%% file: virel.tex
\documentclass{article}
\PassOptionsToPackage{numbers,compress}{natbib}

\usepackage[final]{neurips_2019}
\author{
  Matthew Fellows\thanks{Equal Contribution. Correspondence to \texttt{matthew.fellows@cs.ox.ac.uk} and  \texttt{anuj.mahajan@cs.ox.ac.uk}.} \,\,\,\, Anuj Mahajan\footnotemark[1] \,\,\,\, Tim G. J. Rudner \,\,\,\, Shimon Whiteson\\
  Department of Computer Science\\
  University of Oxford\\
}
\usepackage{hyperref}
\usepackage{amsmath}
\usepackage{url}
\usepackage{amsfonts}   
\usepackage{nicefrac}       
\usepackage{microtype}      
\usepackage{times}
\usepackage{helvet}
\usepackage{courier}
\usepackage{url}
\usepackage{blkarray}
\usepackage{amssymb}
\usepackage{amsthm}
\usepackage[font=small]{caption}
\usepackage{natbib}
\usepackage[capitalize]{cleveref}
\usepackage{autonum}
\usepackage{mathtools}
\usepackage{siunitx}
\usepackage[colorinlistoftodos]{todonotes}
\usepackage{enumitem}
\usepackage{siunitx}
\usepackage{dsfont}
\usepackage[english]{babel}
\usepackage[parfill]{parskip}
\usepackage{bm}
\usepackage{mathrsfs}
\usepackage{multicol}
\usepackage{relsize}

\usepackage{graphicx}
\usepackage{makecell}
\usepackage{array}
\usepackage{booktabs}
\usepackage{subfigure} 
\usepackage{dblfloatfix}
\usepackage{tablefootnote}
\usepackage[flushleft]{threeparttable}
\usepackage[font=small]{caption}
\usepackage{multirow}
\usepackage{wrapfig}
\graphicspath{{figures/}}

\usepackage{algorithm}
\usepackage{algorithmic}
\usepackage{changepage}
\usepackage{bbm}

\newtheorem{assumption}{Assumption}
\newtheorem{theorem}{Theorem}
\newtheorem{lemma}{Lemma}

\newtheorem{definition}{Definition}

\newcommand{\pw}{p_\omega}
\newcommand{\kl}{\textsc{KL}\infdivx}
\newcommand{\elbo}[1]{\textsc{ELBO}\left( #1 \right)}

\newcommand{\closer}[3]{{\kern-#1ex{#2}\kern-#3ex}}
\newcommand{\p}{\pw(h)}

\newcommand{\q}{q_\theta(h)}

\newcommand{\hQ}{\hat{Q}_\omega(h)}
\newcommand{\hQd}{\hat{Q}_\omega(h')}

\newcommand{\pit}{\pi_\theta(a\vert s)}
\newcommand{\pip}{\pi_\phi(a\vert s)}

\newcommand{\piw}{\pi_\omega(a\vert s)}
\newcommand{\piwk}{\pi_{\omega,k}(a\vert s)}

\newcommand{\bw}{\beta_{\omega}(h)}
\newcommand{\bwk}{\beta_{\omega,k}(h)}

\newcommand{\ew}{\varepsilon_\omega}
\newcommand{\ek}{\varepsilon_k}
\newcommand{\ewk}{\varepsilon_{\omega,k}}
\newcommand{\he}{\hat{\varepsilon}_\omega}
\newcommand{\tw}{\mathcal{T}_\omega}
\newcommand{\twk}{\mathcal{T}_{\omega,k}}

\newcommand{\pwk}{p_{\omega,k}(a\vert s)}

\newcommand{\qt}{q_\theta(h)}
\newcommand{\lp}{\mathcal{L}(\omega,\theta)}
\newcommand{\stopgrad}{\mathlarger{\boldsymbol{\dashv}}}
\DeclareMathOperator*{\argmax}{arg\,max}
\DeclareMathOperator*{\argmin}{arg\,min}
\DeclarePairedDelimiterX{\infdivx}[2]{(}{)}{%
  #1\;\delimsize\|\;#2%
}

\usepackage{tikz}
\newcommand*\circled[1]{\tikz[baseline=(char.base)]{
            \node[shape=circle,draw,inner sep=0.5pt] (char) {#1};}}

\usepackage{comment}

\title{\textsc{virel}: A Variational Inference Framework\\for Reinforcement Learning}

\begin{document}
\maketitle

\input{sections/abstract.tex}
\input{sections/introduction.tex}
\input{sections/background.tex}
\input{sections/model.tex}
\input{sections/actor_critic.tex}
\input{sections/experiments.tex}
\input{sections/conclusion.tex}
\input{sections/acknowledgements.tex}

\bibliography{virel.bib}
\bibliographystyle{utilities/icml2019.bst}
\newpage
\input{appendix/appendix.tex}
\typeout{get arXiv to do 4 passes: Label(s) may have changed. Rerun}
\end{document}

%% file: sections/abstract.tex
\vspace{-0.55cm}
\begin{abstract}
Applying probabilistic models to reinforcement learning (RL) enables the uses of powerful optimisation tools such as variational inference in RL. However, existing inference frameworks and their algorithms pose significant challenges for learning optimal policies, for example, the lack of mode capturing behaviour in pseudo-likelihood methods, difficulties learning deterministic policies in maximum entropy RL based approaches, and a lack of analysis when function approximators are used. We propose \textsc{virel}, a theoretically grounded inference framework for RL that utilises a parametrised action-value function to summarise future dynamics of the underlying MDP, generalising existing approaches. \textsc{virel} also benefits from a mode-seeking form of KL divergence, the ability to learn deterministic optimal polices naturally from inference, and the ability to optimise value functions and policies in separate, iterative steps. Applying variational expectation-maximisation to \textsc{virel}, we show that the actor-critic algorithm can be reduced to expectation-maximisation, with policy improvement equivalent to an E-step and policy evaluation to an M-step. We  derive a family of actor-critic methods from \textsc{virel}, including a scheme for adaptive exploration and demonstrate that our algorithms outperform state-of-the-art methods based on soft value functions in several domains.
\end{abstract}

%% file: sections/introduction.tex
\vspace{-0.3cm}
\section{Introduction}
\label{sec:introduction}
\vspace{-0.1cm}
Efforts to combine reinforcement learning (RL) and probabilistic inference have a long history, spanning diverse fields such as control, robotics, and RL \citep{Toussaint06b,Toussaint09a,Peters07,Rawlik10,Heess12,Ziebart08,Ziebart10a,Ziebart10b,Levine13}. Formalising RL as probabilistic inference enables the application of many approximate inference tools to reinforcement learning, extending models in flexible and powerful ways \citep{virl_review}. 
However, existing methods at the intersection of RL and inference suffer from several deficiencies.  Methods that derive from the pseudo-likelihood inference framework \citep{Dayan97, Toussaint06b, Peters07, Hachiya09, Neumann11, Abdolmaleki18} and use expectation-maximisation (EM)  favour risk-seeking policies \citep{Levine14}, which can be suboptimal. Yet another approach, the MERL inference framework \citep{virl_review} (which we refer to as \textsc{merlin}), derives from maximum entropy reinforcement learning (MERL) \citep{Koller00, Ziebart08,Ziebart10a,Ziebart10b}. While \textsc{merlin} does not suffer from the issues of the pseudo-likelihood inference framework, it presents different practical difficulties. These methods do not naturally learn deterministic optimal policies and  constraining the variational policies to be deterministic renders inference intractable \citep{Rawlik10}. As we show by way of counterexample in \cref{sec:MERL}, an optimal policy under the reinforcement learning objective is not guaranteed from the optimal MERL objective. Moreover, these methods rely on soft value functions which are sensitive to a pre-defined temperature hyperparameter. 

Additionally, no existing framework formally accounts for replacing exact value functions with function approximators in the objective; learning function approximators is carried out independently of the inference problem and no analysis of convergence is given for the corresponding algorithms. 

This paper addresses these deficiencies. We introduce \textsc{virel}, an inference framework that translates the problem of finding an optimal policy into an inference problem. Given this framework, we demonstrate that applying EM induces a family of actor-critic algorithms, where the E-step corresponds exactly to policy improvement and the M-step exactly to policy evaluation. Using a variational EM algorithm, we derive analytic updates for both the model and variational policy parameters, giving a unified approach to learning parametrised value functions and optimal policies.

We extensively evaluate two algorithms derived from our framework against DDPG \citep{lillicrap2015continuous} and an existing state-of-the-art actor-critic algorithm, soft actor-critic (SAC) \cite{Haarnoja18}, on a variety of OpenAI gym domains \citep{openai}. While our algorithms perform similarly to SAC and DDPG on simple low dimensional tasks, they outperform them substantially on complex, high dimensional tasks.

The main contributions of this work are: 1) an exact reduction of entropy regularised RL to probabilistic inference using value function estimators; 
2) the introduction of a theoretically justified general framework for developing inference-style algorithms for RL that incorporate the uncertainty in the optimality of the action-value function, $\hQ$, to drive exploration, but that can also learn optimal deterministic policies; and
3) a family of practical algorithms arising from our framework that adaptively balances exploration-driving entropy with the RL objective and outperforms the current state-of-the-art SAC, reconciling existing advanced actor critic methods like A3C \cite{Mnih16}, MPO \citep{Abdolmaleki18} and EPG \cite{epg} into a broader theoretical approach. 

%% file: sections/background.tex
\section{Background}
\label{sec:background}

We assume familiarity with probabilistic inference \citep{Jordan1999} and provide a review in \cref{sec:inference_background}.

\subsection{Reinforcement Learning}
\label{sec:rl_objective}

Formally, an RL problem is modelled as a Markov decision process (MDP) defined by the tuple $\langle \mathcal{S},\mathcal{A},r,p,p_0,\gamma \rangle$ \citep{sutton, rl_algorithms}, where $\mathcal{S}$ is the set of states and $\mathcal{A}\subseteq \mathbb{R}^n$ the set of available actions.
An agent in state $s\in\mathcal{S}$ chooses an action $a\in\mathcal{A}$ according to the policy $a\sim\pi(\cdot\vert s)$, forming a state-action pair $h\in\mathcal{H}$, $h\coloneqq\langle s,a\rangle$.
This pair induces a scalar reward according to the reward function $r_t\coloneqq r(h_t)\in\mathbb{R}$ and the agent transitions to a new state $s'\sim p(\cdot\vert h)$.
The initial state distribution for the agent is given by $s_0\sim p_0$.
We denote a sampled state-action pair at timestep $t$ as $h_t\coloneqq\langle s_t,a_t\rangle$.
As the agent interacts with the environment using $\pi$, it gathers a trajectory $\tau=(h_0,r_0,h_1,r_1,...)$. The value function is the expected, discounted reward for a trajectory, starting in state $s$. The action-value function or $Q$-function is the expected, discounted reward for each trajectory, starting in $h$, $
Q^\pi(h)\coloneqq \mathbb{E}_{\tau\sim p^\pi(\tau\vert h)}\left[\sum_{t=0}^{\infty}\gamma^tr_t\right]$, where $p^\pi(\tau\vert h)\coloneqq p(s_1\vert h_0=h)\prod_{t'=1}^{\infty}p(s_{t'+1}\vert h_{t'})\pi(a_t\vert s_t)$. Any $Q$-function satisfies a Bellman equation $\mathcal{T}^\pi Q^\pi(\cdot)=Q^\pi(\cdot)$ where $\mathcal{T}^\pi\boldsymbol{\cdot}\coloneqq r(h)+\gamma \mathbb{E}_{h'\sim p(s'\vert h)\pi(a'\vert s')}\left[\boldsymbol{\cdot}\right]$ is the Bellman operator. We consider infinite horizon problems with a discount factor $\gamma\in[0,1)$.
The agent seeks an optimal policy $\pi^{*} \in \argmax_{\pi}{J^\pi}$, where
\begin{align}
J^\pi=\mathbb{E}_{h\sim p_0(s)\pi(a\vert s)}\left[ Q^\pi(h)\right]. \label{eq:rl_objective} 
\end{align}
We denote optimal $Q$-functions as $Q^*(\cdot)\coloneqq Q^{\pi^*}(\cdot)$ and the set of optimal policies $\Pi^*\coloneqq\argmax_{\pi}{J^\pi}$. The optimal Bellman operator is $\mathcal{T}^*\boldsymbol{\cdot}\coloneqq r(h)+\gamma \mathbb{E}_{h'\sim p(s'\vert h)}\left[\max_{a'}(\boldsymbol{\cdot})\right]$.

\subsection{Maximum Entropy RL}
\label{sec:MERL}
The MERL objective supplements each reward in the RL objective with an entropy term \citep{Todorov06, Ziebart08,Ziebart10a, Ziebart10b}, $J_{\textrm{merl}}^\pi\coloneqq\mathbb{E}_{\tau\sim p(\tau)}\left[\sum_{t=0}^{T-1}\left( r_t-c\log(\pi (a_t\vert s_t)\right)\right]$. The standard RL, undiscounted objective is recovered for $c\rightarrow0$ and we assume $c=1$ without loss of generality. The MERL objective is often used to motivate the MERL inference framework (which we call \textsc{merlin}) \citep{Levine14}, mapping the problem of finding the optimal policy, $\pi_{\mathrm{merl}}^*(a\vert s)=\argmax_\pi J^\pi_{\textrm{merl}}$, to an equivalent inference problem. A full exposition of this framework is given by \citet{virl_review} and we discuss the graphical model of $\textsc{merlin}$ in comparison to $\textsc{virel}$ in \cref{sec:comparing_virel_merlin}. The inference problem is often solved using a message passing algorithm, where the log backward messages are called soft value functions due to their similarity to classic (hard) value functions \citep{Toussaint09b, Rawlik12, Haarnoja18, Haarnoja17,virl_review}. The soft $Q$-function is defined as
$Q_{\mathrm{soft}}^\pi(h)\coloneqq\mathbb{E}_{\tau\sim q^\pi(\tau\vert h)}\left[ r_0+\sum_{t=1}^{T-1}(r_t-\log \pi(a_t\vert s_t))\right]$,
where $q^\pi(\tau\vert h)\coloneqq p(s_0\vert h)\prod_{t=0}^{T-1}p(s_{t+1}\vert h_t)\pi(a_t\vert s_t)$. The corresponding soft Bellman operator is $\mathcal{T}^{\pi}_{\mathrm{soft}}\boldsymbol{\cdot}\coloneqq r(h)+\mathbb{E}_{h'\sim p(s'\vert h)\pi(a'\vert s')}[\boldsymbol{\cdot}-\log\pi(a'\vert s')]$. Several algorithms have been developed that mirror existing RL algorithms using soft Bellman equations, including maximum entropy policy gradients \citep{virl_review}, soft $Q$-learning \citep{Haarnoja17}, and soft actor-critic (SAC) \citep{Haarnoja18}. MERL is also compatible with methods that use recall traces \citep{Goyal18}. 
 \begin{wrapfigure}{l}{0.45\textwidth}
	\centering
	\vspace{-0.2cm}
		\includegraphics[width=0.45\columnwidth]{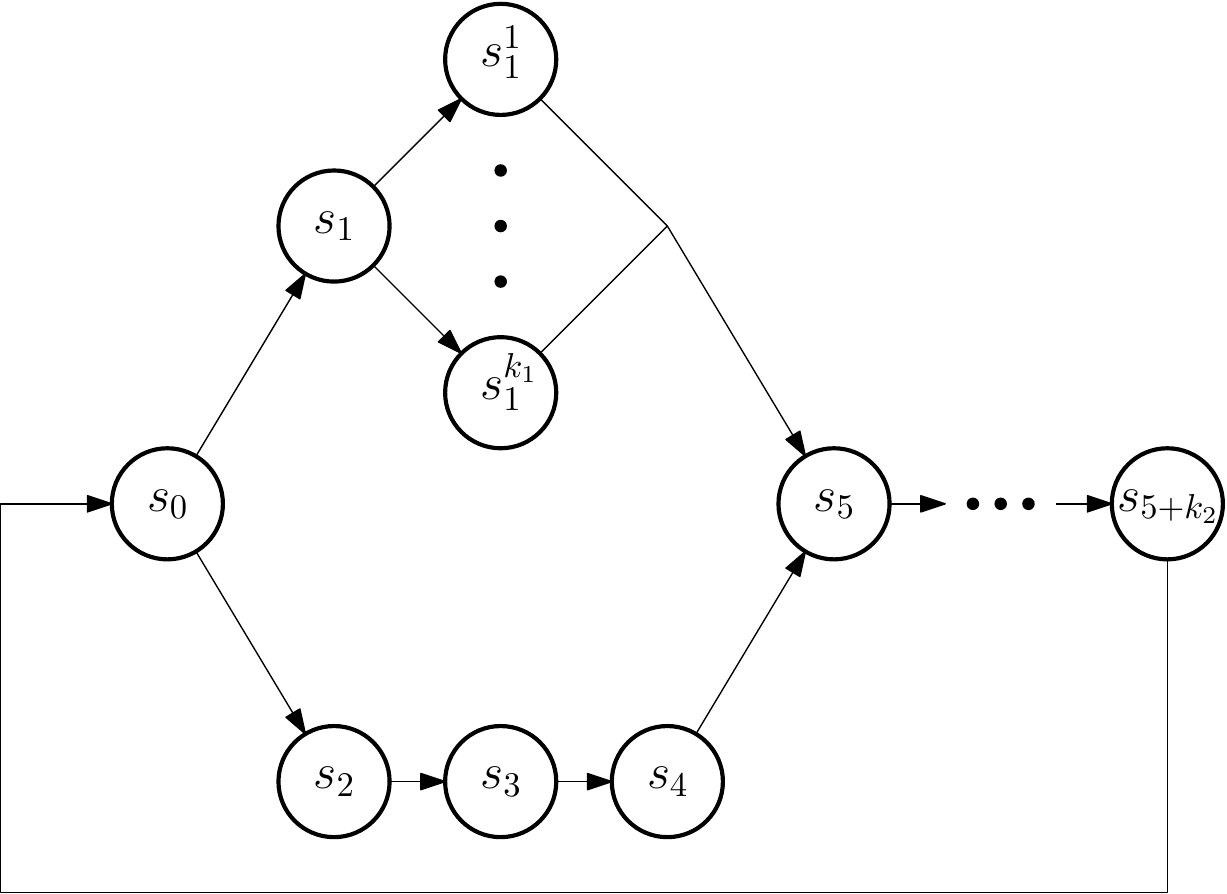}
	\caption{A discrete MDP counterexample for optimal policy under maximum entropy.}
	\label{ce}
	\vspace{-0.5cm}
\end{wrapfigure}

We now outline key drawbacks of \textsc{merlin}. It is well-understood that optimal policies under regularised Bellman operators are more stochastic than under their equivalent unregularised operators \citep{geist19a}. While this can lead to improved exploration, the optimal policy under these operators will still be stochastic, meaning optimal deterministic policies are not learnt naturally. This leads to two difficulties: 1) a deterministic policy can be constructed by taking the action $a^*=\argmax_{a}\pi^*_{\textrm{merl}}(a\vert s)$, corresponding to the maximum a posteriori (MAP) policy, however, in continuous domains, finding the MAP policy requires optimising the $Q$-function approximator for actions, which is often a deep neural network. A common approximation is to use the mean of a variational policy instead; 2) even if we obtain a good approximation, as we show below by way of counterexample, the deterministic MAP policy is not guaranteed to be the optimal policy under $J^\pi$. Constraining the variational policies to the set of Dirac-delta distributions does not solve this problem either, since it renders the inference procedure intractable \citep{Rawlik10,Rawlik12}.

Next, we demonstrate that the optimal policy under $J^{\pi}$ cannot always be recovered from the MAP policy under $J^{\pi}_\textrm{merl}$. Consider the discrete state MDP as shown in \cref{ce}, with action set $\mathcal{A}=\{a_1,a_2,a_1^1,\cdots a_1^{k_1}\}$ and state set $\mathcal{S}=\{s_0,s_1,s_2,s_3,s_4,s_1^1\cdots s_1^{k_1},s_5,\cdots s_{5+{k_2}} \}$. All state transitions are deterministic, with $p(s_1\vert s_0, a_1)=p(s_1\vert s_0, a_2)=p(s_1^i\vert s_1, a_1^i)=1$. All other state transitions are deterministic and independent of action taken, that is, $p(s_j\vert \cdot, s_{j-1})=1\ \forall\ {j>2}$ and $p(s_5\vert \cdot, s_1^{i})=1$. The reward function is $r(s_0,a_2) = 1$ and zero otherwise. Clearly the optimal policy under $J^\pi$ has $\pi^*(a_2|s_0) = 1$. Define a maximum entropy reinforcement learning policy as $\pi_{\textrm{merl}}$ with  $\pi_{\textrm{merl}}(a_1|s_0)=p_1$, $\pi_{\textrm{merl}}(a_2|s_0)=(1-p_1)$ and $\pi_{\textrm{merl}}(a_1^i|s_1)=p_1^i$. For $\pi_{\textrm{merl}}$ and $k_2>>5$, we can evaluate $J^\pi_\textrm{merl}$ for any scaling constant $c$ and discount factor $\gamma$ as:
\begin{align}
    J^\pi_\textrm{merl}=(1-p_1)(1-c\log(1-p_1))-p_1\left(c\log p_1 +\gamma c\sum_{i=1}^k p_1^i \log p_1^i\right). \label{eq:counter_objective}
\end{align}
We now find the optimal MERL policy. Note that  $p_1^i=\frac{1}{k}$ maximises the final term in \cref{eq:counter_objective}. Substituting for $p_1^i=\frac{1}{k_1}$, then taking derivatives of \cref{eq:counter_objective} with respect to $p_1$, and setting to zero, we find $p_1^*=\pi_{\textrm{merl}}^*(a_1\vert s_0)$ as:
\begin{align}
    1-c\log(1-p_1^*)&=\gamma c\log(k_1)-c\log p_1^*,\\
    \implies p_1^*&=\frac{1}{{k_1}^{-\gamma}\exp\left(\frac{1}{c}\right)+1},
\end{align}
hence, for any ${k_1}^{-\gamma}\exp\left(\frac{1}{c}\right) < 1$, we have $ p_1^*>\frac{1}{2}$ and so $\pi^*$ cannot be recovered from $\pi_{\textrm{merl}}^*$, even using the mode action $a_1=\argmax_a\pi_{\textrm{merl}}^*(a\vert s_0)$. The degree to which the MAP policy varies from the optimal unregularised policy depends on both the value of $c$ and $k_1$, the later controlling the number of states with sub-optimal reward. Our counterexample illustrates that when there are large regions of the state-space with sub-optimal reward, the temperature must be comparatively small to compensate, hence algorithms derived from \textsc{merlin} become very sensitive to temperature. As we discuss in \cref{sec:comparing_virel_merlin}, this problem stems from the fact that \textsc{merl} policies optimise for expected reward and long-term expected entropy. While initially beneficial for exploration, this can lead to sub-optimal polices being learnt in complex domains as there is often too little a priori knowledge about the MDP to make it possible to choose an appropriate value or schedule for $c$. 

Finally, a minor issue with \textsc{merlin} is that many existing models are defined for finite-horizon problems \citep{virl_review, Rawlik12}. While it is possible to discount and extend \textsc{merlin} to infinite-horizon problems, doing so is often nontrivial and can alter the objective \citep{Thomas14, Haarnoja18}.
\vspace{-0.2cm}
\subsection{Pseudo-Likelihood Methods}
\vspace{-0.1cm}
\label{sec:PL}

A related but distinct approach is to apply Jensen's inequality directly to the RL objective $J^\pi$. Firstly, we rewrite \cref{eq:rl_objective} as an expectation over $\tau$ to obtain $J=\mathbb{E}_{h\sim p_0(s)\pi(a\vert s)}\left[ Q^\pi(h)\right]=\mathbb{E}_{\tau\sim p(\tau)}\left[ R(\tau) \right]$, where $R(\tau)=\sum_{t=0}^{T-1}\gamma^tr_t$ and $p(\tau)=p_0(s_0)\pi(a_0\vert s_o)\prod^{T-1}_{t=0}p(h_{t+1}\vert h_{t})$. We then treat $p(R,\tau)=R(\tau)p(\tau)$ as a joint distribution, and if rewards are positive and bounded, Jensen's inequality can be applied, enabling the derivation of an evidence lower bound (ELBO). Inference algorithms such as EM can then be employed to find a policy that optimises the pseudo-likelihood objective \citep{Dayan97, Toussaint06b,Peters07,Hachiya09,Neumann11,Abdolmaleki18}. Pseudo-likelihood methods can also be extended to a model-based setting by defining a prior over the environment's transition dynamics. \citet{Furmston2010} demonstrate that the posterior over all possible environment models can be integrated over to obtain an optimal policy in a Bayesian setting. 

Many pseudo-likelihood methods minimise $\kl{p_{\mathcal{O}}}{p_{\pi}}$, where $p_{\pi}$ is the policy to be learnt and $p_{\mathcal{O}}$ is a target distribution monotonically related to reward \citep{virl_review}. Classical RL methods minimise $\kl{p_{\pi}}{p_{\mathcal{O}}}$. The latter encourages learning a mode of the target distribution, while the former encourages matching the moments of the target distribution. If the optimal policy can be represented accurately in the class of policy distributions, optimisation converges to a global optimum and the problem is fully observable, the optimal policy is the same in both cases. Otherwise, the  pseudo-likelihood objective reduces the influence of large negative rewards, encouraging risk-seeking policies.

%% file: sections/model.tex
\vspace{-0.6cm}
\section{\textsc{virel}}
\vspace{-0.2cm}
\label{sec:rl_as_inference}
Before describing our framework, we state some relevant assumptions.
\begin{assumption}[Strictly Positive Optimal $Q$-function]\label{assumption:positive}
The optimal action-value function has a compact domain $\mathcal{H}$ and is finite and strictly positive, i.e., $0<Q^*(h)<\infty\  \forall\ h\in\mathcal{H}$. 
\end{assumption}
\vspace{-0.2cm}
MDPs for which rewards are lower bounded and finite, that is, $R\subset[r_\mathrm{min},\infty)$, satisfy the second condition of \cref{assumption:positive}; we can construct a new MDP by adding $r_\mathrm{min}$ to the reward function, ensuring that all rewards are positive and hence the optimal $Q$-function for the reinforcement learning problem is finite and strictly positive. This does not affect the optimal solution.  Now we introduce a function approximator $\hQ\approx Q^\pi(h)$ in the bounded function space $\hQ\in\mathcal{Q}_\Omega$:
\begin{definition}[Bounded Function Space] Let $\mathcal{Q}_\Omega\coloneqq\left\{f:\Omega\times\mathcal{H}\rightarrow\mathbb{R}\ \Big\vert\ \lvert f\rvert <\infty\ \forall\  \omega\in\Omega,h\in\mathcal{H} \right\}$ be the space of bounded, Lipschitz continuous functions parametrised by $\omega\in\Omega$.
\end{definition}
\begin{assumption}[Representabilty Under Optimisation]\label{assumption:policy_represent_optimal}
Our function approximator can represent the optimal $Q$-function, i.e., $\exists\ \omega^*\in\Omega\ s.t.\ Q^*(\cdot)=\hat{Q}_{\omega^*}(\cdot)$. 
\end{assumption}
\vspace{-0.2cm}
In \cref{sec:relaxation_assumption}, we extend the work of \citet{Maei09} to continuous domains, demonstrating that \cref{assumption:policy_represent_optimal} can be neglected if projected Bellman operators are used. 

\begin{assumption}[Local Smoothness of $Q$-functions ]\label{assumption:smoothness} For $\omega^*$ parametrising $Q^*(\cdot)$ in \cref{assumption:policy_represent_optimal}, $\hat{Q}_{\omega^*}(\cdot)$ has a unique maximum $\hat{Q}_{\omega^*}(h^*)=\sup_h \hat{Q}_{\omega^*}(h)$. $\hat{Q}_{\omega^*}(\cdot)$ is locally $\mathbb{C}^2$-smooth about $h^*$, i.e. $\exists\ \Delta>0\ s.t.\ \hat{Q}_{\omega^*}(h)\in\mathbb{C}^2\ \forall\ h\in\{h\vert \lVert h-h^*\rVert<\Delta$ \}.
\end{assumption}
\vspace{-0.2cm}
This assumption is formally required for the strict convergence of a Boltzmann to a Dirac-delta distribution and, as we discuss in \cref{sec:approx_local_smoothness}, is of more mathematical than practical concern. 
\vspace{-0.3cm}
\subsection{Objective Specification}
\vspace{-0.1cm}
\label{sec:model}
 We now define an objective that we motivate by satisfying three desiderata: \circled{1} In the limit of maximising our objective, a deterministic optimal policy can be recovered and the optimal Bellman equation is satisfied by our function approximator; \circled{2} when our objective is not maximised, stochastic policies can be recovered that encourage effective exploration of the state-action space; and \circled{3} our objective permits the application of powerful and tractable optimisation algorithms from variational inference that optimise the risk-neutral form of KL divergence, $\kl{p_{\pi}}{p_{\mathcal{O}}}$, introduced in \cref{sec:PL}. 
 
 Firstly, we define the residual error $\ew\coloneqq \frac{c}{p}\lVert\tw\hQ-\hQ\rVert_p^p$ where $\tw=\mathcal{T}^{\pi_\omega}\boldsymbol{\cdot}\coloneqq r(h)+\gamma \mathbb{E}_{h'\sim p(s'\vert h)\pi_\omega(a'\vert s')}\left[\boldsymbol{\cdot}\right]$ is the Bellman operator for the Boltzmann policy with temperature $\ew$:
 \begin{align}
\piw \coloneqq \frac{\exp\left(\frac{\hQ}{\ew}\right)}{\int_\mathcal{A} \exp\left(\frac{\hQ}{\ew}\right)da},\label{eq:action_posterior}
\end{align}  
where $c>0$ is an arbitrary finite constant and we assume $p=2$ and without loss of generality. Our main result in \cref{proof:forward} proves that finding a $\omega^*$ that reduces the residual error to zero, i.e., $\varepsilon_{\omega^*}=0$, is a sufficient condition for learning an optimal $Q$-function $\hat{Q}_{\omega^*}(h)=Q^*(h)$. Additionally, the Boltzmann distribution $\piw$ tends towards a Dirac-delta distribution $\piw=\delta(a=\argmax_a' \hat{Q}_{\omega^*}(a',s))$ whenever $\ew\rightarrow0$ (see \cref{proof:convergence_boltzmann}), which is an optimal policy. The simple objective $\argmin(\mathcal{L}(\omega))\coloneqq\argmin(\ew)$ therefore satisfies \circled{1}. Moreover, when our objective is not minimised, we have $\ew>0$ and from \cref{eq:action_posterior} we see that $\piw$ is non-deterministic \textit{for all non-optimal $\omega$}. $\mathcal{L}(\omega)$ therefore satisfies \circled{2} as any agent following $\piw$ will continue exploring until the RL problem is solved. To generalise our framework, we extend $\tw\boldsymbol{\cdot}$ to any operator from the set of target operators $\tw\boldsymbol{\cdot}\in \mathbb{T}$:

\begin{definition}[Target Operator Set]\label{def:target_set}
Define $\mathbb{T}$ to be the set of target operators such that an optimal Bellman operator for $\hQ$ is recovered when the Boltzmann policy in \cref{eq:action_posterior} is greedy with respect to $\hQ$, i.e., $\mathbb{T}\coloneqq\{\tw\boldsymbol{\cdot}\vert\lim_{\ew\rightarrow0}\piw\implies \tw \hQ=\mathcal{T}^*\hQ\}$.
\end{definition}
As an illustration, we prove in \cref{sec:discussion_T} that the Bellman operator $\mathcal{T}^{\pi_\omega}\boldsymbol{\cdot}$ introduced above is a member of $\mathbb{T}$ and can be approximated by several well-known RL targets. We also discuss how $\mathcal{T}^{\pi_\omega}\boldsymbol{\cdot}$ induces a constraint on $\Omega$ due to its recursive definition. As we show in \cref{sec:analysis}, there exists an $\omega$ in the constrained domain that maximises the RL objective under these conditions, so an optimal solution is always feasible. Moreover, we provide an analysis in \cref{sec:analysis_of_approximations} to establish that such a policy is an attractive fixed point for our algorithmic updates, even when we ignore this constraint. Off-policy operators will not constrain $\Omega$: by definition, the optimal Bellman operator $\mathcal{T}^*\boldsymbol{\cdot}$ is a member of $\mathbb{T}$ and does not constrain $\Omega$; similarly, we derive an off-policy operator based on a Boltzmann distribution with a diminishing temperature in \cref{sec:relaxation_constraints} that is a member of  $\mathbb{T}$. Observe that soft Bellman operators are not members of $\mathbb{T}$ as the optimal policy under $J_\text{merl}^\pi$ is not deterministic, hence algorithms such as SAC cannot be derived from the \textsc{virel} framework.

One problem remains: calculating the normalisation constant to sample directly from the Boltzmann distribution in \cref{eq:action_posterior} is intractable for many MDPs and function approximators. As such, we look to variational inference to learn an approximate \textit{variational policy} $\pit\approx\piw$, parametrised by $\theta\in\Theta$ with finite variance and the same support as $\piw$. This suggests optimising a new objective that penalises $\pit$ when $\pit\ne\piw$ but still has a global maximum at $\ew=0$. A tractable objective that meets these requirements is the evidence lower bound (ELBO) on the unnormalised potential of the Boltzmann distribution, defined as $\{\omega^*,\theta^*\}\in\argmax_{\omega,\theta}\mathcal{L}(\omega,\theta)$, \vspace{-0.2cm}
\begin{align}
 \mathcal{L}(\omega,\theta)&\coloneqq\mathbb{E}_{s\sim d(s)}\left[\mathbb{E}_{a\sim \pit} \left[\frac{\hQ}{\ew}\right]+\mathscr{H}(\pit)\right],\label{eq:elborl}
\end{align}
where $\qt\coloneqq d(s)\pit$ is a variational distribution,  $\mathscr{H}(\boldsymbol{\cdot})$ denotes the differential entropy of a distribution, and $d(s)$ is any arbitrary sampling distribution with support over $\mathcal{S}$. From \cref{eq:elborl}, maximising our objective with respect to $\omega$ is achieved when $\ew\rightarrow0$ and hence $\mathcal{L}(\omega,\theta)$  satisfies \circled{1} and \circled{2}. As we show in \cref{proof:objective_infinity}, $\mathscr{H}(\cdot)$ in \cref{eq:elborl} causes $\lp\rightarrow-\infty$ whenever $\pi_\theta(a\vert s)$ is a Dirac-delta distribution for all $\ew>0$. This means our objective heavily penalises premature convergence of our variational policy to greedy Dirac-delta policies except under optimality. We discuss a probabilistic interpretation of our framework in \cref{sec:probablistic_interpretation}, where it can be shown that $\piw$ characterises our model's uncertainty in the optimality of $\hQ$.

We now motivate $\lp$ from an inference perspective: In \cref{sec:derivation_elbo}, we write $\lp$ in terms of the log-normalisation constant of the Boltzmann distribution and the KL divergence between the action-state normalised Boltzmann distribution, $\pw(h)$, and the variational distribution, $\qt$:
\begin{gather}
    \mathcal{L}(\omega,\theta)=\ell(\omega)-\kl{q_\theta(h)}{\pw(h)}-\mathscr{H}(d(s)),\label{eq:objective_ii}\\
    \mathrm{where}\quad\mathcal{\ell}(\omega)\coloneqq\log\int_\mathcal{H}\exp\left(\frac{\hQ}{\ew}\right)dh,\quad \pw(h)\coloneqq \frac{\exp\left(\frac{\hQ}{\ew}\right)}{\int_\mathcal{H}\exp\left(\frac{\hQ}{\ew}\right)dh}.
\end{gather}
As the KL divergence in \cref{eq:objective_ii} is always positive and the final entropy term has no dependence on $\omega$ or $\theta$, maximising our objective for $\theta$ always reduces the KL divergence between $\piw$ and $\pit$ for any $\ew>0$, with $\pit=\piw$ achieved under exact representability (see \cref{proof:policy_posterior}). This yields a tractable way to estimate $\piw$ at any point during our optimisation procedure by maximising $\lp$ for $\theta$. From \cref{eq:objective_ii}, we see that our objective satisfies \circled{3}, as we minimise the mode-seeking direction of KL divergence, $\kl{\qt}{\pw(h)}$, and our objective is an ELBO, which is the starting point for inference algorithms \citep{Jordan1999,Beal03,vi_tutorial}. When the RL problem is solved and $\ew=0$, our objective tends towards infinity for \emph{any} variational distribution that is non-deterministic (see \cref{proof:objective_infinity}). This is of little consequence, however, as whenever $\ew=0$, our approximator is the optimal value function, $\hat{Q}_{\omega^*}(h)=Q^*(h)$ (\cref{proof:forward}), and hence, $\pi^*(a\vert s)$ can be inferred exactly by finding $\max_{a'}\hat{Q}_{\omega^*}(a',s)$ or by using the policy gradient $\nabla_\theta\mathbb{E}_{d(s)\pit}\left[\hat{Q}_{\omega^*}(h)\right]$ (see \cref{sec:ac_discuss}). 
\vspace{-0.6cm}
\subsection{Theoretical Results}
\label{sec:analysis}
\vspace{-0.1cm}
We now formalise the intuition behind \circled{1}-\circled{3}. \cref{proof:convergence_boltzmann} establishes the emergence of a Dirac-delta distribution in the limit of $\ew\rightarrow0$. To the authors' knowledge, this is the first rigorous proof of this result. \cref{proof:forward} shows that finding an optimal policy that maximises the RL objective in \cref{eq:rl_objective} reduces to finding the Boltzmann distribution associated with the parameters $\omega^*\in\argmax_{\omega}\lp$. The existence of such a distribution is a sufficient condition for the policy to be optimal. \cref{proof:policy_posterior} shows that whenever $\ew>0$, maximising our objective for $\theta$ always reduces the KL divergence between $\piw$ and $\pit$, providing a tractable method to infer the current Boltzmann policy. 

\begin{theorem}[Convergence of Boltzmann Distribution to Dirac Delta]\label{proof:convergence_boltzmann}
Let $p_\varepsilon:\mathcal{X}\rightarrow[0,1]$ be a Boltzmann distribution with temperature $\varepsilon\in\mathbb{R}_{\ge0}$, $p_\varepsilon(x)=\frac{\exp\left(\frac{f(x)}{\varepsilon}\right)}{\int_\mathcal{X}\exp\left(\frac{f(x)}{\varepsilon}\right)dx},$ where $f:\mathcal{X}\rightarrow\mathcal{Y}$ is a function that satisfies \cref{def:locally_smooth}. In the limit $\varepsilon\rightarrow 0$, $p_\varepsilon(x)\rightarrow \delta(x=\sup_{x'} f(x'))$.
\begin{proof}
\vspace{-0.4cm}
See \cref{sec:convergence_boltzmann}
\vspace{-0.2cm}
\end{proof}
\end{theorem}
\begin{lemma}[Lower and Upper limits of $\lp$] i) For any $\ew>0$ and $\pit=\delta(a^*)$, we have $\lp=-\infty$. ii) For $\hQ>0$ and any non-deterministic $\pit$, $\lim_{\ew\rightarrow0}\lp=\infty$.
\label{proof:objective_infinity}
\begin{proof}
\vspace{-0.4cm}
See \cref{proof:rl_elbo_ap}.
\vspace{-0.2cm}
\end{proof}
\end{lemma}

\begin{theorem}[Optimal Boltzmann Distributions as Optimal Policies]
	\label{proof:forward}
	For $\omega^*$ that maximises $\lp$ defined in \cref{eq:elborl}, the corresponding Boltzmann policy induced must be optimal, i.e., $\{\omega^*,\theta^*\} \in \argmax_{\omega,\theta}\lp
	\implies  \pi_{\omega^*}(a \vert  s) \in \Pi^*$.
\begin{proof}
\vspace{-0.4cm}
See \cref{proof:rl_elbo_ap}.
\vspace{-0.2cm}
\end{proof}
\end{theorem}
\begin{theorem}[Maximising the ELBO for $\theta$]
\label{proof:policy_posterior}
 For any $\ew>0$, $\max_\theta\lp=\mathbb{E}_{d(s)}\left[\min_\theta\kl{\pit}{\piw}\right]$ with $\piw=\pit$ under exact representability. 
\begin{proof}
\vspace{-0.4cm}
See \cref{sec:proof_kl_app}.
\vspace{-0.2cm}
\end{proof}
\end{theorem}

\subsection{Comparing \textsc{virel} and \textsc{merlin} Frameworks} 
\label{sec:comparing_virel_merlin}
 \begin{wrapfigure}{r}{0.42\textwidth}
\vspace{-1.6cm}
    \begin{center}
    \includegraphics[scale=0.167]{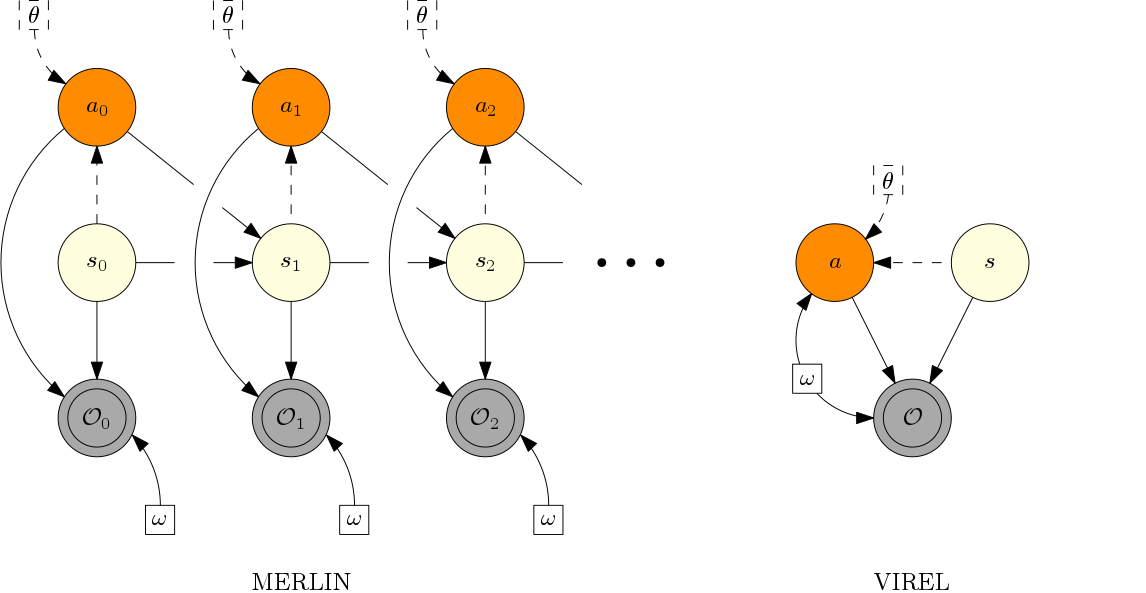}
    \end{center}
  \caption{Graphical models for \textsc{merlin} and \textsc{virel} (variational approximations are dashed).}
  \label{fig:rl_vi_graphs}
  \vspace{-0.5cm}
\end{wrapfigure}

To compare \textsc{merlin} and \textsc{virel}, we consider the probabilistic interpretation of the two models discussed in \cref{sec:probablistic_interpretation}; introducing a binary variable $\mathcal{O}\in\{0,1\}$ defines a graphical model for our inference problem whenever $\ew>0$. Comparing the graphs in \cref{fig:rl_vi_graphs}, observe that \textsc{merlin} models exponential \emph{cumulative} rewards over entire trajectories. By contrast, \textsc{virel}'s variational policy models a single step and a function approximator is used to model future \emph{expected} rewards. The resulting KL divergence minimisation for \textsc{merlin} is therefore much more sensitive to the value of temperature, as this affects how much future entropy influences the variational policy. For \textsc{virel}, temperature is defined by the model, and updates to the variational policy will not be as sensitive to errors in its value or linear scaling as its influence only extends to a single interaction. We hypothesise that \textsc{virel} may afford advantages in higher dimensional domains where there is greater chance of encountering large regions of state-action space with sub-optimal reward; like our counterexample from \cref{sec:background}, $c$ must be comparatively small to balance the influence of entropy in these regions to prevent $\textsc{merlin}$ algorithms from learning sub-optimal policies.

\cref{proof:convergence_boltzmann} demonstrates that, unlike in \textsc{merlin}, \textsc{virel} naturally learns optimal deterministic policies directly from the optimisation procedure while still maintaining the benefits of stochastic policies in training. While Boltzmann policies with fixed temperatures have been proposed before \citep{Sallans04}, the adaptive temperature $\ew$ in \textsc{virel}'s Boltzmann policy is unique in that it ensures adaptive exploration that becomes less stochastic as $\hQ\rightarrow Q^*(h)$. Like in \textsc{merlin}, we discuss a probabilistic interpretation of this policy as a posterior over actions conditioned on $\mathcal{O}=1$ in \cref{sec:probablistic_interpretation}.

%% file: sections/actor_critic.tex
\vspace{-0.2cm}
\section{Actor-Critic and EM}
\vspace{-0.1cm}

\label{sec:acem}
We now apply the expectation-maximisation (EM) algorithm \citep{Dempster77maximumlikelihood,Gunawardana05} to optimise our objective $\lp$. (See \cref{sec:inference_background} for an exposition of this algorithm.) In keeping with RL nomenclature, we refer to $\hQ$ as the \emph{critic} and $\pit$ as the \emph{actor}. We establish that the expectation (E-) step is equivalent to carrying out policy improvement and the maximisation (M-)step to policy evaluation. This formulation reverses the situation in most pseudo-likelihood methods, where the E-step is related to policy evaluation and the M-step is related to policy improvement, and is a direct result of optimising the forward KL divergence, $\kl{q_\theta(h)}{\pw(h\vert \mathcal{O})}$, as opposed to the reverse KL divergence used in pseudo-likelihood methods. As discussed in \cref{sec:PL}, this mode-seeking objective prevents the algorithm from learning risk-seeking policies.
We now introduce an extension to \cref{assumption:policy_represent_optimal} that is sufficient to guarantee convergence.

\begin{assumption}[Universal Variational Representability]\label{assumption:policy_represent}
Every Boltzmann policy can be represented as $\pi_\theta(a\vert s)$, i.e., $\forall\ \omega\in\Omega\ \exists\ \theta\in\Theta\ s.t.\ \pit=\piw$.
\end{assumption}
\cref{assumption:policy_represent} is strong but, like in variational inference, our variational policy $\pi_\theta(a\vert s)$ provides a useful approximation when \cref{assumption:policy_represent} does not hold. As we discuss in \cref{sec:relaxation_assumption}, using projected Bellman errors also ensures that our M-step always converges no matter what our current policy is.

\vspace{-0.3cm}
\subsection{Variational Actor-Critic}
\vspace{-0.1cm}
\label{sec:vacem}
In the E-step, we keep the parameters of our critic $\omega_k$ constant while updating the actor's parameters by maximising the ELBO with respect to $\theta$: $\theta_{k+1}\leftarrow \argmax_\theta\mathcal{L}(\omega_k,\theta)$. Using gradient ascent with step size $\alpha_{\textrm{actor}}$, we optimise $\varepsilon_{\omega_k}\mathcal{L}(\omega_k,\theta)$ instead, which prevents ill-conditioning and does not alter the optimal solution, yielding the update (see \cref{proof:estep} for full derivation):
\vspace{-0.2cm}
\paragraph{ E-Step (Actor):} $\theta_{i+1}\leftarrow\theta_{i} + \alpha_{\textrm{actor}}\left(\varepsilon_{\omega_k}\nabla_\theta \mathcal{L}(\omega_k,\theta)\right)\rvert_{\theta=\theta_i}$,\\
\vspace{-0.2cm}
\begin{align}
\varepsilon_{\omega_k}\nabla_\theta \mathcal{L}(\omega_k,\theta)=\mathbb{E}_{s\sim d(s)}\left[\mathbb{E}_{a\sim\pit}\left[ \hat{Q}_{\omega_k}(h)\nabla_\theta\log\pit\right]+\varepsilon_{\omega_k}\nabla_\theta\mathscr{H}(\pit)\right].\label{eq:e_step}
\end{align}
In the M-step, we maximise the ELBO with respect to $\omega$ while holding the parameters $\theta_{k+1}$ constant. Hence expectations are taken with respect to the variational policy found in the E-step: $\omega_{k+1}\leftarrow \argmax_\omega\mathcal{L}(\omega,\theta_{k+1})$. We use gradient ascent with step size $\alpha_\textrm{critic}(\varepsilon_{\omega_i})^2$ to optimise $\mathcal{L}(\omega,\theta_{k+1})$ to prevent ill-conditioning, yielding (see \cref{proof:mstep} for full derivation):
\paragraph{M-Step (Critic):} $
\omega_{i+1}\leftarrow\omega_{i} + \alpha_{\textrm{critic}}(\varepsilon_{\omega_i})^2\nabla_\omega \mathcal{L}(\omega,\theta_{k+1})\rvert_{\omega=\omega_i},$
\begin{align}
&(\varepsilon_{\omega_i})^2\nabla_\omega \mathcal{L}(\omega,\theta_{k+1})=\varepsilon_{\omega_i}\mathbb{E}_{ d(s)\pi_{\theta_{k+1}}(a\vert s)}\left[ \nabla_\omega\hQ\right]-\mathbb{E}_{ d(s)\pi_{\theta_{k+1}}(a\vert s)}\left[ \hat{Q}_{\omega_i}(h)\right]\nabla_\omega\ew.\label{eq:m_step}
\end{align}
\vspace{-0.6cm}
\subsection{Discussion}
\label{sec:ac_discuss}
From an RL perspective, the E-step corresponds to training an actor using a policy gradient method \citep{Sutton00} with an adaptive entropy regularisation term \citep{Williams91,Mnih16}. The M-step update corresponds to a policy evaluation step, as we seek to reduce the MSBE in the second term of \cref{eq:m_step}. We derive $\nabla_\omega\ew$ exactly in \cref{sec:gradient_residual_error}. Note that this term depends on $(\tw\hQ-\hQ)\nabla_\omega\tw\hQ$, which typically requires evaluating two independent expectations. For convergence guarantees, techniques such as residual gradients \citep{Baird95} or GTD2/TDC \citep{Maei09} need to be employed to obtain an unbiased estimate of this term. If guaranteed convergence is not a priority, dropping gradient terms allows us to use direct methods \citep{Sutton17}, which are often simpler to implement. We discuss these methods further in \cref{sec:approx_gradients} and provide an analysis in \cref{sec:analysis_of_approximations} demonstrating that the corresponding updates act as a variational approximation to $Q$-learning \citep{qlearning,dqn}. A key component of our algorithm is the behaviour when $\varepsilon_{\omega^*}=0$; under this condition, there is no M-step update (both $\varepsilon_{\omega_k}=0$ and $\nabla_\omega\ew=0$) and $Q_{\omega^*}(h)=Q^*(h)$ (see \cref{proof:forward}), so our E-step reduces exactly to a policy gradient step, $\theta_{k+1}\leftarrow\theta_{k} + \alpha_{\textrm{actor}}\mathbb{E}_{h\sim d(s)\pit}\left[ Q^*(h)\nabla_\theta\log\pit\right]$, recovering the optimal policy in the limit of convergence, that is, $\pit\rightarrow\pi^*(a\vert s)$.

From an inference perspective, the E-step improves the parameters of our variational distribution to reduce the gap between the current Boltzmann posterior and the variational policy, $\kl{\pit)}{\pi_{\omega_k}(a\vert s)}$ (see \cref{proof:policy_posterior}). This interpretation makes precise the intuition that how much we can improve our policy is determined by how similar $\hat{Q}_{\omega_k}(h)$ is to $Q^*(h)$, limiting policy improvement to the complete E-step:
$\pi_{\theta_{k+1}}(a\vert s)=\pi_{\omega_k}(a\vert s)$. We see that the common greedy policy improvement step, $\pi_{\theta_{k+1}}(a\vert s)=\delta(a\in\argmax_{a'}(\hat{Q}_{\omega_k}(a',s)))$ acts as an approximation to the Boltzmann form in \cref{eq:action_posterior}, replacing the softmax with a hard maximum.

If \cref{assumption:policy_represent} holds and any constraint induced by $\tw\boldsymbol{\cdot}$ does not prevent convergence to a complete E-step, the EM algorithm alternates between two convex optimisation schemes, and is guaranteed to converge to at least a local optimum of $\mathcal{L}(\omega,\theta)$ \citep{Wu83}. In reality, we cannot carry out complete E- and M-steps for complex domains, and our variational distributions are unlikely to satisfy \cref{assumption:policy_represent}. Under these conditions, we can resort to the empirically successful variational EM algorithm \citep{Jordan1999}, carrying out partial E- and M-steps instead, which we discuss further in \cref{sec:approx_gradients}.

\subsection{Advanced Actor-Critic Methods}
\label{sec:relationship_to_pg}
A family of actor-critic algorithms follows naturally from our framework: 1) we can use powerful inference techniques such as control variates \citep{qprop} or variance-reducing baselines by subtracting any function that does not depend on the action \citep{Schulman15a}, e.g., $V(s)$, from the action-value function, as this does not change our objective, 2) we can manipulate \cref{eq:e_step} to obtain variance-reducing gradient estimators such as EPG \citep{epg-journal}, FPG \citep{fpg}, and SVG0 \citep{Heess15}, and 3) we can take advantage of $d(s)$ being any general decorrelated distribution by using replay buffers \citep{dqn} or empirically successful asynchronous methods that combine several agents' individual gradient updates at once \citep{Mnih16}. As we discuss in \cref{sec:E-step_discussion}, the manipulation required to derive the estimators in 2) is not strictly justified in the classic policy gradient theorem \citep{Sutton00} and MERL formulation \citep{Haarnoja18}.

MPO is a state-of-the-art EM algorithm derived from the pseudo-likelihood objective \citep{Abdolmaleki18}. In its derivation, policy evaluation does not naturally arise from either of its EM steps and must be carried out separately. In addition, its E step is approximated, giving rise to the one step KL regularised update. As we demonstrate in \cref{sec:recovering_mpo}, under the probabilistic interpretation of our model, including a prior of the form $p_\phi(h)=\mathcal{U}(s)\pip$ in our ELBO and specifying a hyper-prior $p(\omega)$, the MPO objective with an adaptive regularisation constant can be recovered from \textsc{virel}: 
\begin{align}
\mathcal{L}^\textsc{mpo}(\omega,\theta,\phi)=\mathbb{E}_{s\sim d(s)}\left[\mathbb{E}_{a\sim \pit}\left[ \frac{\hQ}{\ew}\right]-\kl{\pit}{\pip }\right]+\log p(\omega).
\end{align}
We also show in \cref{sec:recovering_mpo} that applying the (variational) EM algorithm from \cref{sec:acem} yields the MPO updates with the missing policy evaluation step and without approximation in the E-step.

%% file: sections/experiments.tex
\vspace{-0.5cm}
\section{Experiments}
\vspace{-0.1cm}
\label{sec:experiments}
We evaluate our EM algorithm using the direct method approximation outlined in \cref{sec:approx_gradients} with $\tw$, ignoring constraints on $\Omega$. The aim of our evaluation is threefold: Firstly, as explained in \cref{sec:model}, algorithms using soft value functions cannot be recovered from \textsc{virel}. We therefore demonstrate that using hard value functions does not affect performance. Secondly, we provide evidence for our hypothesis stated in \cref{sec:comparing_virel_merlin} that using soft value functions may harm performance in higher dimensional tasks. Thirdly, we show that even under all practical approximations discussed, the algorithm derived in \cref{sec:acem} still outperforms advanced actor-critic methods.

We compare our methods to the state-of-the-art SAC\footnote{We use implementations provided by the authors \url{https://github.com/haarnoja/sac} for v1 and \url{https://github.com/vitchyr/rlkit} for v2.} and DDPG \citep{lillicrap2015continuous}  algorithms on MuJoCo tasks in OpenAI gym \citep{openai} and in rllab~\citep{rllab}. We use SAC as a baseline because \citet{Haarnoja18} show that it outperforms PPO \citep{Schulman2017ppo}, Soft $Q$-Learning \citep{Haarnoja17}, and TD3 \citep{fujimoto2018addressing}. We compare to DDPG \citep{lillicrap2015continuous}  because, like our methods, it can learn deterministic optimal policies. We consider two variants: In the first one, called \textit{virel}, we keep the scale of the entropy term in the gradient update for the variational policy constant $\alpha$; in the second, called \textit{beta}, we use an estimate $\he$ of $\ew$ to scale the corresponding term in \cref{eq:E-step_full}. We compute $\he$ using a buffer to draw a fixed number of samples $N_{\varepsilon}$ for the estimate. 
\begin{figure*}[!htb]
\vspace{-0.4cm}
	\begin{center}
		\minipage{0.33\textwidth}
		\includegraphics[width=\linewidth]{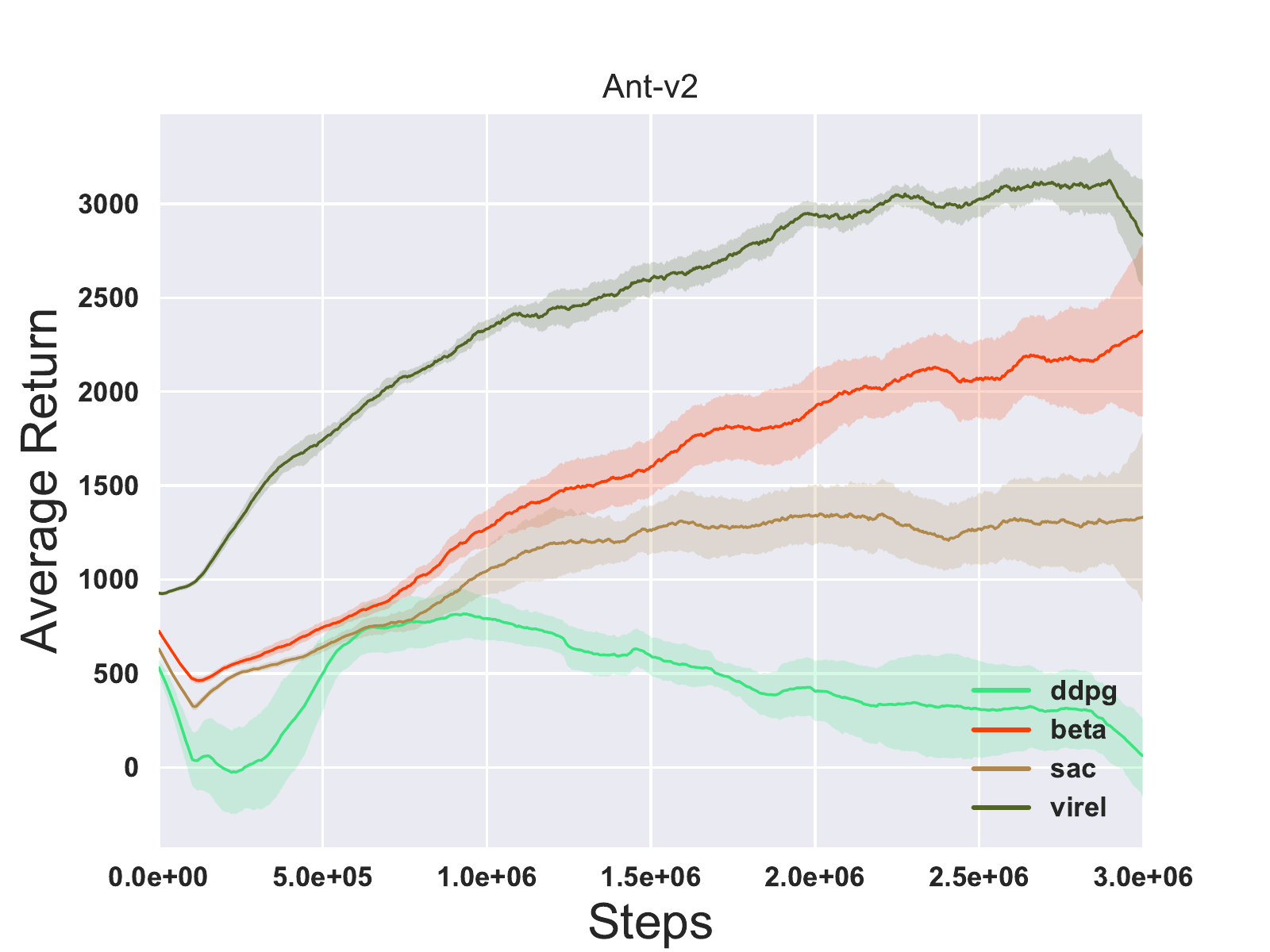}
		\label{fig:antv2}
		\endminipage
		\minipage{0.33\textwidth}
		\includegraphics[width=\linewidth]{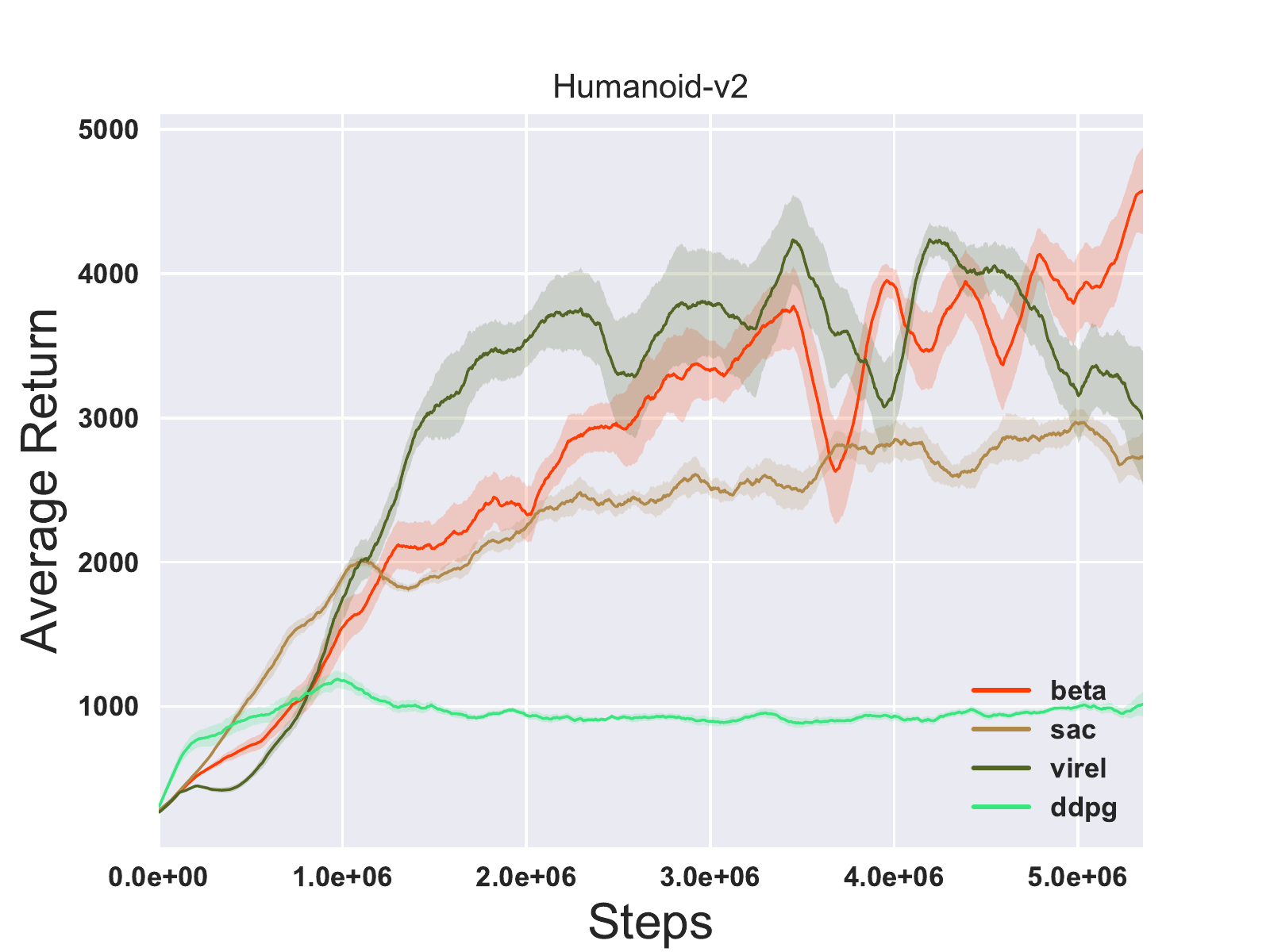}
		\label{fig:humanoidv2}
		\endminipage
		\minipage{0.33\textwidth}
		\includegraphics[width=\linewidth]{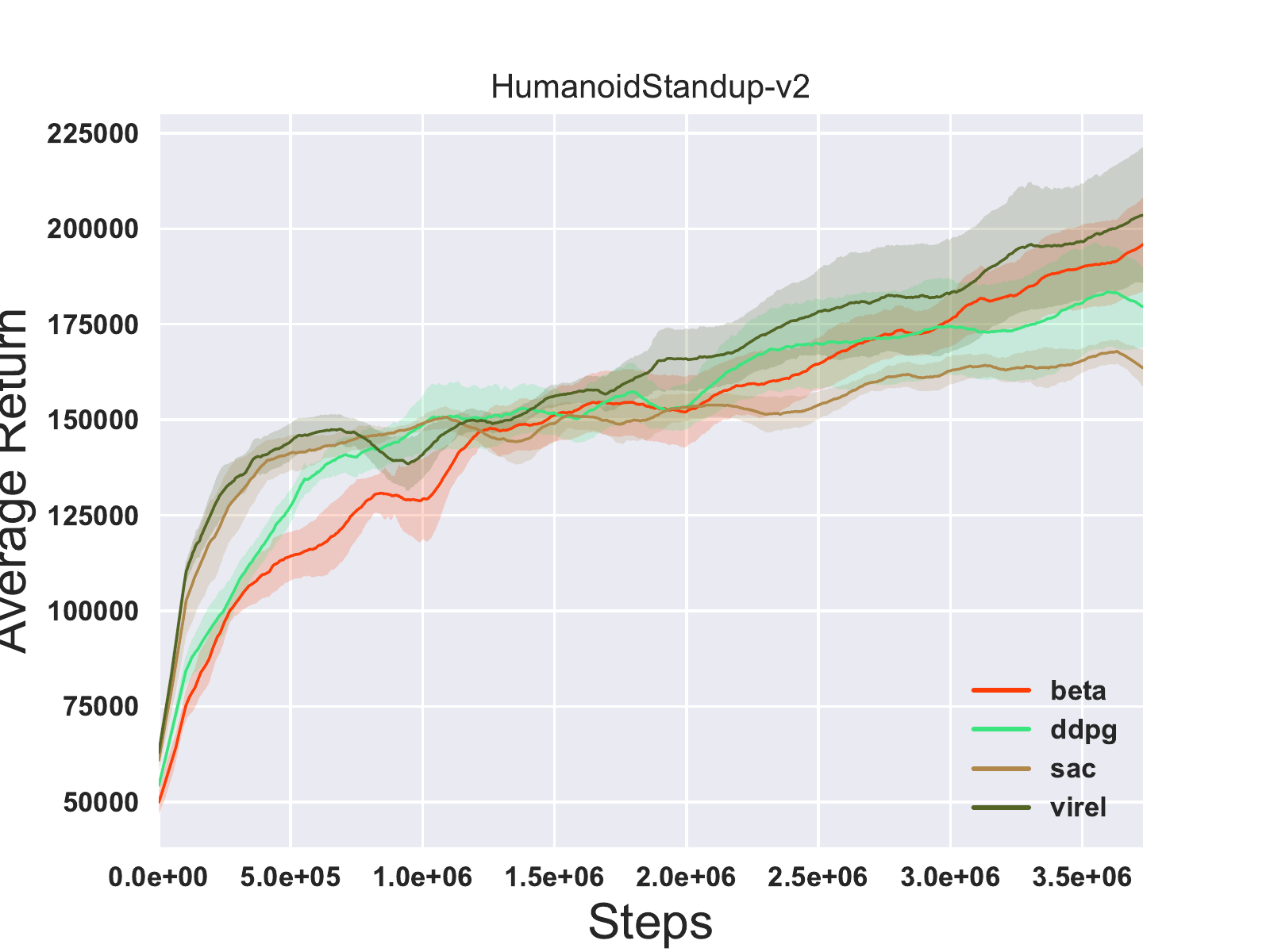}
		\label{fig:reacherv2}
		\endminipage\\
	\end{center}
\vspace{-5mm}
	\caption{Training curves on continuous control benchmarks gym-Mujoco-v2 : High-dimensional domains.}
		\vspace{-5mm}
	\label{experimentsv2}
\end{figure*}

To adjust for the relative magnitude of the first term in \cref{eq:E-step_full} with that of $\ew$ scaling the second term, we also multiply the estimate $\he$ by a scalar $\lambda\approx \frac{1-\gamma}{r_{avg}}$, where $r_{avg}$ is the average reward observed; $\lambda^{-1}$ roughly captures the order of magnitude of the first term and allows $\he$ to balance policy changes between exploration and exploitation. We found performance is poor and unstable without $\lambda$. To reduce variance, all algorithms use a value function network $V(\phi)$ as a baseline and a Gaussian policy, which enables the use of the reparametrisation trick. Pseudocode can be found in \cref{app:pseudocode}. All experiments use 5 random initialisations and parameter values are given in \cref{paremeter_values}.

\cref{experimentsv2} gives the training curves for the various algorithms on high-dimensional tasks for on gym-mujoco-v2. In particular, in Humanoid-v2 (action space dimensionality: 17, state space dimensionality: 376) and Ant-v2 (action space dimensionality: 8, state space dimensionality: 111), DDPG fails to learn any reasonable policy.  We believe that this is because the Ornstein-Uhlenbeck noise that DDPG uses for exploration is insufficiently adaptive in high dimensions. While SAC performs better, \textit{virel} and \textit{beta} still significantly outperform it. As hypothesised in \cref{sec:comparing_virel_merlin}, we believe that this performance advantage arises because the gap between optimal unregularised policies and optimal variational policies learnt under \textsc{merlin} is sensitive to temperature $c$. This effect is exacerbated in high dimensions where there may be large regions of the state-action space with sub-optimal reward. All algorithms learn optimal policies in simple domains, the training curves for which can be found  in \cref{experimentsv2_app} in \cref{sec:v2_experiments}. Thus, as the state-action dimensionality increases, algorithms derived from \textsc{virel} outperform SAC and DDPG.

\begin{figure*}[!htb]
\vspace{-0.3cm}
\begin{center}
\minipage{0.33\textwidth}
  \includegraphics[width=\linewidth]{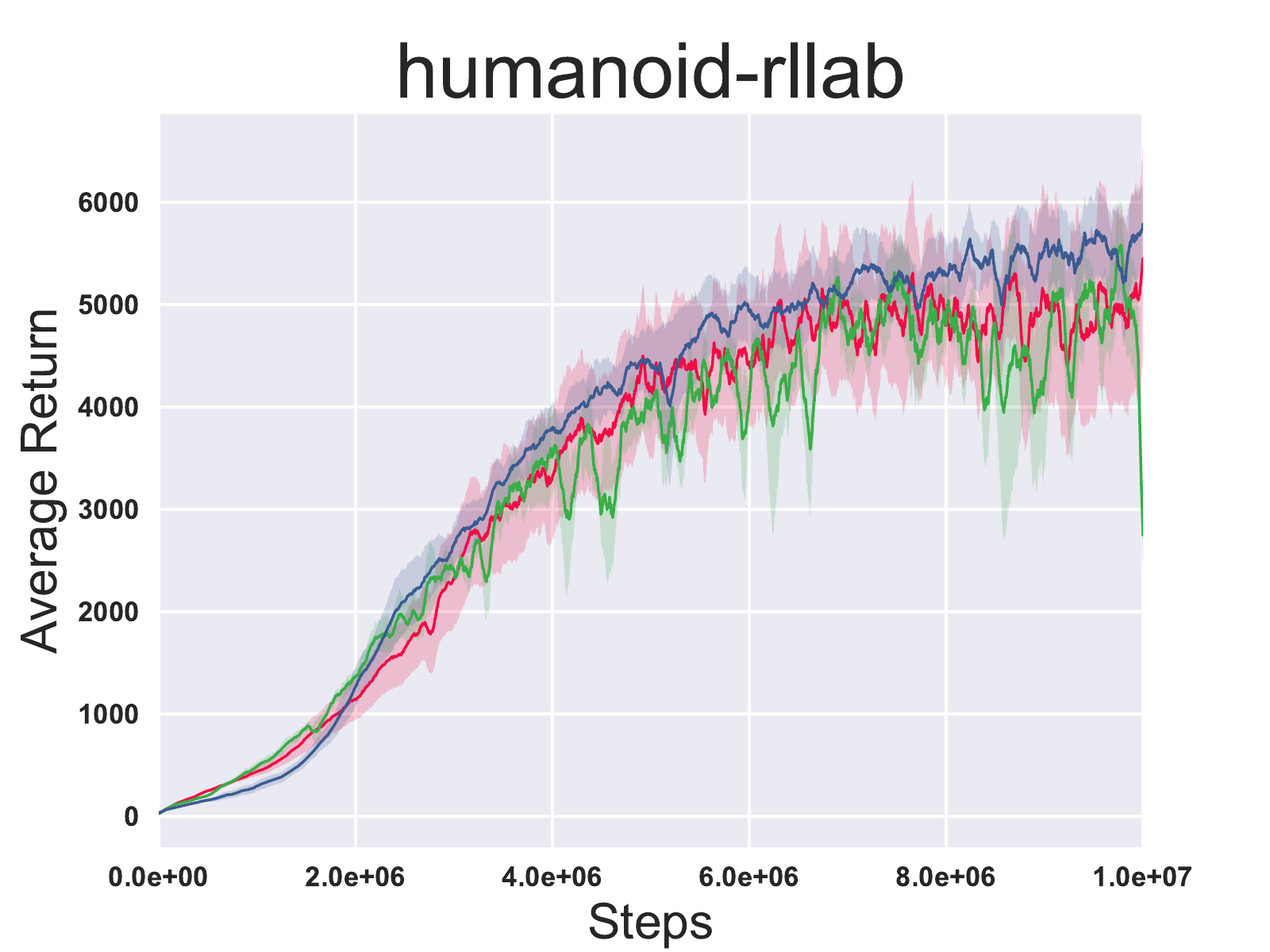}
  \label{fig:swimmer}
\endminipage
\minipage{0.33\textwidth}
  \includegraphics[width=\linewidth]{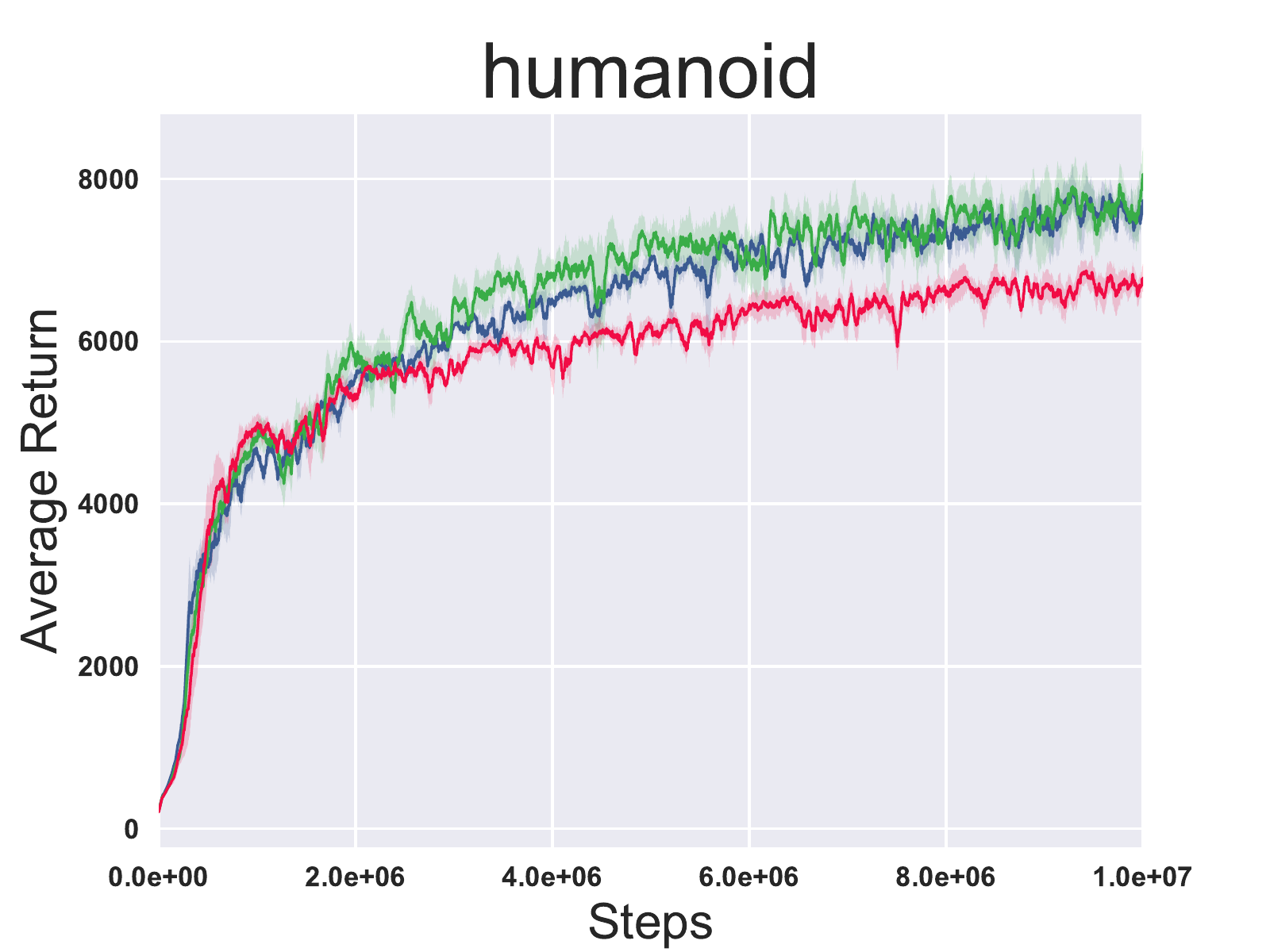}
  \label{fig:humanoid}
\endminipage
\minipage{0.33\textwidth}
  \includegraphics[width=\linewidth]{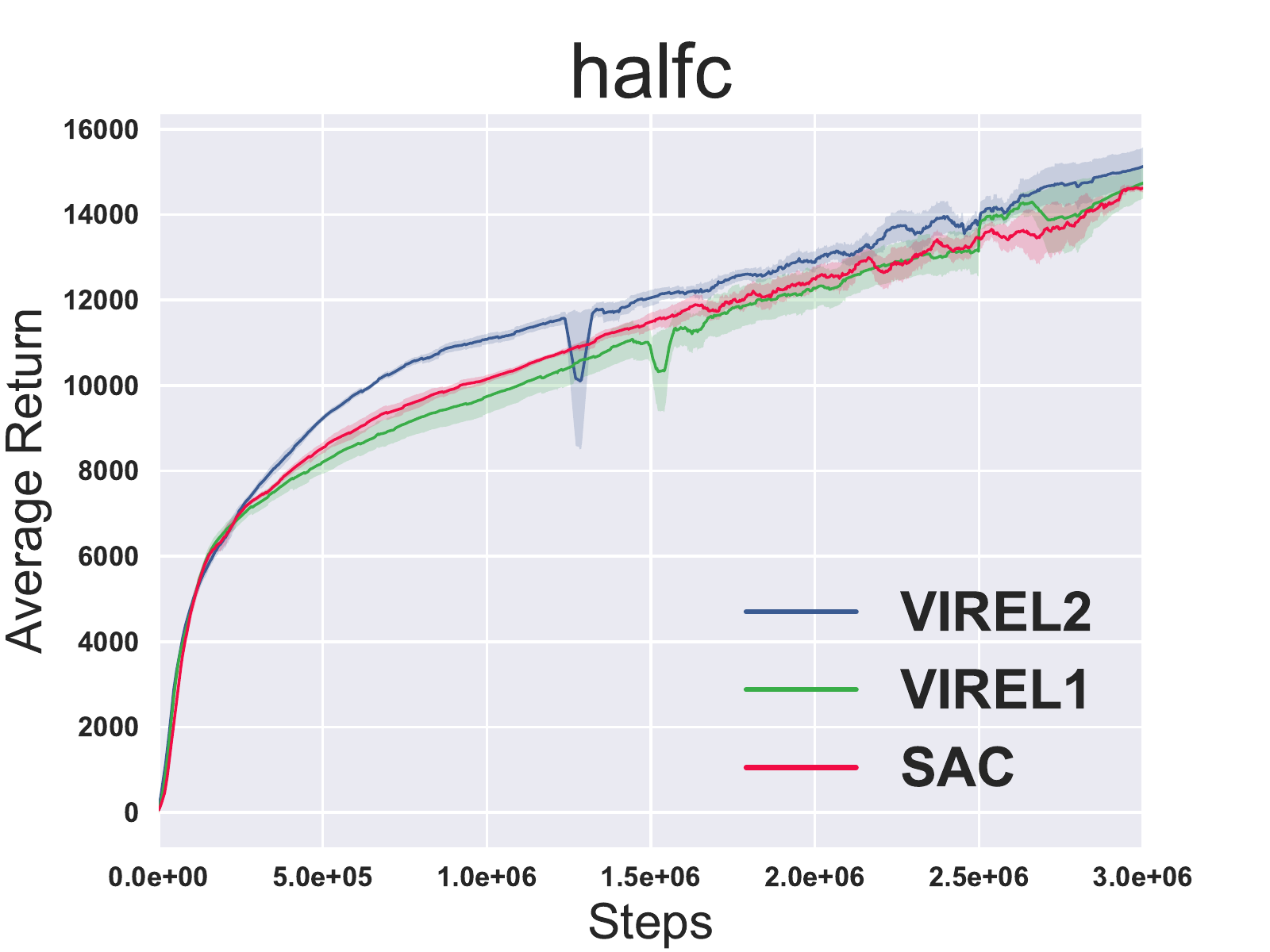}
  \label{fig:walker}
\endminipage
\end{center}
\vspace{-0.5cm}
\caption{Training curves on continuous control benchmarks gym-Mujoco-v1.}
\vspace{-0.5cm}
\label{experiments}
\end{figure*}

\citet{fujimoto2018addressing} and \citet{doubleq} note that using the minimum of two randomly initialised action-value functions helps mitigate the positive bias introduced by function approximation in policy gradient methods. Therefore, a variant of SAC uses two soft critics.  We compare this variant of SAC to two variants of \textit{virel}: \textit{virel1}, which uses two hard $Q$-functions and \textit{virel2}, which uses one hard and one soft $Q$-function. We scale the rewards so that the means of the $Q$-function estimates in \textit{virel2} are approximately aligned. \cref{experiments} shows the training curves on three gym-Mujoco-v1 domains, with additional plots shown in \cref{experimentsv1_app} in \cref{sec:v1_experiments}. Again, the results demonstrate that \textit{virel1} and \textit{virel2} perform on par with SAC in simple domains like Half-Cheetah and outperform it in challenging high-dimensional domains like humanoid-gym and -rllab (17 and 21 dimensional action spaces, 376 dimensional state space). 

%% file: sections/conclusion.tex
\vspace{-0.1cm}
\section{Conclusion and Future Work}
\vspace{-0.2cm}
\label{sec:conclusion}
This paper presented \textsc{virel}, a novel framework that recasts the reinforcement learning problem as an inference problem using function approximators. We provided strong theoretical justifications for this framework and compared two simple actor-critic algorithms that arise naturally from applying variational EM on the objective. Extensive empirical evaluation shows that our algorithms perform on par with current state-of-the-art methods on simple domains and substantially outperform them on challenging high dimensional domains. As immediate future work, our focus is to find better estimates of $\ew$ to provide a principled method for uncertainty based exploration; we expect it to help attain sample efficiency in conjunction with various methods like \citep{mahajan2017asymmetry,mahajan2017symmetry}. Another avenue of research would extend our framework to multi-agent settings, in which it can be used to tackle the sub-optimality induced by representational constraints used in MARL algorithms \citep{mahajan2019maven}.

%% file: sections/acknowledgements.tex
\section{Acknowledgements}
\vspace{-0.3 cm}
This project has received funding from the European Research Council (ERC) under the European Unions Horizon
2020 research and innovation programme (grant agreement
number 637713). The experiments were made possible by
a generous equipment grant from NVIDIA. Matthew Fellows is funded by the EPSRC. Anuj Mahajan is funded by Google DeepMind and the Drapers Scholarship. Tim G. J. Rudner is funded by the Rhodes Trust and the EPSRC. We would like to thank Yarin Gal and Piotr Miloś for helpful comments.

%% file: appendix/appendix.tex
\appendix
\section{A Brief Review of EM and Variational Inference}
\label{sec:inference_background}
\cref{fig:inference} shows the representation of a generative graphical model that produces observations $x$ from a distribution $x\sim \pw(x\vert h)$, has hidden variables $h$, and is parameterised by a set of parameters, $\omega$. In learning a model, we often seek the parameters that maximises the log-marginal-likelihood (LML), which can be found by marginalising the joint distribution $\pw(x,h)$ over hidden variables:
\begin{wrapfigure}{r}{0.24\textwidth}
\vspace{1cm}
    \begin{center}
    \includegraphics[scale=0.22]{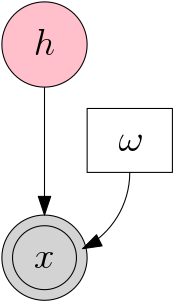}
    \end{center}
  \caption{Graphical model of inference problem.}
  \label{fig:inference}
  \vspace{-1cm}
\end{wrapfigure}
\begin{align}
\ell_\omega(x)\coloneqq\log\pw(x)=\log\left(\int_\mathcal{H} \pw(x,h)dh\right). \label{eq:marginal_likelihood}
\end{align}
In many cases, we also need to infer the corresponding posterior,
\begin{align}
 \pw(h\vert x)=\frac{\pw(x,h)}{\int_\mathcal{H} \pw(x,h)dh}.
\end{align}

Evaluating the marginal likelihood in \cref{eq:marginal_likelihood} and obtain the corresponding posterior, however, is intractable for most distributions. To compute the marginal likelihood and $\omega^*$, we can use the EM algorithm \citep{Dempster77maximumlikelihood} and variational inference (VI). We review these two methods now.

For any valid probability distribution $q(h)$ with support over $h$ we can rewrite the LML as a difference of two divergences \citep{Jordan1999},
\begin{align}
\label{eq:likelihood_decomposition}
\ell_\omega(x)=&\int_\mathcal{H} q(h)\log\left(\frac{\pw(x,h)}{q(h)}\right)dh -\int_\mathcal{H} q(h)\log\left(\frac{\pw(h\vert x)}{q(h)}\right)dh,\\
=& \mathcal{L}(\omega,q(h))+\kl{q(h)}{\pw(h\vert x)},\label{eq:lml_kl_elbo}
\end{align}
where $\mathcal{L}(\omega,q(h))\coloneqq\int_\mathcal{H} q(h)\log\left(\frac{\pw(x,h)}{q(h)}\right)dh$ is known as the evidence lower bound (ELBO).
Intuitively, as $\kl{q(h)}{\pw(h\vert x)}\ge 0$, it follows that $\ell_\omega(x)\ge\elbo{q(h);\omega}$, hence $\ell_\omega(x)\ge\elbo{q(h);\omega}$ is a lower bound for the LML. The derivation of this bound can also be viewed as applying Jensen's inequality directly to \cref{eq:marginal_likelihood} \citep{vi_review}. Note that when the ELBO and marginal likelihood are identical, the resulting KL divergence between the function $q(h)$ and the posterior $p(h\vert x)$ is zero, implying that $q(h)=\pw(h\vert x)$.

Maximising the LML now reduces to maximising the ELBO, which can be achieved iteratively using EM \citep{Dempster77maximumlikelihood, Wu83}; an expectation step (E-step) finds the posterior for the current set of model parameters and then a maximisation step (M-step) maximises the ELBO with respect to $\omega$ while keeping $q(h)$ fixed as the posterior from the E-step.  

As finding the exact posterior in the E-step is still typically intractable, we resort to variational inference (VI), a powerful tool for approximating the posterior using a parametrised variational distribution $\q$ \citep{Jordan1999,Beal03}. VI aims to reduce the KL divergence between the true posterior and the variational distribution, $\kl{\q}{\pw(h\vert x)}$. Typically VI never brings this divergence to zero but nonetheless yields useful posterior approximations. As minimising $\kl{\q}{\pw(h\vert x)}$ is equivalent to maximising the ELBO for the variational distribution (see \cref{eq:lml_kl} from \cref{app:proof_kl}), the variational E-step amounts to maximising the ELBO with respect to $\theta$ while keeping $\omega$ constant. The variational EM algorithm can be summarised as:
\begin{align}
\textrm{Variational E-Step: } &\theta_{k+1}\leftarrow\argmax_\theta\mathcal{L}(\omega_k,\theta),\\
\textrm{Variational M-Step: } &\omega_{k+1}\leftarrow\argmax_\omega\mathcal{L}(\omega,\theta_{k+1}).
\end{align}
\section{A Probabilistic Interpretation of \textsc{virel}} 
\label{sec:probablistic_interpretation}
We now motivate our inference procedure and Boltzmann distribution $\piw$ from a probabilistic perspective, demonstrating that $\piw$ can be interpreted as an action-posterior that characterises the uncertainty our model has in the optimality of $\hQ$. Moreover, maximising $\lp$ for $\theta$ is equivalent to carrying our variational inference on the graphical model in \cref{fig:virel_graph} for any $\ew>0$.

\subsection{Model Specification}

Like previous work, we introduce a binary variable $\mathcal{O}\in\{0,1\}$ in order to define a formal graphical model for our inference problem when $\ew>0$. The likelihood of $\mathcal{O}$ therefore takes the form of a Bernoulli distribution:
 \begin{align}
&\pw(\mathcal{O}\vert h) =y_\omega(h)^\mathcal{O}(1-y_\omega(h))^{(1-\mathcal{O})}, \label{eq:virel_likelihood}
\end{align}
where 
\begin{align}
    y_\omega(h)\coloneqq \exp\left(\frac{\hQ-\max _{a'}\hat{Q}_\omega(a',s)}{\ew}\right).
\end{align}
In most existing frameworks, $\mathcal{O}=1$ is understood to be the event that the agent is acting optimally \citep{virl_review, Toussaint09b}. As we are using function approximators in \textsc{virel}, $\mathcal{O}=1$ can be interpreted as the event that the agent is behaving optimally under $\hQ$. Exploring the semantics of $\mathcal{O}$ further, consider the likelihood when $\mathcal{O}=1$:
\begin{align}
    \pw(\mathcal{O}=1\vert h)=\exp\left(\frac{\hQ-\max _{a'}\hat{Q}_\omega(a',s)}{\ew}\right), 
\end{align}
\begin{wrapfigure}{r}{0.24\textwidth}
\vspace{-0.7cm}

    \begin{center}
    \includegraphics[scale=0.25]{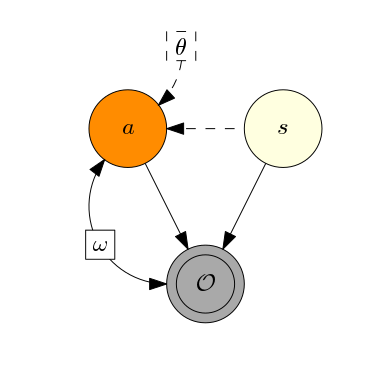}
    \end{center}
    \vspace{-0.2cm}
  \caption{Graphical model for \textsc{virel} (variational approximation dashed)}
  \label{fig:virel_graph}
  \vspace{-0.6cm}
\end{wrapfigure}
Observe that $0\le\pw(\mathcal{O}=1\vert \cdot)\le1$ $\forall\ \omega\in\Omega\ s.t.\ \ew>0$. For any state $s$ and any action $a^*$ such that $\pw(\mathcal{O}=1\vert s,a^*)=1$, such an action must be optimal under $\hQ$ in the sense that it is the greedy action $a^*\in\argmax_{a}\hQ$. If we find $\pw(\mathcal{O}=1\vert h)=1\ \forall\  h\in\mathcal{H}$, then all observed state-action pairs have been generated from a greedy policy $\pi(a\vert s)=\delta(a\in\argmax_{a'}\hat{Q}_\omega(a'\vert s))$. From \cref{proof:forward}, the closer the residual error $\ew$ is to zero, the closer $\hQ$ becomes to representing an optimal action-value function. When $\ew\approx0$, any $a$ observed such that $\pw(\mathcal{O}=1\vert a,\cdot)=1$ will be very nearly an action sampled from an optimal policy, that is  $a\sim\pi(a\vert \cdot)\approx\delta(a\in\argmax_{a'}Q^*(a'\vert \cdot))$. We caution readers that in the limit $\ew\rightarrow 0$, our likelihood is not well-defined for any $a\in\argmax _{a'}\hat{Q}_\omega(a',s)$. Without loss of generality, we condition on optimality for the rest of this section, writing $\mathcal{O}$ in place of $\mathcal{O}=1$. Defining the function $y_\omega(s)\coloneqq\exp\left(-\frac{\max _{a'}\hat{Q}_\omega(a',s)}{\ew}\right)$, our likelihood takes the convenient form:
\begin{align}
    \pw(\mathcal{O}\vert h)=\exp\left(\frac{\hQ}{\ew}\right)y_\omega(s), 
\end{align}

Defining the prior distribution as the uniform distribution $p(h)=\mathcal{U}(h)$ completes our model, the graph for which is shown in \cref{fig:virel_graph}. Using Bayes' rule, we find our posterior distribution is: 
\begin{align}
    \pw(h\vert\mathcal{O})&=\frac{\pw(\mathcal{O}\vert h)p(h)}{p_\omega(\mathcal{O})},\\
    &=\frac{\pw(\mathcal{O}\vert h)p(h)}{\int_\mathcal{H}\pw(\mathcal{O}\vert h)p(h)dh},\\
    &=\frac{\exp\left(\frac{\hQ}{\ew}\right)y_\omega(s)}{\int_\mathcal{H}\exp\left(\frac{\hQ}{\ew}\right)y_\omega(s)dh}.\label{eq:posterior}
\end{align}
We can also derive our action-posterior, $\pw(a\vert s,\mathcal{O})$, which we will find to be equivalent to the Boltzmann policy from \cref{eq:action_posterior}. Using Bayes' rule, it follows:
\begin{align}
    \pw(a\vert s,\mathcal{O})&=\frac{\pw(h\vert \mathcal{O})}{\pw(s\vert \mathcal{O})}.
\end{align}
Now, we find $\pw(s\vert \mathcal{O})$ by marginalising our posterior over actions. Substituting $\pw(s\vert \mathcal{O})=\int\pw(h\vert \mathcal{O})da$ yields : 
\begin{align}
    \pw(a\vert s,\mathcal{O})&=\frac{\pw(h\vert \mathcal{O})}{\int_\mathcal{A}\pw(h\vert \mathcal{O})da}.
\end{align}
Substituting for our posterior from \cref{eq:posterior}, we obtain:
\begin{align}
    \pw(a\vert s,\mathcal{O})&=\frac{\exp\left(\frac{\hQ}{\ew}\right)y_\omega(s)}{\int_\mathcal{A}\exp\left(\frac{\hQ}{\ew}\right)y_\omega(s)da}\cdot\frac{\int_\mathcal{H}\exp\left(\frac{\hQ}{\ew}\right)y_\omega(s)dh}{\int_\mathcal{H}\exp\left(\frac{\hQ}{\ew}\right)y_\omega(s)dh},\\
    &=\frac{\exp\left(\frac{\hQ}{\ew}\right)y_\omega(s)}{\left(\int_\mathcal{A}\exp\left(\frac{\hQ}{\ew}\right)da\right) y_\omega(s)},\\
    &=\frac{\exp\left(\frac{\hQ}{\ew}\right)}{\int_\mathcal{A}\exp\left(\frac{\hQ}{\ew}\right)da},\\
    &=\piw,
\end{align}
proving that our action-posterior is exactly the Boltzmann policy introduced in \cref{sec:model}. From a Bayesian perspective, the action-posterior $\pw(a\vert s,\mathcal{O})$ characterises the uncertainty we have in deducing the optimal action for a given state $s$ under $\hQ$; whenever $\ew\approx 0$ and hence $\hQ\approx Q^*(h)$, the uncertainty will be very small as $\pw(a\vert s,\mathcal{O})$ will have near-zero variance, approximating a Dirac-delta distribution. Our model is therefore highly confident that the maximum-a-posteriori (MAP) action $a\in\argmax_{a'}\hat{Q}_\omega(a',s)$ is an optimal action, with all of the probability mass being close to this point. In light of this, we can interpret the greedy policy $\piw=\delta(a\in\argmax_{a'}\hat{Q}_\omega(a',s))$ as one that always selecting the MAP action across all states. 

As our model incorporates the uncertainty in the optimality of $\hQ$ into the variance of $\piw$, we can benefit directly by sampling trajectories from $\piw$ which drives exploration to gather data that is beneficial to reducing the residual error $\ew$. Unfortunately, calculating the normalisation constant $\int_\mathcal{A}\exp\left(\frac{\hQ}{\ew}\right)da$ is intractable for most function approximators and MDPs of interest. As such, we resort to variational inference, a powerful technique to infer an approximation to a posterior distribution from a tractable family of variational distributions  \citep{Jordan1999,Beal03,vi_review}. As before $\pit$ is known as the variational policy, is parametrised by $\theta\in\Theta$ and with the same support as $\piw$. Like in \cref{sec:model}, we define a variational distribution as $\qt\coloneqq d(s)\pit$, where $d(s)$ is an arbitrary sampling distribution with support over $\mathcal{S}$. We fix $d(s)$, as in our model-free paradigm we do not learn the state transition dynamics and only seek to infer the action-posterior. 

The goal of variational inference is to find $\qt$ closest in KL-divergence to $\pw(h\vert \mathcal{O})$, giving an objective:
\begin{align}
    \theta^*\in\argmin_{\theta}\kl{\qt}{\pw(h\vert \mathcal{O})}.
\end{align}
This objective still requires the intractable computation of $\int\exp\left(\frac{\hQ}{\ew}\right)y_\omega(s)dh$. Using \cref{eq:lml_kl_elbo}, we can overcome this by writing the KL divergence in terms of the ELBO:
\begin{gather}
    \kl{\qt}{\pw(h\vert \mathcal{O})}=\ell_\omega-\mathcal{L}_\omega(\theta)
,\\
    \mathrm{where}\quad\ell_\omega\coloneqq\log\int_\mathcal{H}\exp\left(\frac{\hQ}{\ew}\right)y_\omega(s)dh,\quad \mathcal{L}_\omega(\theta)\coloneqq \mathbb{E}_{h\sim \qt}\left[\log\left(\frac{\exp\left(\frac{\hQ}{\ew}\right)y_\omega(s)}{\qt}\right)\right].
\end{gather}
We see that minimising the KL-divergence for $\theta$ is equivalent to maximising the ELBO for $\theta$, which is tractable. This affords a new objective:
\begin{align}
    \theta^*\in\argmax_{\theta}\mathcal{L}_\omega(\theta).
\end{align}
Expanding the ELBO yields:
\begin{align}
\mathcal{L}_\omega(\theta)&= \mathbb{E}_{h\sim \qt}\left[\log\left(\frac{\exp\left(\frac{\hQ}{\ew}\right)y_\omega(s)}{\qt}\right)\right],\\
&= \mathbb{E}_{s\sim d(s)}\left[\mathbb{E}_{a\sim \pit}\left[\log\left(\frac{\exp\left(\frac{\hQ}{\ew}\right)y_\omega(s)}{\qt}\right)\right]\right],\\
&=  \mathbb{E}_{s\sim d(s)}\left[\mathbb{E}_{a\sim \pit}\left[\frac{\hQ}{\ew}\right]+\mathbb{E}_{a\sim \pit}\left[\log y_\omega(s)\right]-\mathbb{E}_{a\sim \pit}\left[\log(\pit d(s))\right]\right],\\
&=  \mathbb{E}_{s\sim d(s)}\left[\mathbb{E}_{a\sim \pit}\left[\frac{\hQ}{\ew}\right]+\log y_\omega(s)-\log d(s)-\mathbb{E}_{a\sim \pit}\left[\log\pit\right]\right],\\
&=  \mathbb{E}_{s\sim d(s)}\left[\mathbb{E}_{a\sim \pit}\left[\frac{\hQ}{\ew}\right]-\mathbb{E}_{a\sim \pit}\left[\log\pit\right]\right]+\mathbb{E}_{s\sim d(s)}\left[\log\left(\frac{y_\omega(s)}{d(s)}\right)\right],\\
&=  \mathbb{E}_{s\sim d(s)}\left[\mathbb{E}_{a\sim \pit}\left[\frac{\hQ}{\ew}\right]+\mathscr{H}(\pit)\right]+\mathbb{E}_{s\sim d(s)}\left[\log\left(\frac{y_\omega(s)}{d(s)}\right)\right].
\end{align}
As the final term $\mathbb{E}_{s\sim d(s)}\left[\log\left(\frac{y_\omega(s)}{d(s)}\right)\right]$ has no dependency on $\theta$, we can neglect it from our objective, recovering the \textsc{virel}  objective from \cref{eq:elborl}:
\begin{align}
    \mathcal{L}_\omega(\theta)&=\mathbb{E}_{s\sim d(s)}\left[\mathbb{E}_{a\sim \pit}\left[\frac{\hQ}{\ew}\right]+\mathscr{H}(\pit)\right].
\end{align}
Finally, \cref{proof:policy_posterior} guarantees that minimising $\mathcal{L}_\omega(\theta)$ always minimises the expected KL divergence between $\piw$ and $\pit$, allowing us to learn a variational approximation for the action-posterior.

\section{A Discussion of the Target Set $\mathbb{T}$}
\label{sec:discussion_T}
We now prove that the Bellman operator for the Boltzmann policy, $\mathcal{T}^{\pi_\omega}\boldsymbol{\cdot}\coloneqq r(h)+\gamma \mathbb{E}_{h'\sim p(s'\vert h)\pi_\omega(a'\vert s')}\left[\boldsymbol{\cdot}\right]$, is a member of $\mathbb{T}$. Taking the limit $\ew\rightarrow0$ of $\mathcal{T}^{\pi_\omega}\hQ$, we find:
\begin{align}
    \lim_{\ew\rightarrow0}\mathcal{T}^{\pi_\omega}\hQ&=r(h)+\lim_{\ew\rightarrow0}\gamma \mathbb{E}_{h'\sim p(s'\vert h)\pi_\omega(a'\vert s')}\left[\hQd\right].
\end{align}
 From \cref{proof:forward}, evalutating $\lim_{\ew\rightarrow0}\gamma \mathbb{E}_{h'\sim p(s'\vert h)\pi_\omega(a'\vert s')}\left[\boldsymbol{\cdot}\right]$ recovers a Dirac-delta distribution:
\begin{align}
    \lim_{\ew\rightarrow0}\mathcal{T}^{\pi_\omega}\hQ&=r(h)+\gamma \mathbb{E}_{h'\sim p(s'\vert h)\delta(a'=\argmax_{a}\hat{Q}_\omega(a,s))}\left[\hQd\right],\\
    &=r(h)+\gamma \mathbb{E}_{h'\sim p(s'\vert h)}\left[\max_{a'}(\hQd)\right],\\
    &=\mathcal{T}^*\hQ.
\end{align}
which is sufficient to demonstrate membership of $\mathbb{T}$.

Observe that using $\mathcal{T}^{\pi_\omega}\boldsymbol{\cdot}$ implies $\hQ$ cannot represent the true $Q$-function of any $\piw$ except for the optimal Q-function. To see this, imagine there exists some $\ew>0$ such that $Q^{\pi_\omega}(\cdot)=\hat{Q}(\cdot)$. Under these conditions, it holds that $\mathcal{T}^{\pi_\omega}\hat{Q}(\cdot)=\hat{Q}(\cdot)\implies\ew=0$, which is a contradiction. More generally, as $\piw$ is defined in terms of $\ew$, which itself depends on $\piw$ from the definition of $\mathcal{T}^{\pi_\omega}\boldsymbol{\cdot}$, any $\omega$  satisfying this recursive definition forms a constrained set $\Omega^c\subseteq\Omega$. Crucially, we show in \cref{proof:forward} that there always exists some $\omega^*\in\Omega^c$ such that $\hat{Q}_{\omega^*}$ can represent the action-value function for an optimal policy. Note that there may exist other policies that are not Boltzmann distributions such that $\hQ=Q^\pi(h)$ for some $\omega\in\Omega^c$. We discuss operators that don't constrain $\Omega$ in \cref{sec:relaxation_constraints}. 

Finally, we can approximate $\mathcal{T}^{\pi_\omega}$ using any TD target sampled from $\piw$ (see \citet{Sutton17} for an overview of TD methods). Likewise, the optimum Bellman operator $\mathcal{T}^*\boldsymbol{\cdot}=r(h)+\gamma \mathbb{E}_{h'\sim p(s'\vert h)}\left[\max_{a'}(\boldsymbol{\cdot})\right]$ is by definition a member of $\mathbb{T}$ and can be approximated using the Q-learning target \citep{qlearning}.

\section{Proofs for \cref{sec:rl_as_inference}}
\label{sec:model_proofs}

\addtocounter{lemma}{-1}
\addtocounter{theorem}{-3}

\subsection{Derivation of Lower Bound in terms of KL Divergence}
\label{sec:derivation_elbo}
We need to show that 
\begin{gather}
\mathcal{L}(\omega,\theta)=\ell(\omega)-\kl{q_\theta(h)}{\p}-\mathscr{H}(d(s)),\label{eq:objective_ii_app}\\
    \mathrm{where}\quad\mathcal{\ell}(\omega)\coloneqq\log\int_\mathcal{H}\exp\left(\frac{\hQ}{\ew}\right)dh,\quad \p\coloneqq \frac{\exp\left(\frac{\hQ}{\ew}\right)}{\int_\mathcal{H}\exp\left(\frac{\hQ}{\ew}\right)dh}.
\end{gather}
Starting with the LHS of \cref{eq:objective_ii_app}, and recalling the definition of $\lp$ from \cref{eq:elborl}, we have:
\begin{align}
    \lp=\mathbb{E}_{s\sim d(s)}\left[\mathbb{E}_{a\sim \pit} \left[\frac{\hQ}{\ew}\right]+\mathscr{H}(\pit)\right].
\end{align}
Expanding the definition of differential entropy, $\mathscr{H}(\pit)$:
\begin{align}
    \lp&=\mathbb{E}_{s\sim d(s)}\left[\mathbb{E}_{a\sim \pit} \left[\frac{\hQ}{\ew}\right]-\mathbb{E}_{a\sim \pit}\left[\log\pit\right]\right],\\
    &=\mathbb{E}_{s\sim d(s)}\left[\mathbb{E}_{a\sim \pit} \left[\frac{\hQ}{\ew}\right]-\mathbb{E}_{a\sim \pit}\left[\log\left(\frac{\pit d(s)}{\p}\cdot\frac{\p}{d(s)}\right)\right]\right],\\
    &=\mathbb{E}_{s\sim d(s)}\Bigg[\mathbb{E}_{a\sim \pit} \left[\frac{\hQ}{\ew}\right]-\mathbb{E}_{a\sim \pit}\left[\log\left(\frac{\qt}{\p}\right)\right]\\
    &\quad-\mathbb{E}_{a\sim \pit}\left[\log\p\right]+\log(d(s))\Bigg],\\
     &=\mathbb{E}_{h\sim \qt}\left[\frac{\hQ}{\ew}\right]-\kl{\qt}{\p}-\mathscr{H}(d(s))-\mathbb{E}_{h\sim \qt}\left[\log\p\right].
\end{align}
Substituting for the definition of $\p$ in the final term yields our desired result:
\begin{align}
    \lp&=\mathbb{E}_{h\sim \qt}\left[\frac{\hQ}{\ew}\right]-\kl{\pit}{\piw}-\mathscr{H}(d(s))\\
    &\quad-\mathbb{E}_{a\sim \pit}\left[\log\left(\frac{\exp\left(\frac{\hQ}{\ew}\right)}{\int_{\mathcal{H}}\exp\left(\frac{\hQ}{\ew}\right)dh}\right)\right]\Bigg],\\
    &=\mathbb{E}_{h\sim \qt}\left[\frac{\hQ}{\ew}\right]-\kl{\pit}{\piw}-\mathscr{H}(d(s))\\
    &\quad-\mathbb{E}_{h\sim \qt}\left[\frac{\hQ}{\ew}\right]+\log\int_{\mathcal{H}}\exp\left(\frac{\hQ}{\ew}\right)dh\Bigg],\\
    &=\ell(\omega)-\kl{q_\theta(h)}{\p}-\mathscr{H}(d(s)).
\end{align}
\subsection{Convergence of Boltzmann Distribution to Dirac-Delta}
\label{sec:convergence_boltzmann}

\begin{theorem}[Convergence of Boltzmann Distribution to Dirac Delta]\label{proof:app_convergence_boltzmann}
Let $p_\varepsilon:\mathcal{X}\rightarrow[0,1]$ be a Boltzmann distribution with temperature $\varepsilon\in\mathbb{R}_{\ge0}$
\begin{align}
    p_\varepsilon(x)=\frac{\exp\left(\frac{f(x)}{\varepsilon}\right)}{\int_\mathcal{X}\exp\left(\frac{f(x)}{\varepsilon}\right)dx},
\end{align}
where $f:\mathcal{X}\rightarrow\mathcal{Y}$ is a function with a unique maximum $f(x^*)=\sup_x f$, a compact domain $\mathcal{X}$ and bounded range $\mathcal{Y}$. Let $f$ be locally $\mathbb{C}^2$ smooth about $x^*$, that is $\exists\ \Delta>0\ s.t. f(x)\in\mathbb{C}^2\ \forall\ x\in\{x\vert \lVert x-x^*\rVert<\Delta$ \}. In the limit $\varepsilon\rightarrow 0$, $p_\varepsilon(x)\rightarrow \delta(x^*)$, that is:
\begin{align}
    \lim_{\varepsilon\rightarrow0}\int_{\mathcal{X}} \varphi(x)p_\varepsilon(x)dx=\varphi(x^*),\label{eq:delta_condition}
\end{align}
for any smooth test function $\varphi\in\mathbb{C}_0^\infty(\mathcal{X})$.
\begin{proof}
Firstly, we define the auxiliary function to be     
\begin{align}
    g(x)\coloneqq f(x)-f(x^*).
\end{align}

Note, $g(x)\le 0$ with equality at $g(x^*)=0$. Substituting  $f(x)=g(x)+f(x^*)$ into $p_\varepsilon(x)$:
\begin{align}
p_\varepsilon(x)=&\frac{\exp\left(\frac{g(x)+f(x^*)}{\varepsilon}\right)}{\int_\mathcal{X}\exp\left(\frac{g(x)+f(x^*)}{\varepsilon}\right)dx},\\
=&\frac{\exp\left(\frac{g(x)}{\varepsilon}\right)\exp\left(\frac{f(x^*)}{\varepsilon}\right)}{\int_\mathcal{X}\exp\left(\frac{g(x)}{\varepsilon}\right)\exp\left(\frac{f(x^*)}{\varepsilon}\right)dx},\\
=&\frac{\exp\left(\frac{g(x)}{\varepsilon}\right)}{\int_\mathcal{X}\exp\left(\frac{g(x)}{\varepsilon}\right)dx}.\label{eq:p_as_g}
\end{align}
Now, substituting \cref{eq:p_as_g} into the limit in \cref{eq:delta_condition} yields:
\begin{align}
    \lim_{\varepsilon\rightarrow0}\int_{\mathcal{X}} \varphi(x)p_\varepsilon(x)dx=\lim_{\varepsilon\rightarrow0}\left(\int_{\mathcal{X}} \varphi(x)\frac{\exp\left(\frac{g(x)}{\varepsilon}\right)}{\int_\mathcal{X}\exp\left(\frac{g(x)}{\varepsilon}\right)dx}dx\right).\label{eq:limit_g}
\end{align}
Using the substitution $u\coloneqq \frac{(x^*-x)}{\sqrt{\varepsilon}}$ to transform the integrals in \cref{eq:limit_g}, we obtain
\begin{align}
    \lim_{\varepsilon\rightarrow0}\int_{\mathcal{X}} \varphi(x)p_\varepsilon(x)dx&=\lim_{\varepsilon\rightarrow0}\left(\int_{\mathcal{U}} \varphi(x^*-\sqrt{\varepsilon}u)\frac{\exp\left(\frac{g(x^*-\sqrt{\varepsilon}u)}{\varepsilon}\right)}{\int_\mathcal{U}\exp\left(\frac{g(x^*-\sqrt{\varepsilon}u)}{\varepsilon}\right)\sqrt{\varepsilon}du}\sqrt{\varepsilon}du\right),\\
    &=\lim_{\varepsilon\rightarrow0}\left(\frac{\int_{\mathcal{U}} \varphi(x^*-\sqrt{\varepsilon}u)\exp\left(\frac{g(x^*-\sqrt{\varepsilon}u)}{\varepsilon}\right)du}{\int_\mathcal{U}\exp\left(\frac{g(x^*-\sqrt{\varepsilon}u)}{\varepsilon}\right)du}\right).\label{eq:limit_u}
\end{align}
We now find $\lim_{\varepsilon\rightarrow0}\left(\frac{g(x^*-\sqrt{\varepsilon}u)}{\varepsilon}\right)$. Denoting the partial derivative $\partial_{\sqrt{\varepsilon}}\coloneqq \frac{\partial}{\partial\sqrt{\varepsilon}}$ and using L'H\^{o}pital's rule to the second derivative with respect to $\sqrt{\epsilon}$, we find the limit as:
\begin{align}
    \lim_{\varepsilon\rightarrow0}\left(\frac{g(x^*-\sqrt{\varepsilon}u)}{\varepsilon}\right)&=\lim_{\varepsilon\rightarrow0}\left(\frac{\partial_{\sqrt{\varepsilon}}g(x^*-\sqrt{\varepsilon}u)}{\partial_{\sqrt{\varepsilon}}\varepsilon}\right),\\
    &=\lim_{\varepsilon\rightarrow0}\left(\frac{\partial_{\sqrt{\varepsilon}}f(x^*-\sqrt{\varepsilon}u)}{\partial_{\sqrt{\varepsilon}}\varepsilon}\right),\\
    &=\lim_{\varepsilon\rightarrow0}\left(\frac{-u^\top\nabla f(x^*-\sqrt{\varepsilon}u)}{2\sqrt{\varepsilon}}\right),\\ &=\lim_{\varepsilon\rightarrow0}\left(\frac{-\partial_{\sqrt{\varepsilon}}\left(u^\top\nabla f(x^*-\sqrt{\varepsilon}u)\right)}{\partial_{\sqrt{\varepsilon}}(2\sqrt{\varepsilon})}\right),\\
    &=\lim_{\varepsilon\rightarrow0}\left(\frac{u^\top\nabla^2 f(x^*-\sqrt{\varepsilon}u)u}{2}\right),\\
    &=\frac{u^\top\nabla^2 f(x^*)u}{2}.
\end{align}
The integrand in the numerator in \cref{eq:limit_u} therefore converges pointwise to $\varphi(x^*)\exp\left(\frac{u^\top\nabla^2 f(x^*)u}{2}\right)$, that is 
\begin{align}
    \lim_{\varepsilon\rightarrow0}\left(\varphi(x^*-\sqrt{\varepsilon}u)\exp\left(\frac{g(x^*-\sqrt{\varepsilon}u)}{\varepsilon}\right)\right)=\varphi(x^*)\exp\left(\frac{u^\top\nabla^2 f(x^*)u}{2}\right),\label{eq:point_num}
\end{align}
and the integrand in the denominator converges pointwise to $\exp\left(\frac{u^\top\nabla^2 f(x^*)u}{2}\right)$, that is 
\begin{align}
    \lim_{\varepsilon\rightarrow0}\left(\exp\left(\frac{g(x^*-\sqrt{\varepsilon}u)}{\varepsilon}\right)\right)=\exp\left(\frac{u^\top\nabla^2 f(x^*)u}{2}\right).\label{eq:point_dom}
\end{align}
From the second order sufficient conditions for $f(x^*)$ to be a maximum, we have $u^\top\nabla^2 f(x^*)u \le 0$ $\forall\ u\in\mathcal{U}$ with equality only when $u=0$ \citep{Liberzon11}.  This implies that \cref{eq:point_num} and \cref{eq:point_dom} are both bounded functions. 

By definition, we have $g(x^*-\sqrt{\epsilon} u)\le0\ \forall\ u\in\mathcal{U}$, which implies that $|\exp\left(\frac{g(x^*-\sqrt{\varepsilon}u)}{\varepsilon}\right)|\le1$. Consequently, the integrand in the numerator of \cref{eq:limit_u} is dominated by $\lVert \varphi(\cdot)\rVert_\infty$, that is 
\begin{align}
    \left\lvert\varphi(x^*-\sqrt{\varepsilon}u)\exp\left(\frac{g(x^*-\sqrt{\varepsilon}u)}{\varepsilon}\right)\right\rvert\le\lVert \varphi(\cdot)\rVert_\infty,\label{eq:dom_num}
\end{align}
and the integrand in the denominator is dominated by $1$, that is
\begin{align}
    \left\lvert\exp\left(\frac{g(x^*-\sqrt{\varepsilon}u)}{\varepsilon}\right)\right\rvert\le1.\label{eq:dom_dom}
\end{align}

Together \cref{eq:dom_num,eq:dom_dom,eq:point_num,eq:point_dom} are the sufficient conditions for applying the dominated convergence theorem \citep{Bass13}, allowing us to commute all limits and integrals in \cref{eq:limit_u}, yielding our desired result:
\begin{align}
     \lim_{\varepsilon\rightarrow0}\int_{\mathcal{X}} \varphi(x)p_\varepsilon(x)dx&=\lim_{\varepsilon\rightarrow0}\left(\frac{\int_{\mathcal{U}} \varphi(x^*-\sqrt{\varepsilon}u)\exp\left(\frac{g(x^*-\sqrt{\varepsilon}u)}{\varepsilon}\right)du}{\int_\mathcal{U}\exp\left(\frac{g(x^*-\sqrt{\varepsilon}u)}{\varepsilon}\right)du}\right),\\
     &=\frac{\int_{\mathcal{U}}\lim_{\varepsilon\rightarrow0}\left( \varphi(x^*-\sqrt{\varepsilon}u)\exp\left(\frac{g(x^*-\sqrt{\varepsilon}u)}{\varepsilon}\right)\right)du}{\int_\mathcal{U}\lim_{\varepsilon\rightarrow0}\left(\exp\left(\frac{g(x^*-\sqrt{\varepsilon}u)}{\varepsilon}\right)\right)du},\\
     &=\frac{\int_{\mathcal{U}} \varphi(x^*)\exp\left(u^\top\nabla^2 f(x^*)u\right)du}{\int_\mathcal{U}\exp\left(u^\top\nabla^2 f(x^*)u\right)du},\\
     &=\varphi(x^*)\frac{\int_{\mathcal{U}} \exp\left(u^\top\nabla^2 f(x^*)u\right)du}{\int_\mathcal{U}\exp\left(u^\top\nabla^2 f(x^*)u\right)du},\\
     &=\varphi(x^*).
\end{align}

\end{proof}
\end{theorem}

\subsection{Optimal Boltzmann Distributions as Optimal Policies}
\label{proof:rl_elbo_ap}
\begin{lemma}[Lower and Upper limits of $\lp$] i) For any $\ew>0$ and $\pit=\delta(a^*)$, we have $\lp=-\infty$. ii) For $\hat{Q}_\omega(\cdot)>0$ and any non-deterministic $\pit$, $\lim_{\ew\rightarrow0}\lp=\infty$.
\label{proof:objective_infinity_ap}
\begin{proof}
To prove i), we substitute $\pit=\delta(a^*)$ into $\lp$ from \cref{eq:elborl}, yielding:
\begin{align}
    \lp&=\mathbb{E}_{s\sim d(s)}\left[\mathbb{E}_{a\sim \delta(a^*)} \left[\frac{\hQ}{\ew}\right]+\mathcal{H}(\delta(a^*))\right],\\
    &=\mathbb{E}_{s\sim d(s)}\left[\frac{\hat{Q}_\omega(a^*,s)}{\ew}+\mathcal{H}(\delta(a^*))\right],\label{eq:l_delta}
\end{align}

We now prove that $\mathscr{H}(\delta(a^*))=-\infty$ for any $a^*$. Let $p:\mathcal{X}\rightarrow[0,1]$ be any zero-mean, unit variance distribution. Using a transformation of variables, we have $\mathcal{A}=\sigma \mathcal{X}+a^*$ and hence $p(a)=\frac{1}{\sigma}p(\sigma x-a^*)$. We can therefore write our Dirac-delta distribution as
\begin{align}
\delta(a^*)=\lim_{\sigma\rightarrow0}p(a)=\lim_{\sigma\rightarrow0}\frac{1}{\sigma}p(\sigma x-a^*).
\end{align}
Substituting into the definition of differential entropy, we obtain:
\begin{align}
\mathscr{H}(\delta(a^*))&=\lim_{\sigma\rightarrow0}\mathscr{H}(p(a))\\
&=\lim_{\sigma\rightarrow0}\mathscr{H}\left(\frac{1}{\sigma}p(\sigma x-a^*)\right),\\
&=-\lim_{\sigma\rightarrow0}\int_\mathcal{A} \frac{1}{\sigma}p(\sigma x-a^*)\log\left(\frac{1}{\sigma}p(\sigma x-a^*)\right)da,\\
&=-\lim_{\sigma\rightarrow0}\int_\mathcal{A} \frac{1}{\sigma}p(\sigma x-a^*)\log\left(p(\sigma x-a^*)\right)da+\lim_{\sigma\rightarrow0}\int_\mathcal{A} \frac{1}{\sigma}p(\sigma x-a^*)\log\left(\sigma\right)da,\\
&=-\int_\mathcal{A} \delta(a^*)\log\left(p(-a^*)\right)da+\lim_{\sigma\rightarrow0}\log\left(\sigma\right),\\
&=-\log(p(-a^*))+\lim_{\sigma\rightarrow0}\log\left(\sigma\right),\label{eq:limit_entropy}\\
&=-\infty.
\end{align}
Substituting for $\mathscr{H}(\delta(a^*))$ from \cref{eq:limit_entropy} in \cref{eq:l_delta} yields our desired result:
\begin{align}
     \lp&=\mathbb{E}_{s\sim d(s)}\left[\frac{\hat{Q}_\omega(a^*,s)}{\ew}\right]+\mathbb{E}_{s\sim d(s)}\left[\mathcal{H}(\delta(a^*))\right],\\
     &=\frac{\mathbb{E}_{s\sim d(s)}\left[\hat{Q}_\omega(a^*,s)\right]}{\ew}+(\lim_{\sigma\rightarrow0}\log\left(\sigma\right)-\log(p(-a^*)))\mathbb{E}_{s\sim d(s)}\left[1\right],\\
     &=-\infty,
\end{align}
where our final line follows from the first term being finite for any $\ew>0$. 

To prove ii), we take the limit $\ew\rightarrow0$ of $\lp$ in \cref{eq:elborl}:
\begin{align}
    \lim_{\ew\rightarrow0}\lp& =\lim_{\ew\rightarrow0}\left( \frac{\mathbb{E}_{d(s)\pit}\left[\hQ\right]}{\ew}+\mathbb{E}_{d(s)}\left[\mathscr{H}(\pit)\right] \right),\\
    & =\lim_{\ew\rightarrow0}\left( \frac{\mathbb{E}_{d(s)\pit}\left[\hQ\right]}{\ew} \right)+\mathbb{E}_{d(s)}\left[\mathscr{H}(\pit)\right],\\
    & =\infty.
\end{align}
where our last line follows from $\mathscr{H}(\pit)$ being finite for any non-deterministic $\pit$ and $\hat{Q}_\omega(\cdot)>0\implies\mathbb{E}_{d(s)\pit}\left[\hQ\right]>0$.

\end{proof}
\end{lemma}

\begin{theorem}[Optimal Boltzmann Distributions as Optimal Policies]
For any pair $\{\omega^*,\theta^*\}$ that maximises $\lp$ defined in \cref{eq:elborl}, the corresponding variational policy induced must be optimal, i.e. $\{\omega^*,\theta^*\} \in \argmax_{\omega,\theta}\lp\implies \pi_{\omega^*}(a \vert  s)\in \Pi^*$. Moreover, any $\theta^*\ s.t.\ \pi_{\theta^*}(a\vert s)=\pi_{\omega^*}(a \vert  s) \implies\theta^*\in\argmax_{\omega,\theta}\lp $.
	\begin{proof}
		Our proof is structured as follows: Firstly, we prove that $\varepsilon_{\omega^*}=0$ is both a necessary and sufficient condition for any $\omega^*\in\argmax_{\omega,\theta}\lp$ with $\hat{Q}_{\omega^*}(\cdot)>0$. We then verify that $\hat{Q}_{\omega^*}(\cdot)>0$ is satisfied by our framework and $\varepsilon_{\omega^*}=0$ is feasible. Finally, we prove that $\varepsilon_{\omega^*}=0$ is sufficient for $\pi_{\omega^*}(a \vert  s)\in \Pi^*$.
		
		To prove necessity, assume there exists an optimal $\omega^*$ such that $\varepsilon_{\omega^*}\ne0$. As $\varepsilon_{\omega}\ge0$, it must be that $\varepsilon_{\omega^*}>0$. Consider $\lp$ as defined in \cref{eq:elborl}:
		\begin{align}
		     \lp=\frac{\mathbb{E}_{d(s)\pit}\left[\hQ\right]}{\ew}+\mathbb{E}_{d(s)}\left[\mathscr{H}(\pit)\right].
		\end{align}
		As $\pit$ has finite variance, $\mathscr{H}(\pit)$ is upper bounded, and as $\hat{Q}_\omega(\cdot)$ is upper bounded, $\mathbb{E}_{d(s)\pit}\left[\hQ\right]$ is upper bounded too. Together, this implies that $\mathbb{E}_{d(s)\pit}\left[\hQ\right]$ is upper bounded for $\varepsilon_{\omega^*}>0$. From \cref{assumption:policy_represent_optimal}, there exists $\omega^\diamond\in\Omega$ such that $\varepsilon_{\omega^\diamond}=0$. From \cref{proof:objective_infinity}, there exists $\theta^*$ such that $\lim_{\varepsilon_{\omega^*}\rightarrow0}\mathcal{L}(\omega^\diamond,\theta^*)=\infty$, implying $\mathcal{L}(\omega^*,\theta^*)<\mathcal{L}(\omega^\diamond,\theta^*)$ which is a contradiction.
		
		To prove sufficiency, we take $\argmax_\omega\lp$:
		\begin{align}
		\argmax_\omega\lp=&
		\argmax_\omega\left(\mathbb{E}_{d(s)\pit}\left[ \frac{\hQ}{\ew}\right]+\mathbb{E}_{d(s)}\left[\mathscr{H}(\pit)\right]\right),\\
		=&\argmax_\omega\left(\mathbb{E}_{d(s)\pit}\left[ \frac{\hQ}{\ew}\right]\right),\\
		=&\argmax_\omega\left( \frac{\mathbb{E}_{d(s)\pit}\left[\hQ\right]}{\ew}\right).\label{eq:expected_Q}\\
		\end{align}
		Assume that \circled{i} $\hat{Q}_{\omega^*}(\boldsymbol{\cdot})>0$. It then follows:
		\begin{align}
		\argmax_\omega\lp=&\argmax_\omega\left( \frac{\mathbb{E}_{d(s)\pit}\left[\hQ\right]}{\ew}\right),\\
		=&\argmin_\omega\left(\frac{\ew}{\mathbb{E}_{ d(s)\pit}\left[\hQ\right]}\right),\\
		=&\argmin_\omega\ew,\\
		\end{align}
		which, as $\ew\ge0$, is satisfied for any $\omega^*\in\Omega\ s.t.\ \varepsilon_{\omega^*}=0$, proving sufficiency.
		
		Assume now \circled{ii} $\hat{Q}_{\omega^*}(\boldsymbol{\cdot})$ is locally smooth with a unique maximum over actions according to \cref{def:locally_smooth}. Under this condition we can apply \cref{proof:app_convergence_boltzmann} and our Boltzmann distribution tends towards a Dirac-delta function:
		\begin{align}
		\pi_{\omega^*}(a\vert s)=\lim_{\ew\rightarrow0}\left(\frac{\exp\left(\frac{\hat{Q}_{\omega^*}(h)}{\ew}\right)}{\int \exp\left(\frac{\hat{Q}_{\omega^*}(h)}{\ew}\right)da}\right)=\delta(a=\argmax_{a'}\hat{Q}_{\omega^*}(s,a')), \label{eq:policy_optimal}
		\end{align}
		which is a greedy policy w.r.t. $\hat{Q}_{\omega^*}(\boldsymbol{\cdot})$. From \cref{def:target_set}, when $\lim_{\ew\rightarrow0}\pi_{\omega}(a\vert s)$ we have $\mathcal{T}_{\omega}\hQ=\mathcal{T}^*\hQ$. Substituting into $\varepsilon_{\omega^*}=0$ shows our our function approximator must satisfy an optimal Bellman equation:
		\begin{gather}
		    \varepsilon_{\omega^*}=\frac{c}{p}\lVert\mathcal{T}^*\hQ-\hQ\rVert_p^p=0,\\
		    \implies\mathcal{T}^*\hat{Q}_{\omega^*}(\boldsymbol{\cdot})=\hat{Q}_{\omega^*}(\boldsymbol{\cdot}),
		\end{gather}
		hence $\hat{Q}_{\omega^*}(\boldsymbol{\cdot})=Q^*(\boldsymbol{\cdot})$. Under \cref{assumption:policy_represent_optimal}, we see that there exists $\omega^*\in\Omega\ s.t.\ \varepsilon_{\omega^*}=0$ for $\hat{Q}_{\omega^*}(\boldsymbol{\cdot})=Q^*(\boldsymbol{\cdot})$, hence $\varepsilon_{\omega^*}=0$ is feasible. Moreover, our assumptions \circled{i} and \circled{ii} are satisfied for $\hat{Q}_{\omega^*}(\boldsymbol{\cdot})=Q^*(\boldsymbol{\cdot})$ under Assumptions \ref{assumption:policy_represent_optimal} and \ref{assumption:smoothness} respectively. Substituting for $\hat{Q}_{\omega^*}(\boldsymbol{\cdot})=Q^*(\boldsymbol{\cdot})$ into $\pi_{\omega^*}(a\vert s)$ from \cref{eq:policy_optimal} we recover our desired result:
		\begin{gather}
		\omega^* \in \argmax_\omega\lp\\ \implies  	\pi_{\omega^*}(a\vert s)=\delta(a=\argmax_{a'}Q^*(s,a'))\in \Pi^*.
		\end{gather}
		From \cref{proof:objective_infinity}, we have that $\lp\rightarrow\infty=\max_{\omega,\theta}\lp$ when $\ew=0$ for any $\theta^*\in\Theta$ such that the variational policy is non-deterministic, hence 
		\begin{align}
		    \{\omega^*,\theta^*\} \in \argmax_{\omega,\theta}\lp\implies \pi_{\omega^*}(a \vert  s)\in \Pi^*,
		\end{align}
		as required. 
	\end{proof}
\end{theorem}

\subsection{Maximising the ELBO for $\theta$}
\label{sec:proof_kl_app}
\begin{theorem}
[Maximising the ELBO for $\theta$]
Maximsing $\lp$ for $\theta$ with $\ew>0$ is equivalent to minimising the expected KL divergence between $\piw$ and $\pit$, i.e.  for any $\ew>0$, $\max_\theta\lp=\min_\theta\mathbb{E}_{d(s)}\left[\kl{\pit}{\piw}\right]$ with $\piw=\pit$ under exact representability.
\begin{proof}
    \label{app:proof_kl}
    Firstly, we write $\lp$ in terms of $\ell(\omega)$ and $\kl{\q}{\p}$ from \cref{eq:objective_ii}, ignoring the entropy term which has no dependency on $\omega$ and $\theta$:
    \begin{align}
    \lp = \ell(\omega)-\kl{\q}{\p},
    \end{align}
    which, for any $\ew>0$, implies
    \begin{align}
    \max_\theta \lp&=\max_\theta\left(\ell(\omega)-\kl{\q}{\p}\right).\\&=\min_\theta\left(\kl{\q}{\p}\right).\label{eq:lml_kl}
    \end{align}
    We now introduce the definition
    \begin{align}
        \pw(s)\coloneqq\frac{\int_\mathcal{A}\exp\left(\frac{\hQ}{\ew}\right)da}{\int_\mathcal{H}\exp\left(\frac{\hQ}{\ew}\right)dh}.
    \end{align}
    We now show that we can decompose $\p$ as $\p=\piw\pw(s)$:
    \begin{align}
        \p&=\frac{\exp\left(\frac{\hQ}{\ew}\right)}{\int_\mathcal{H}\exp\left(\frac{\hQ}{\ew}\right)dh},\\
        &=\frac{\exp\left(\frac{\hQ}{\ew}\right)}{\int_\mathcal{H}\exp\left(\frac{\hQ}{\ew}\right)dh}\cdot\frac{\int_\mathcal{A}\exp\left(\frac{\hQ}{\ew}\right)da}{\int_\mathcal{A}\exp\left(\frac{\hQ}{\ew}\right)da},\\
        &=\frac{\exp\left(\frac{\hQ}{\ew}\right)}{\int_\mathcal{A}\exp\left(\frac{\hQ}{\ew}\right)da}\cdot\frac{\int_\mathcal{A}\exp\left(\frac{\hQ}{\ew}\right)da}{\int_\mathcal{H}\exp\left(\frac{\hQ}{\ew}\right)dh},\\
        &=\piw\pw(s).
    \end{align}
    Substituting for $\p=\piw\pw(s)$ and $\q=d(s)\pit$ into the KL divergence from \cref{eq:lml_kl} yields:
	\begin{align}
	    \kl{\q}{\p}&=\mathbb{E}_{d(s)\pit}\left[\log\left(\frac{d(s)\pit}{\pw(s)\piw}\right)\right],\\
	    &=\mathbb{E}_{d(s)\pit}\left[\log\left(\frac{d(s)}{\pw(s)}\right)\right]+\mathbb{E}_{d(s)\pit}\left[\log\left(\frac{\pit}{\piw}\right)\right],\\
	    &=\mathbb{E}_{d(s)}\left[\log\left(\frac{d(s)}{\pw(s)}\right)\right]\mathbb{E}_{\pit}\left[1\right]+\mathbb{E}_{d(s)\pit}\left[\log\left(\frac{\pit}{\piw}\right)\right],\\
	    &=\mathbb{E}_{d(s)}\left[\log\left(\frac{d(s)}{\pw(s)}\right)\right]+\mathbb{E}_{d(s)}\left[\mathbb{E}_{\pit}\left[\log\left(\frac{\pit}{\piw}\right)\right]\right],\\
	    &=\kl{d(s)}{\pw(s)}+\mathbb{E}_{d(s)}\left[\kl{\pit}{\piw}\right].\label{eq:kl_split}
	\end{align}
	Observe that the first term in \cref{eq:kl_split} does not depend on $\theta$, hence taking the minimum yields our desired result:
	\begin{align}
	\max_\theta \lp&=\min_\theta\left(\kl{d(s)}{\pw(s)}+\mathbb{E}_{d(s)}\left[\kl{\pit}{\piw}\right]\right),\\
	&=\min_\theta\mathbb{E}_{d(s)}\left[\kl{\pit}{\piw}\right].
	\end{align}    
	Since $\kl{\pit}{\piw}\ge0$, it follows that  under exact representability, that is there exists $\theta\in\Theta$ s.t. $\pit=\piw$ and hence $\kl{\pit}{\piw}=0$, we have
	$\min_\theta\mathbb{E}_{d(s)}\left[\kl{\pit}{\piw}\right]=0$.
	\end{proof}
\end{theorem}

\section{Deriving the EM Algorithm}

\subsection{E-Step}
\label{proof:estep}
Here we provide a full derivation of our E-step of our variational actor-critic algorithm. The ELBO for our model from \cref{eq:elborl} with $\omega_k$ fixed is:
\begin{align}
    \mathcal{L}(\omega_k,\theta)=&\mathbb{E}_{s\sim d(s)}\left[\frac{\mathbb{E}_{a\sim\pit}\left[ \hat{Q}_{\omega_k}(h)\right]}{\varepsilon_{\omega_k}}+\mathscr{H}(\pit)\right].
\end{align}
Taking derivatives of the $\omega$-fixed ELBO with respect to $\theta$ yields:
\begin{align}
\nabla_\theta\mathcal{L}(\omega_k,\theta)=&\mathbb{E}_{s\sim d(s)}\left[\frac{\nabla_\theta\mathbb{E}_{a\sim\pit}\left[ \hat{Q}_{\omega_k}(h)\right]}{\varepsilon_{\omega_k}}+\nabla_\theta\mathscr{H}(\pit)\right],\\
=&\mathbb{E}_{s\sim d(s)}\left[\frac{\mathbb{E}_{a\sim\pit}\left[ \hat{Q}_{\omega_k}(h)\nabla_\theta\log\pit\right]}{\varepsilon_{\omega_k}}+\nabla_\theta\mathscr{H}(\pit)\right],
\end{align}
where we have used the log-derivative trick \citep{Sutton00} in deriving the final line. Note that in this form, when $\varepsilon_{\omega_k}\approx0$, our gradient signal becomes very large. To prevent ill-conditioning, we multiply our objective by the constant $\varepsilon_{\omega_k}$. As $\varepsilon_{\omega_k}>0$ for all non-optimal $\omega_k$ (see \cref{proof:forward}), this will not change the solution to the E-step optimisation. Our gradient becomes:
\begin{align}
\varepsilon_{\omega_k}\nabla_\theta\mathcal{L}(\omega_k,\theta)=&\mathbb{E}_{s\sim d(s)}\left[\mathbb{E}_{a\sim\pit}\left[\hat{Q}_{\omega_k}(h)\nabla_\theta\log\pit\right]+\varepsilon_{\omega_k}\nabla_\theta\mathscr{H}(\pit)\right],
\label{eq:E-step_full}
\end{align}
as required.

\subsection{M-Step}
\label{proof:mstep}
Here we provide a full derivation of the M-step for our variational actor-critic algorithm. The ELBO from \cref{eq:elborl} with $\theta_{k+1}$ fixed is:
\begin{align}
\mathcal{L}(\omega,\theta_{k+1})=\mathbb{E}_{ d(s)}\left[ \frac{\mathbb{E}_{\pi_{\theta_{k+1}}(a\vert s)}\left[\hQ\right]}{\ew}+\mathscr{H}(\pi_{\theta_{k+1}}(a\vert s))\right]
\end{align}
Taking derivatives of the with respect to $\omega$ yields:
\begin{align}
\nabla_\omega\mathcal{L}(\omega,\theta_{k+1})&=\mathbb{E}_{  d(s)\pi_{\theta_{k+1}}(a\vert s)}\left[ \nabla_\omega\left(\frac{\hQ}{\ew}\right)\right],\\
&=\mathbb{E}_{  d(s)\pi_{\theta_{k+1}}(a\vert s)}\left[ \frac{\nabla_\omega\hQ}{\ew}-\frac{\hQ}{(\ew)^2}\nabla_\omega\ew\right],\\
&=\frac{1}{\ew}\mathbb{E}_{  d(s)\pi_{\theta_{k+1}}(a\vert s)}\left[ \nabla_\omega\hQ\right]-\frac{1}{(\ew)^2}\mathbb{E}_{  d(s)\pi_{\theta_{k+1}}(a\vert s)}\left[\hQ\right]\nabla_\omega\ew,
\end{align}
where we note that $\ew$ does not depend on $h$, which allowed us to move it in and out of the expectation in deriving the final line. The gradient depends on terms up to $\frac{1}{(\ew)^2}$, and so we multiply our objective by $(\varepsilon_{\omega_i})^2$ to prevent ill-conditioning when $\ew\approx 0$. As $(\varepsilon_{\omega_i})^2>0$ for all non-convergent $\omega^*$, this does not change the solution to our M-step optimisation and can be seen as introducing an adaptive step size which supplements $\alpha_\textrm{critic}$. Observe that $\frac{\varepsilon_{\omega_i}}{\ew}\big\lvert_{\omega=\omega_i}=1$, which, with a slight abuse of notation, yields our desired result:
\begin{align}
(\varepsilon_{\omega_i})^2\nabla_\omega\mathcal{L}(\omega,\theta_{k+1})=\varepsilon_{\omega_i}\mathbb{E}_{ d(s)\pi_{\theta_{k+1}}(a\vert s)}\left[ \nabla_\omega\hQ\right]-\mathbb{E}_{ d(s)\pi_{\theta_{k+1}}(a\vert s)}\left[ \hQ\right]\nabla_\omega\ew.
\end{align}
In general, calculating the exact gradient of $\ew$ is non-trivial. We now derive this update for three important cases:

\subsection{Gradient of the Residual Error}
\label{sec:gradient_residual_error}
We define $\bw\coloneqq \mathcal{T}_\omega \hQ-\hQ$ and use the notation $\mathbb{E}[\cdot]\triangleq \mathbb{E}_{h\sim\mathcal{U}(h)}[\cdot]$. Taking the derivative yields:
\begin{align}
    \nabla_\omega\ew &=\frac{1}{2\lvert \mathcal{H}\rvert}\nabla_\omega\lVert\bw^2\rVert_2^2,\\ &=\frac{1}{2}\nabla_\omega\mathbb{E}\left[\bw^2\right],\\
    &=\mathbb{E}\left[\bw\nabla_\omega\bw\right].\label{eq:nabla_ew}
\end{align}
For targets that do not depend on $\piw$, the gradient of $\nabla_\omega\bw$ can be computed directly. As an example, consider the update for the optimal Bellman operator target $\mathcal{T}^*\boldsymbol{\cdot}\coloneqq r(h)+\gamma \mathbb{E}_{h'\sim p(s'\vert h)}\left[\max_{a'}(\boldsymbol{\cdot})\right]$:
\begin{align}
    \nabla_\omega\bw=\mathbb{E}_{s'\sim p(s'\vert h)}\left[\nabla_\omega\hat{Q}_\omega(a^*,s')\right]-\nabla_\omega\hQ,
\end{align}
where $a^*=\argmax_{a}\hat{Q}(a,s')$. 

Consider instead the Bellman operator target $\mathcal{T}^{\pi_\omega}\hQ=r(h)+\gamma \mathbb{E}_\omega\left[\hQd \right]$ for the Bellman policy $\piw$, which does have dependency on $\piw$. For convenience, we denote the expectation $\mathbb{E}_{h'\sim p(s'\vert h)\pi_\omega(a'\vert s')}\left[\cdot \right]$ as $\mathbb{E}_\omega\left[\cdot\right]$. To obtain an analytic gradient, we must solve a recursive equation for $\nabla_\omega\piw$. Consider the gradient of $\bw$ with respect to $\omega$ using this operator:
\begin{align}
    \nabla_\omega \bw&=\nabla_\omega\left(r(h)+\gamma \mathbb{E}_\omega\left[\hQd \right]-\hQ\right),\\
    &=\nabla_\omega\gamma \mathbb{E}_\omega\left[\hQd \right]-\nabla_\omega\hQ,\\
    &=\gamma \mathbb{E}_\omega\left[(\nabla_ \omega\log\pi_\omega(a'\vert s'))\hQd+\nabla_\omega\hQd \right]-\nabla_\omega\hQ,\\
    &=\gamma \mathbb{E}_\omega\left[(\nabla_ \omega\log\pi_\omega(a'\vert s'))\hQd\right]+\gamma \mathbb{E}_\omega\left[\nabla_\omega\hQd \right]-\nabla_\omega\hQ,\\
    &=\gamma \mathbb{E}_\omega\left[(\nabla_ \omega\log\pi_\omega(a'\vert s'))\hQd\right]+\Gamma_\omega(h),\label{eq:gradient_beta}
\end{align}
where $\Gamma_\omega(h)\coloneqq \gamma \mathbb{E}_\omega\left[\nabla_\omega\hQd \right]-\nabla_\omega\hQ$. Substituting \cref{eq:gradient_beta} into \cref{eq:nabla_ew}, we obtain:
\begin{align}
    \nabla_\omega\ew&=\mathbb{E}\left[\bw\nabla_\omega\bw\right],\\
    &=\gamma \mathbb{E}\left[\bw\mathbb{E}_\omega\left[(\nabla_ \omega\log\pi_\omega(a'\vert s'))\hQd\right]\right]+\mathbb{E}\left[\bw\Gamma_\omega(h)\right]\label{eq:gradient_beta_ii}
\end{align}

To find an analytic expression for the first term of \cref{eq:gradient_beta_ii}, we rely on the following theorem:

\begin{theorem}[Analytic Expression for Derivative of Boltzmann Policy Under Expectation]
\label{proof:piw_gradient}
If $\piw$ is the Boltzmann policy defined in \cref{eq:action_posterior}, it follows that:
\begin{align}
    \mathbb{E}\left[\bw\mathbb{E}_\omega\left[(\nabla_ \omega\log\pi_\omega(a'\vert s'))\hQd\right]\right]=\frac{\ew\mathbb{E}\left[\bw\Gamma_\omega(h)\right]\mathcal{E}_\omega\hQ+\mathcal{E}_\omega\left[\nabla_\omega\hQ\right]}{(\ew)^2\left(1+\gamma\mathbb{E}\left[\bw \mathbb{E}_\omega\left[\hQd\right]\right]\right)},
\end{align}
where $\mathcal{E}_\omega$ is the operator $\mathcal{E}_\omega\boldsymbol{\cdot}\coloneqq\mathbb{E}\left[\bw\mathbb{E}_\omega\left[\hQd\mathcal{M}_\omega\boldsymbol{\cdot} \right]\right]$ and $\mathcal{M}_\omega$ denotes the operator $\mathcal{M}_\omega[\boldsymbol{\cdot}]\coloneqq \boldsymbol{\cdot}-\mathbb{E}_{a\sim\piw}\left[\boldsymbol{\cdot}\right] $
\begin{proof}
consider the derivative  $\piw\nabla_\omega\log\piw$:
\begin{align}
    \piw\nabla_\omega\log\piw&=\nabla_\omega\piw,\\
    &=\nabla_\omega \frac{\exp\left(\frac{\hQ}{\ew}\right)}{\int_\mathcal{A}\exp\left(\frac{\hQ}{\ew}\right)da},\\
    &=\nabla_\omega \left(\frac{\hQ}{\ew}\right) \frac{\exp\left(\frac{\hQ}{\ew}\right)}{\int_\mathcal{A}\exp\left(\frac{\hQ}{\ew}\right)da}\\
    &\quad-\frac{\exp\left(\frac{\hQ}{\ew}\right)}{\int_\mathcal{A}\exp\left(\frac{\hQ}{\ew}\right)da}\cdot\frac{\int_\mathcal{A}\nabla_\omega \left(\frac{\hQ}{\ew}\right)\exp\left(\frac{\hQ}{\ew}\right)da}{\int_\mathcal{A}\exp\left(\frac{\hQ}{\ew}\right)da},\\
    &=\nabla_\omega \left(\frac{\hQ}{\ew}\right) \piw-\piw\int_\mathcal{A}\nabla_\omega \left(\frac{\hQ}{\ew}\right)\piw da,\\
    &=\nabla_\omega \left(\frac{\hQ}{\ew}\right) \piw-\piw\mathbb{E}_{a\sim\piw}\left[\nabla_\omega \left(\frac{\hQ}{\ew}\right)\right],\\
    &=\piw\left(\nabla_\omega \left(\frac{\hQ}{\ew}\right) -\mathbb{E}_{a\sim\piw}\left[\nabla_\omega \left(\frac{\hQ}{\ew}\right)\right]\right).\label{eq:piw_derivative}
\end{align}
Finding an expression for $\nabla_\omega \left(\frac{\hQ}{\ew}\right)$, we have:
\begin{align}
    \nabla_\omega \left(\frac{\hQ}{\ew}\right)=\frac{1}{(\ew)^2}\left(\ew\nabla_\omega\hQ-\hQ\nabla_\omega\ew\right).
\end{align}
Substituting into \cref{eq:piw_derivative}, we obtain:
\begin{align}
    \piw\nabla_\omega\log\piw&=\frac{\piw}{(\ew)^2}\bigg(\ew\nabla_\omega\hQ-\hQ\nabla_\omega\ew\\
    &\quad-\mathbb{E}_{a\sim\piw}\left[\ew\nabla_\omega\hQ-\hQ\nabla_\omega\ew\right]\bigg),\\
    &=\frac{\piw}{(\ew)^2}\bigg(\ew\left(\nabla_\omega\hQ-\mathbb{E}_{a\sim\piw}\left[\nabla_\omega\hQ\right]\right)\\
    &\quad+\nabla_\omega\ew\left(\mathbb{E}_{a\sim\piw}\left[\hQ\right]-\hQ\right)\bigg),\\
    &=\frac{\piw}{(\ew)^2}\left(\ew \mathcal{M}_\omega\left[\nabla_\omega\hQ\right] -\nabla_\omega\ew\mathcal{M}_\omega\hQ \right),
\end{align}
where $\mathcal{M}_\omega$ denotes the operator $\mathcal{M}_\omega[\boldsymbol{\cdot}]\coloneqq \boldsymbol{\cdot}-\mathbb{E}_{a\sim\piw}\left[\boldsymbol{\cdot}\right] $. Dividing both sides by $\piw$ yields:
\begin{align}
    \nabla_\omega\log\piw=\frac{1}{(\ew)^2}\left(\ew \mathcal{M}_\omega\left[\nabla_\omega\hQ\right] -\nabla_\omega\ew\mathcal{M}_\omega\hQ \right).
\end{align}
Now, substituting for $\nabla_\omega \ew=\mathbb{E}\left[\bw\nabla_\omega\bw\right]$ from \cref{eq:nabla_ew} yields:
\begin{align}
    \nabla_\omega\log\piw=\frac{1}{(\ew)^2}\left(\ew \mathcal{M}_\omega\left[\nabla_\omega\hQ\right] -\mathbb{E}\left[\bw\nabla_\omega\bw\right]\mathcal{M}_\omega\hQ \right).
\end{align}
Now substituting for $\nabla_\omega\bw=\gamma \mathbb{E}_\omega\left[(\nabla_ \omega\log\pi_\omega(a'\vert s'))\hQd\right]+\Gamma_\omega(h)$ from \cref{eq:gradient_beta}, and re-arranging for $\nabla_\omega\log\pi_\omega(a\vert s)$:
\begin{gather}
    \nabla_\omega\log\piw=\frac{1}{(\ew)^2}\bigg(\ew \mathcal{M}_\omega\left[\nabla_\omega\hQ\right] -\gamma \mathbb{E}\left[\bw\mathbb{E}_\omega\left[(\nabla_ \omega\log\pi_\omega(a'\vert s'))\hQd\right]\right]\\
    \quad+\mathbb{E}\left[\bw\Gamma_\omega(h)\right]\mathcal{M}_\omega\hQ \bigg),\\
    \nabla_\omega\log\piw+\gamma \mathbb{E}\left[\bw\mathbb{E}_\omega\left[(\nabla_ \omega\log\pi_\omega(a'\vert s'))\hQd\right]\right]=\frac{1}{(\ew)^2}\bigg(\ew \mathcal{M}_\omega\left[\nabla_\omega\hQ\right] \\
    \quad+\mathbb{E}\left[\bw\Gamma_\omega(h)\right]\mathcal{M}_\omega\hQ \bigg).
\end{gather}
Now, to obtain our desired result, we first multiply both sides by $\hQ$, take the expectation $\mathbb{E}_\omega$, multiply by $\bw$ and finally take the expectation $\mathbb{E}$:
\begin{align}
    &\mathbb{E}\left[\bw\mathbb{E}_\omega\left[(\nabla_ \omega\log\pi_\omega(a'\vert s'))\hQd\right]\right]\left(1+\gamma\mathbb{E}\left[\bw \mathbb{E}_\omega\left[\hQd\right]\right]\right) \\
    &\quad=\frac{1}{(\ew)^2}\bigg(\ew\mathbb{E}\left[\bw\mathbb{E}_\omega\left[\hQd \mathcal{M}_\omega\left[\nabla_\omega\hQd\right]\right]\right]\\
&\quad\quad\quad+\mathbb{E}\left[\bw\Gamma_\omega(h)\right]\mathbb{E}\left[\bw\mathbb{E}_\omega\left[\hQd\mathcal{M}_\omega\hQd \right]\right]\bigg).\\
&\mathbb{E}\left[\bw\mathbb{E}_\omega\left[(\nabla_ \omega\log\pi_\omega(a'\vert s'))\hQd\right]\right]=\frac{\mathbb{E}\left[\bw\mathbb{E}_\omega\left[\hQd \mathcal{M}_\omega\left[\nabla_\omega\hQd\right]\right]\right]}{\ew\left(1+\gamma\mathbb{E}\left[\bw \mathbb{E}_\omega\left[\hQd\right]\right]\right)}\\
&\quad\quad\quad+\frac{\mathbb{E}\left[\bw\Gamma_\omega(h)\right]\mathbb{E}\left[\bw\mathbb{E}_\omega\left[\hQd\mathcal{M}_\omega\hQd \right]\right]}{(\ew)^2\left(1+\gamma\mathbb{E}\left[\bw \mathbb{E}_\omega\left[\hQd\right]\right]\right)},\\
&\quad\quad=\frac{\ew\mathbb{E}\left[\bw\Gamma_\omega(h)\right]\mathcal{E}_\omega\hQ+\mathcal{E}_\omega\left[\nabla_\omega\hQ\right]}{(\ew)^2\left(1+\gamma\mathbb{E}\left[\bw \mathbb{E}_\omega\left[\hQd\right]\right]\right)},
\end{align}
as required.
\end{proof}
\end{theorem}
Using \cref{proof:piw_gradient} to substitute for $\mathbb{E}\left[\bw\mathbb{E}_\omega\left[(\nabla_ \omega\log\pi_\omega(a'\vert s'))\hQd\right]\right]$ into \cref{eq:gradient_beta}, we obtain the result:
\begin{align}
    \nabla\ew&=\frac{\ew\mathbb{E}\left[\bw\Gamma_\omega(h)\right]\mathcal{E}_\omega\hQ+\mathcal{E}_\omega\left[\nabla_\omega\hQ\right]}{(\ew)^2\left(1+\gamma\mathbb{E}\left[\bw \mathbb{E}_\omega\left[\hQd\right]\right]\right)}+\mathbb{E}\left[\bw\Gamma_\omega(h)\right].\label{eq:nabla_ew_full}
\end{align}
The second term of \cref{eq:nabla_ew_full} is the standard policy evaluation gradient and the first term changes $\piw$ in the direction of increasing $\ew$. We see that all expectations in \cref{eq:nabla_ew_full} can be approximated by sampling from our variational policy $\pit\approx\piw$. After a complete E-step, and under \cref{assumption:policy_represent}, we have $\pit=\piw$ and the gradient is exact. 

While the first term in \cref{eq:nabla_ew_full} is certainly tractable, it presents a formidable challenge for the programmer to implement, especially if unbiased estimates are required; several expressions which involve the multiplication of more than one expectation $\mathbb{E}_\omega$ need to be evaluated. In all of these cases, expectations approximated using the same data will introduce bias, however it is often infeasible to sample more than once from the same state in the environment. Like in \citet{Sutton09}, a solution to this problem is to learn a function approximator for one of the expectations that is updated at a slower rate than the other expectation. Alternatively, these function approximators can be updated using separate data batches from a replay buffer.

A radical approach is to simply neglect this gradient term, which we discuss in \cref{sec:approx_gradients}. A more considered approach is to use an operator that does not constraint $\Omega$. Consider the operator introduced in \cref{sec:relaxation_constraints}, 
\begin{align}
    \twk\boldsymbol{\cdot}=r(h)+\gamma\mathbb{E}_{\omega,k}\left[\boldsymbol{\cdot}\right],
\end{align}
where we have used the shorthand for expectation $\mathbb{E}_{\omega,k}\left[\boldsymbol{\cdot}\right]\coloneqq \mathbb{E}_{h'\sim p(s'\vert h)p_{\omega,k}(a'\vert s')}\left[\boldsymbol{\cdot}\right]$ and the Boltzmann distribution is defined as 
\begin{align}
   \pwk \coloneqq \frac{\exp\left(\frac{\hQ}{\ek}\right)}{\int_\mathcal{A} \exp\left(\frac{\hQ}{\ek}\right)da}.
\end{align}
The incremental residual error is defined as $\ewk\coloneqq \frac{1}{2\lvert \mathcal{H}\rvert}\lVert\bwk\rVert_2^2+\varepsilon_k$ and $\bwk\coloneqq\twk\hQ-\hQ$. Taking gradients of $\ewk$ directly yields:
\begin{align}
    \nabla_\omega\ewk=\mathbb{E}\left[\bwk\nabla_\omega\bwk\right].
\end{align}
where
\begin{align}
    \nabla_\omega\bwk&=\nabla_\omega\mathbb{E}_{\omega,k}\left[\hQd\right]-\nabla_\omega\hQ,\\
    &=\nabla_\omega\mathbb{E}_{\omega,k}\left[\hQd\right]-\nabla_\omega\hQ,\\
    &=\mathbb{E}_{\omega,k}\left[\nabla_\omega\log p_{\omega,k}(a'\vert s') +\nabla_\omega\hQd\right]-\nabla_\omega\hQ.\label{eq:nabla_bwk}
\end{align}
Now, $\nabla_\omega\log p_{\omega,k}(a'\vert s')$ can be computed directly as:
\begin{align}
    \nabla_\omega\log p_{\omega,k}(a'\vert s')&=\nabla_\omega\left(\frac{\hQd}{\ek}-\log\int_\mathcal{A} \exp\left(\frac{\hQd}{\ek}\right)da\right),\\
    &=\frac{\nabla_\omega\hQd}{\ek}-\int_\mathcal{A}\frac{\nabla_\omega\hQd}{\ek}\frac{\exp\left(\frac{\hQd}{\ek}\right)}{\int_\mathcal{A} \exp\left(\frac{\hQ}{\ek}\right)da}da,\\
    &=\frac{\nabla_\omega\hQd}{\ek}-\int_\mathcal{A}\frac{\nabla_\omega\hQ}{\ek} p_{\omega,k}(a'\vert s')da,\\
    &=\frac{\nabla_\omega\hQd}{\ek}-\mathbb{E}_{a'\sim p_{\omega,k}(a'\vert s')}\left[\frac{\nabla_\omega\hQ}{\ek} \right],\\
    &=\mathcal{M}_{\omega,k}\left[\frac{\nabla_\omega\hQd}{\ek}\right],
\end{align}
where where $\mathcal{M}_{\omega,k}$ denotes the operator $\mathcal{M}_{\omega,k}[\boldsymbol{\cdot}]\coloneqq \boldsymbol{\cdot}-\mathbb{E}_{a\sim\pwk}\left[\boldsymbol{\cdot}\right] $. Substituting into \cref{eq:nabla_bwk} yields:
\begin{align}
    \nabla_\omega\bwk=\mathbb{E}_{\omega,k}\left[\mathcal{M}_{\omega,k}\left[\frac{\nabla_\omega\hQd}{\ek}\right] +\nabla_\omega\hQd\right]-\nabla_\omega\hQ.
\end{align}
\subsection{Discussion of E-step}
\label{sec:E-step_discussion}

We now explore the relationship between classical actor-critic methods and the E-step. The policy gradient theorem \citep{Sutton00} derives an update for the derivative of the RL objective \eqref{eq:rl_objective} with respect to the policy parameters 
\begin{align}
\nabla_\theta J(\theta) = \mathbb{E}_{s\sim\rho^\pi(s)}\left[\mathbb{E}_{a\sim\pit}\left[ Q^\pi(h)\nabla_\theta\log\pit \right]\right],
\end{align}
where $\rho^\pi(s)$ is the discounted-ergodic occupancy, defined formally in \citet{epg-journal}, and in general not a normalised distribution. To obtain practical algorithms, we collect rollouts and treat them as samples from the steady-state distribution instead. 

By contrast, the \textsc{virel} policy update in \cref{eq:E-step_full} involves an expectation over $d(s)$, which can be any sampling distribution decorrelated from $\pi$ ensuring all states are visited infinitely often. As $\hQ$ is also independent of $\pit$, we can move the gradient operator $\nabla_\theta$ out of the inner integral to obtain 
\begin{align}
    \mathbb{E}_{s\sim d(s)}\left[\mathbb{E}_{a\sim\pit}\left[ \hQ\nabla_\theta\log\pit\right]\right]=\mathbb{E}_{s\sim d(s)}\left[\nabla_\theta\mathbb{E}_{a\sim\pit}\left[ \hQ\right]\right]
\end{align}
This transformation is essential in deriving powerful policy gradient methods such as Expected and Fourier Policy Gradients \citep{epg,fpg} and holds for deterministic polices \citep{dpg}. However, unlike in \textsc{virel}, it is not strictly justified in the classic policy gradient theorem \citep{Sutton00} and MERL formulation \citep{Haarnoja18}.
\section{Relaxations and Approximations}
\label{sec:relaxation}
\subsection{Relaxation of Representability of $Q$-functions}
In our analysis, \cref{assumption:policy_represent_optimal} is required by \cref{proof:forward} to ensure that a maximum to the optimisation problem exists, however it can be completely neglected provided that projected Bellman operators are used; moreover, if projected Bellman operators are used, our M-step is also always guaranteed to converge, even if our E-step does not. Consequently, we can terminate the algorithm by carrying out a complete M-step at any time using our variational approximation and still be guaranteed convergence to a sub-optimal point.  

We now introduce the assumption that our action-value function approximator is three-times differentiable over $\Omega$, which is required for convergence guarantees.
 \begin{assumption}[Universal Smoothness of $\hQ$] \label{assumption:hQ_c3} We require that $\hQ\in\mathbb{C}^3(\Omega)$ for all $h\in\mathcal{H}$,
 \end{assumption}
 We now extend the analysis of \citet{Maei09} to continuous domains. Consider the local linearisation of the function approximator $\hQ\approx b_\omega^\top (h)\omega$, where $b_\omega(h)\coloneqq\nabla_\omega \hQ$. We define the projection operator $\mathcal{P}_\omega Q(\cdot)\coloneqq b_\omega^\top(h)\omega'$ where $\tilde{\omega}$ are the parameters that minimise the difference between the action-value function and the local linearisation:
 \begin{align}
     \tilde{\omega}\coloneqq \argmin_{\omega'}\frac{1}{2\lvert\mathcal{H}\rvert}\lVert Q(h)- b_\omega^\top(h)\omega'\rVert^2_2.\label{eq:projection_min}
 \end{align}
Using the notation $\mathbb{E}[\cdot]\triangleq \mathbb{E}_{h\sim\mathcal{U}(h)}[\cdot]$ and taking derivatives of \cref{eq:projection_min} with respect to $\omega'$ yields:
\begin{align}
    \nabla_{\omega'}\frac{1}{2\lvert\mathcal{H}\rvert}\lVert Q(h)-{\omega'}^\top\rVert_2^2&=\frac{1}{2} \nabla_{\omega'}\mathbb{E}\left[ (Q(h)- b_\omega^\top(h)\omega')^2\right],\\
    &=\frac{1}{2}\mathbb{E}\left[\nabla_{\omega'}(Q(h)^2 -2 b_\omega^\top(h)\omega'Q(h)+  b_\omega^\top(h)\omega'b_\omega^\top(h) \omega'\right],\\
    &=\mathbb{E}\left[ b_\omega(h)b_\omega^\top(h)\omega'- b_\omega(h)Q(h)\right].
\end{align}
Equating to zero and solving for $\tilde{\omega}$, we obtain:
\begin{align}
    \tilde{\omega}=\mathbb{E} \left[b_\omega(h)b_\omega^\top(h)\right]^{-1}\mathbb{E} \left[b_\omega(h)Q(h)\right].
\end{align}
Substituting into our operator yields:
\begin{align}
    \mathcal{P}_\omega \boldsymbol{\cdot}=b_\omega^\top(h)\mathbb{E} \left[b_\omega(h)b_\omega^\top(h)\right]^{-1}\mathbb{E} \left[b_\omega(h)\boldsymbol{\cdot}\right].
\end{align}
We can therefore interpret $\mathcal{P}$ as an operator that projects an action-value function onto the tangent space of $\hQ$ at $\omega$. For linear function approximators of the form $\hQ=b^\top(h)\omega$, the projection operator is independent of $\omega$ and projects $Q$ directly onto the nearest function approximator and the operator  \citep{Sutton09a}.

We now replace the residual error in \cref{sec:model} with the projected residual error, 
\begin{align}
    \ew\coloneqq \frac{1}{2\lvert\mathcal{H}\rvert}\left\lVert \mathcal{P}_\omega\left(\mathcal{T}_\omega \hQ-\hQ
\right)\right\rVert_2^2.\label{eq:projected_error}
\end{align}
By definition, there always exists fixed point $\omega\in\Omega$ for which $\varepsilon_{\omega}=0$, which means that $\ew$ now satisfies all requirements in \cref{proof:forward} without \cref{assumption:policy_represent_optimal}. We can also carry out a complete partial variational M-step by minimising the surrogate $\ew$, keeping $\piw=\pit$ in all expectations. At convergence, we have $\ew=0$ in this case.

We now derive the more convenient form of $\ew$ from Lemma 1 in \citet{Maei09}, extending this result to continuous domains. Let $\bw\coloneqq \mathcal{T}_\omega \hQ-\hQ$. Substituting into \cref{eq:projected_error}, we obtain:
\begin{align}
    2\ew&=\frac{1}{\lvert\mathcal{H}\rvert}\left\lVert \mathcal{P}_\omega\bw\right\rVert_2^2,\\
&=\frac{1}{\lvert\mathcal{H}\rvert}\left\lVert b_\omega^\top(h)\mathbb{E} \left[b_\omega(h)b_\omega^\top(h)\right]^{-1}\mathbb{E} \left[b_\omega(h)\bw\right]\right\rVert_2^2,\\
&=\mathbb{E}\left[ \mathbb{E} \left[b^\top_\omega(h)\bw\right]\mathbb{E} \left[b_\omega(h)b_\omega^\top(h)\right]^{-1}b_\omega(h)b_\omega^\top(h)\mathbb{E} \left[b_\omega(h)b_\omega^\top(h)\right]^{-1}\mathbb{E} \left[\bw b_\omega(h)\right]\right],\\
&=\mathbb{E} \left[b^\top_\omega(h)\bw\right]\mathbb{E} \left[b_\omega(h)b_\omega^\top(h)\right]^{-1}\mathbb{E}\left[b_\omega(h)b_\omega^\top(h)\right]\mathbb{E} \left[b_\omega(h)b_\omega^\top(h)\right]^{-1}\mathbb{E} \left[\bw b_\omega(h)\right],\\
&=\mathbb{E} \left[b^\top_\omega(h)\bw\right]\mathbb{E} \left[b_\omega(h)b_\omega^\top(h)\right]^{-1}\mathbb{E} \left[\bw b_\omega(h)\right].
\end{align}
Denoting $\zeta_\omega\coloneqq \mathbb{E} \left[b_\omega(h)b_\omega^\top(h)\right]^{-1}\mathbb{E} \left[\bw b_\omega(h)\right]$ following the analysis in \citep{Maei09}, we find the derivative of $\ew$ as:
\begin{align}
\nabla_\omega\ew=\mathbb{E} \left[(\nabla_\omega\bw)b^\top_\omega(h)\zeta_\omega\right]+\mathbb{E} \left[(\bw-b^\top_\omega(h)\zeta_\omega)\nabla_\omega^2\hQ \zeta_\omega\right].
\end{align}
Following the method of \citet{Pearlmutter94}, the multiplication between the Hessian and $\zeta_\omega$ can be calculated in $O(n)$ time, which bounds the overall complexity of our algorithm. To avoid bias in our estimate, we learn a set of weights $\hat{\zeta}\approx \zeta_\omega$ on a slower timescale, which we update as:
\begin{align}
    \hat{\zeta}_{k+1}\leftarrow \hat{\zeta}_{k}+\alpha_{\zeta k}\left(\bw-b^\top_\omega(h)\zeta_k\right)b_\omega(h),\label{eq:zeta_update}
\end{align}
where $\alpha_{\zeta k}$ is a step size chosen to ensure that $\alpha_{\zeta k}<\alpha_\textrm{critic}$.  The weights are then used to find our gradient term:
\begin{align}
\nabla_\omega\ew=\mathbb{E} \left[(\nabla_\omega\bw)b^\top_\omega(h)\hat{\zeta}\right]+\mathbb{E} \left[(\bw-b^\top_\omega(h)\hat{\zeta})\nabla_\omega^2\hQ \zeta_\omega\right].\label{eq:ew_projected_gradient}
\end{align}
In our framework, the term $\nabla\bw$ is specific to our choice of operator. In \citet{Maei09}, a TD-target is used and parameter updates for $\omega$ are given as:
\begin{align}
    \omega_{k+1}&=\mathfrak{P}\left(\omega_k+\alpha_{\omega k}(b_k-\gamma b_k')b_k^\top\hat{\zeta}_k-q_k\right),\label{eq:omega_update}\\
    q_k&\coloneqq \left(\beta_{\omega_k}(h_k)-b_k^\top\hat{\zeta}_k\right)\nabla_\omega^2\hat{Q}_{\omega_k}(h_k)\hat{\zeta}_k
\end{align}
where $b_k\coloneqq b_{\omega_k}(h_k)$ and $\mathfrak{P}(\cdot)$ is an operator that projects $\omega_k$ into any arbitrary compact set with a smooth boundary, $\mathcal{C}$. The projection $\mathfrak{P}(\cdot)$ is introduced for mathematical formalism and, provided $\mathcal{C}$ is large enough to contain all solutions $\left\{\omega\vert\mathbb{E}\left[\bw \nabla_\omega\hQ\right]=0\right\}\subseteq\mathcal{C}$, has no bearing on the updates in practice. Under \cref{assumption:hQ_c3}, provided the step size conditions $\sum_k^\infty \alpha_{\zeta k}=\sum_k^\infty \alpha_{\omega k}=\infty $, $\sum_k^\infty \alpha_{\zeta k}^2<,\sum_k^\infty \alpha_{\omega k}^2<\infty $ and $\lim_{k\rightarrow \infty}\frac{\alpha_{\zeta k}}{\alpha_{\omega k}}=0$ hold and $\mathbb{E}[b_\omega(h)b_\omega^\top(h)]$ is non-singular $\forall\omega\in\Omega$, the analysis in Theorem 2 of \citet{Maei09} applies and the updates in \cref{eq:zeta_update,eq:omega_update} are guaranteed to converge to the TD fixed point.  This demonstrates using data sampled from any variational policy $\pit$ to update $\omega_k$ as \cref{eq:zeta_update,eq:omega_update}, $\omega_k$ will converge to a fixed point.

\label{sec:relaxation_assumption}

\subsection{Off-Policy Bellman Operators}
\label{sec:relaxation_constraints}
As discussed in \cref{sec:model}, using the Bellman operator $\mathcal{T}^{\pi_\omega}\boldsymbol{\cdot}$ induces a constraint on the set of parameters $\Omega$. While this constraint can be avoided using the optimal Bellman operator $\mathcal{T}^*\boldsymbol{\cdot}\coloneqq r(h)+\gamma \mathbb{E}_{h'\sim p(s'\vert h)}\left[\max_{a'}(\boldsymbol{\cdot})\right]$, evaluating $\max_{a'}(\hQd)$ may be difficult in large continuous domains. We now make a slight modification to our model in \cref{sec:model} to accommodate a Bellman operator that avoids these two practical difficulties.

Firstly, we introduce a new Boltzmann distribution $\pwk$:
 \begin{align}
\pwk \coloneqq \frac{\exp\left(\frac{\hQ}{\ek}\right)}{\int_\mathcal{A} \exp\left(\frac{\hQ}{\ek}\right)da},\label{eq:boltzmann_k}
\end{align}  
where $\{\ek\}$ is a sequence of positive constants $\ek\ge0$, $\lim_{k\rightarrow\infty}\ek=0$. We now introduce a new operator $\twk\boldsymbol{\cdot}$, defined as is the Bellman operator for $\pwk$:
 \begin{align}
     \twk\boldsymbol{\cdot}&\coloneqq\mathcal{T}^{p_{\omega,k}}\boldsymbol{\cdot}=r(h)+\gamma\mathbb{E}_{h'\sim p(s'\vert h)p_{\omega,k}(a'\vert s')}\left[\boldsymbol{\cdot}\right].\label{eq:twk}
 \end{align}
Let $\piwk$ be the Boltzmann policy:
 \begin{align}
\piwk \coloneqq \frac{\exp\left(\frac{\hQ}{\ewk}\right)}{\int_\mathcal{A} \exp\left(\frac{\hQ}{\ewk}\right)da},\label{eq:boltzmann_k}
\end{align}  
 where the residual error $\ewk\coloneqq \frac{c}{p}\lVert\twk\hQ-\hQ\rVert_p^p+\varepsilon_k$. It is clear that $\twk\boldsymbol{\cdot}$ does not constrain $\Omega$ as $\ek$ has no dependency on $\omega$ and $\piwk$ is well defined for all $\omega\in\Omega$.
 
 We now formally prove that $\min_\omega\lim_{k\rightarrow\infty}\ewk=\min_\omega\ew$, and so minimising $\ewk$ is the same as minimising the objective $\ew$ from \cref{sec:model} and that $\twk\boldsymbol{\cdot}\in\mathbb{T}$. We also prove that $\min_\omega\lim_{k\rightarrow\infty}\ewk=\lim_{k\rightarrow\infty}\min_\omega\ewk$ (i.e. that $\min$ and $\lim$ commute), which allows us to minimise our objective incrementally over sequences $\ewk$.

\begin{theorem}[Incremental Optimisation of $\ewk$]
 Let $\ewk\coloneqq \frac{c}{p}\lVert\twk\hQ-\hQ\rVert_p^p+\varepsilon_k$ and $\twk$ be the Bellman operator defined in \cref{eq:twk}. It follows that i) $\twk\boldsymbol{\cdot}\in\mathbb{T}$, ii) $\min_\omega\lim_{k\rightarrow\infty}\ewk=\min_\omega\ew$ and iii) $\min_\omega\lim_{k\rightarrow\infty}\ewk=\lim_{k\rightarrow\infty}\min_\omega\ewk$
\begin{proof}
 
 To prove i), we take the limit $\lim_{k\rightarrow\infty}\twk\hQ=\mathcal{T}^*\hQ$:
 \begin{align}
     \lim_{k\rightarrow\infty}\twk\hQ&=r(h)+\lim_{k\rightarrow\infty}\gamma\mathbb{E}_{h'\sim p(s'\vert h)p_{\omega,k}(a'\vert s')}\left[\hQ\right].
 \end{align}
 Observe that from \cref{proof:convergence_boltzmann}, we have
 \begin{align}
    \lim_{\ek\rightarrow\infty}\gamma\mathbb{E}_{h'\sim p(s'\vert h)p_{\omega,k}(a'\vert s')}\left[\hQ\right]=\gamma\mathbb{E}_{h'\sim p(s'\vert h)\delta(a=\argmax_{a'}(\hat{Q}_\omega(a',s))}\left[\hQ\right],
 \end{align}
 hence:
\begin{align}
     \lim_{k\rightarrow\infty}\twk\hQ&=r(h)+\lim_{k\rightarrow\infty}\gamma\mathbb{E}_{h'\sim p(s'\vert h)p_{\omega,k}(a'\vert s')}\left[\hQ\right],\\
     &=r(h)+\gamma\mathbb{E}_{h'\sim p(s'\vert h)\delta(a=\argmax_{a'}(\hat{Q}_\omega(a',s))}\left[\hQ\right],\\
     &=r(h)+\gamma\mathbb{E}_{s'\sim p(s'\vert h)}\left[\max_{a'}(\hQ)\right],\\
     &=\mathcal{T}^*\hQ.
 \end{align}
Our operator is therefore constructed such that in the limit $k\rightarrow\infty$, we recover the optimal Bellman operator. Observe too that as $\frac{c}{p}\lVert\twk\hQ-\hQ\rVert_p^p\ge0$, we have $\ewk>0$ for all $\ek>0$. From \cref{proof:convergence_boltzmann}, it follows that $\piwk\rightarrow \delta(a=\argmax_{a'}(\hat{Q}_\omega(a',s))$ only in the limit $\lim_{k\rightarrow\infty}\ek=0$ and when $\ewk=0$. Under this limit, we have  $\lim_{k\rightarrow\infty}\twk=\mathcal{T}^*$ and so $\twk\in\mathbb{T}$, as required for i).

To prove ii), consider taking the limit of $\ewk$ directly:
\begin{align}
\lim_{k\rightarrow\infty}\ewk&=\lim_{k\rightarrow\infty}\left(\frac{c}{p}\lVert\twk\hQ-\hQ\rVert_p^p+\varepsilon_k\right),\\
&=\lim_{k\rightarrow\infty}\left(\frac{c}{p}\lVert\twk\hQ-\hQ\rVert_p^p\right)+\varepsilon_{\infty},\\
&=\frac{c}{p}\lVert\lim_{k\rightarrow\infty}\twk\hQ-\hQ\rVert_p^p,\\
&=\frac{c}{p}\lVert\mathcal{T}^*\hQ-\hQ\rVert_p^p,\\
&=\ew,\label{eq:lim_ewk}
\end{align}
as required.

To prove iii), let $\tilde{\omega}_k$ be the minimiser of $\ewk$, that is $\tilde{\omega}_k=\argmin_\omega\ewk$. Let $\tilde{\omega}$ be the limit of all such sequences $\tilde{\omega}=\lim_{k\rightarrow\infty}\tilde{\omega}_k$ and let $\omega^*=\argmin_\omega\ew$. By definition, we have $\varepsilon_{\tilde{\omega}_k,k}\le\ewk$. Taking the limit $k\rightarrow\infty$ and then the $\min$, we have: 
\begin{gather}
    \min\lim_{k\rightarrow\infty}\varepsilon_{\tilde{\omega}_k,k}\le\min\lim_{k\rightarrow\infty}\ewk,\\
    \implies\varepsilon_{\tilde{\omega},\infty}\le\min\lim_{k\rightarrow\infty}\ewk.\label{eq:minlim}
\end{gather}
Using \cref{assumption:policy_represent_optimal} and \cref{eq:lim_ewk}, it follows that the right hand side of \cref{eq:minlim} is $\min\lim_{k\rightarrow\infty}\ewk=\min\ew=0$, hence $\varepsilon_{\tilde{\omega},\infty}\le0$. By definition, $\varepsilon_{\tilde{\omega},\infty}\ge0$, and so equality must hold. It therefore follows $\lim_{k\rightarrow\infty}\min_\omega\ewk=\varepsilon_{\tilde{\omega},\infty}=0$, which implies $\min_\omega\lim_{k\rightarrow\infty}\ewk=\lim_{k\rightarrow\infty}\min_\omega\ew=0$ as required.
\end{proof}
\end{theorem}

Overall, this result permits us to carry out separate optimisations over $\ewk$ while gradually increasing $k\rightarrow\infty$ to obtain the same result as minimising $\ew$ directly. The advantage to this method is that each minimisation $\ewk$ involves the operator $\twk$, which is tractable, mathematically convenient and does not constrain $\Omega$. Note too that, as calculated in \cref{sec:gradient_residual_error}, the gradient $\nabla_\omega\ewk$ is straightforward to implement in comparison with $\nabla_\omega\ew$ using $\mathcal{T}^{\pi_\omega}$. We save investigating this operator further for future work.

\subsection{Approximate Gradient Methods and Partial Optimisation}
\label{sec:approx_gradients}
A common trick in policy evaluation is to use a direct method \citep{Baird95, Sutton17}. Like in supervised methods \citep{Bishop2006}, direct methods treat the term $\tw\hQ$ as a fixed target, rather than a differential function. Introducing the notation $\mathbb{E}[\cdot]\triangleq \mathbb{E}_{h\sim\mathcal{U}(h)}[\cdot]$, the gradient can easily be derived as:
\begin{align}
    \nabla_\omega\ew&=\frac{1}{2}\nabla_\omega \mathbb{E}\left[\left(\stopgrad\left[\tw\hQ\right]-\hQ\right)^2\right],\\
    &= -\mathbb{E}\left[\left(\hQ-\tw\hQ\right)\nabla_\omega\hQ\right]
\end{align}
where $\stopgrad\left[\boldsymbol{\cdot}\right]$ is the stopgrad operator, which sets the gradient of its operand to zero, $\stopgrad\left[\boldsymbol{\cdot}\right]=\boldsymbol{\cdot}$, $\nabla\stopgrad\left[\boldsymbol{\cdot}\right]=0$ \citep{Foerster18}. For general function approximators, direct methods have no convergence guarantees, and indeed there exist several famous examples of divergence when used with classic RL targets \citep{Bertsekas96, Tsitsiklis1997, Williams93}, however its ubiquity in the RL community is testament to its ease of implementation and empirical success \citep{dqn,Sutton17}. We therefore see no reason why it should not be successful for \textsc{virel}, a claim which we verify in \cref{sec:experiments}. In our setting, we replace our M-step with the simplified objective $\omega_{k+1}\leftarrow \argmin_\omega\ew$. This is justified because $\argmin_\omega\ew$ was the original objective motivated in \cref{sec:model}, and so the only limitation to minimising this directly is obtaining a good enough variational policy $\piw\approx\pit$. More formally, our objective $\lp$ is maximised for any $\ew\rightarrow0$, so $\argmin_\omega\ew$ can be considered a surrogate objective for $\lp$. Using direct methods, M-step update becomes:
\paragraph{M-Step (Critic) direct:} $
\omega_{i+1}\leftarrow\omega_{i} - \alpha_{\textrm{critic}}\nabla_\omega \ew\rvert_{\omega=\omega_i},$
\begin{align}
&\nabla_\omega \ew=\mathbb{E}\left[\left(\hQ-\tw\hQ\right)\nabla_\omega\hQ\right].
\end{align}
We can approximate $\tw\hQ$ by sampling from the variational distribution $\pit$ and by using any appropriate RL target. Another important approximation that we make is that we perform only partial E- and M-steps, halting optimisation before convergence. From a practical perspective, convergence can often only occur in a limit of infinite time steps anyway, and if good empirical performance result from taking partial E- and M-steps, computation may be wasted carrying out many sub-optimisation steps for little gain.

As analysed by \citet{Gunawardana05}, such algorithms fall under the umbrella of the generalised alternating maximisation (GAM) framework, and convergence guarantees are specific to the form of function approximator and MDP. Like in many inference settings, we anticipate that most function approximators and MDPs of interest will not satisfy the conditions required to prove convergence, however variational EM procedures are known to be to empirically successful even when convergence properties are not guaranteed \citep{Gunawardana05,Turner11}. We demonstrate in \cref{sec:experiments} that taking partial EM steps does not hinder our performance. 

\subsection{Local Smoothness of $\hat{Q}_{\omega^*}({\cdot})$}
\label{sec:approx_local_smoothness}
For \cref{proof:convergence_boltzmann} to hold, we require that $\hat{Q}_{\omega^*}({\cdot})$ is locally smooth about its maximum. Our choice of function approximator may prevent this condition from holding, for example, a neural network with ReLU elements can introduce a discontinuity in gradient at $\max_h\hat{Q}_{\omega^*}(h)$. In practice, a formal Dirac-delta function can only ever emerge in the limit of convergence $\ew\rightarrow0$. In finite time, we obtain, at best, a nascent delta function; that is a function with very small variance that is `on the way to convergence' (see, for example, \citet{Kelly08} for a formal definition). The mode of a nascent delta function therefore approximates the true Dirac-delta distribution. When $\hat{Q}_{\omega^*}({\cdot})$ is not locally smooth, functions that behave similarly to nascent delta functions will still emerge at finite time, the mode of which we anticipate provides an approximation to the hardmax behaviour we require for most RL settings.

We also require that $\hat{Q}_{\omega^*}({\cdot})$ has a single, unique global maximum for any state. In reality, optimal Q-functions may have more than one global maxima for a single state corresponding to the existence of multiple optimal policies. To ensure \cref{assumption:smoothness} strictly holds, we can arbitrarily reduce the reward for all but one optimal policy. We anticipate that this is unnecessary in practice, as our risk-neutral objective means that a variational policy will be encouraged fit to a single mode anyway. In addition, these assumptions are required to characterise behaviour under convergence to a solution and will not present a problem in finite time where $\hQ$ is very unlikely to have more than one global optimum.
\subsection{Analysis of Approximate EM Algorithms}
\label{sec:analysis_of_approximations}
We now provide two separate analyses of our EM algorithm, replacing the Bellman operator $\mathcal{T}^{\pi_\omega}\cdot$ with its unconstrained variational approximation $\mathcal{T}^{\pi_\theta}\cdot$ (effectively substituting for $\pi_\omega(\cdot\vert s)\approx\pi_\theta(\cdot\vert s)$ under expectation). In our first analysis, we make no simplifying assumptions on $\varepsilon_\omega$, showing that our EM algorithm reduces exactly to policy iteration and evaluation. In our second analysis, we use a direct method, treating $\mathcal{T}^{\pi_\theta}\cdot$ as a fixed target as outlined in \cref{sec:approx_gradients}, showing that the algorithm reduces exactly to Q-learning. 

In both analyses, we assume a complete E- and M- step can be carried out and our class of function approximators is rich enough to represent any action-value function. Let $\pi_{\theta_0}(a\vert \cdot)$ be any initial policy and $\hat{Q}_{\omega_0}(\cdot)$ an arbitrary initialisation of the function approximator. For notational convenience we write $\pi_k(a\vert \cdot)\coloneqq \pi_{\theta_k}(a\vert \cdot)$.

\paragraph{Analysis with $\omega$-Dependent Target}

As we prove in \cref{proof:forward}, we can always maximise our objective with respect to $\omega$ by finding $\omega^*$ s.t. $\varepsilon_{\omega^*}=0$. This gives the M-step update:
\begin{gather}
    \omega_1=\argmin_\omega \varepsilon_{\omega},\\
    \implies \varepsilon_{\omega_1}=0,\\
    \implies \mathcal{T}^{\pi_0}\hat{Q}_{\omega_1}=\hat{Q}_{\omega_1} ,\\
    \implies \hat{Q}_{\omega_1}=Q^{\pi_0}(\cdot).
\end{gather}
Our E-step amounts to calculating the Boltzmann distribution with $\varepsilon_{\omega_1}=0$, which from \cref{proof:convergence_boltzmann} takes the form of a Dirac-delta distribution:
\begin{gather}
    \pi_1(a\vert \cdot)=\delta\left(a=\argmax_{a'}Q^{\pi_0}(a',\cdot)\right).
\end{gather}
We can generalise to the $k$th EM update as:
\begin{gather}
\hat{Q}_{\omega_k}(\cdot)\leftarrow Q^{\pi_{k-1}}(\cdot), \label{eq:pol_it_i}
\end{gather}
\begin{gather}
\pi_k(a\vert \cdot)\leftarrow \delta\left(a=\argmax_{a'}Q^{\pi_{k-1}}(a',\cdot)\right).\label{eq:pol_it_ii}
\end{gather}
Together Eqs \ref{eq:pol_it_i} and \ref{eq:pol_it_ii} are exactly the updates for policy iteration, an algorithm which is known to converge to the optimal policy \citep{rl_algorithms,Sutton17}. We therefore see that, even ignoring the constraint on $\Omega$, the optimal solution is still an attractive fixed point when our algorithms are carried out exactly. Using partial E- and M-steps give a variational approximation to the complete EM algorithm. We now provide a similar analysis using the fixed target of direct methods introduced in \cref{sec:approx_gradients}.

\paragraph{Analysis with Fixed Target}
Using a direct method, we replace the residual error with the fixed target residual error $\varepsilon_{\omega}\approx\varepsilon_{\omega,k}\coloneqq\frac{c}{p}\lVert \mathcal{T}^{\pi_k}\hat{Q}_{\omega_k}- \hat{Q}_{\omega}\rVert_p^p$, giving the M-step update:
\begin{gather}
    \omega_1=\argmin_\omega \varepsilon_{\omega,0} 
\end{gather}
which, under our assumption of representability, is achieved for
\begin{gather}
\varepsilon_{\omega_1,0}=0,\\
\implies \hat{Q}_{\omega_1}(\cdot)=\mathcal{T}^{\pi_0}\hat{Q}_{\omega_0}.
\end{gather}
As with our $\omega$-dependent target, the E-step amounts to calculating the Boltzmann distribution with $\varepsilon_{\omega_1,0}=0$, which from \cref{proof:convergence_boltzmann} takes the form of a Dirac-delta distribution:
\begin{gather}
    \pi_1(a\vert \cdot)=\delta\left(a=\argmax_{a'}\hat{Q}_{\omega_1}(a',\cdot)\right).
\end{gather}
We see that for any policy and function approximator, carrying out a complete E- and M- step results in a deterministic policy being learnt in this approximate regime. We generalise to the $k$th EM updates for $k>2$ as:
\begin{gather}
    \pi_{k-1}(a\vert \cdot)=\delta\left(a=\argmax_{a'}\hat{Q}_{\omega_{k-1}}(a',\cdot)\right),\\
    \omega_k=\argmin_\omega \varepsilon_{\omega,k-1}= \argmin_\omega \frac{c}{p}\lVert \mathcal{T}^{\pi_{k-1}}\hat{Q}_{\omega_{k-1}}- \hat{Q}_{\omega}\rVert_p^p ,\\
    \implies \varepsilon_{\omega_1,0}=0,\\
    \implies \hat{Q}_{\omega_k}(\cdot)=\mathcal{T}^{\pi_{k-1}}\hat{Q}_{\omega_{k-1}}(\cdot),\\
    =r(\cdot)+\mathbb{E}_{s'\vert \cdot}\left[\max_{a'}\hat{Q}_{\omega_{k-1}}(s',a')\right],\\
    =\mathcal{T}^*\hat{Q}_{\omega_{k-1}}(\cdot). \label{eq:q_it}
    \end{gather}
From \cref{eq:q_it}, we see that the EM algorithm with complete E- and M- steps implements $Q$-learning updates on our function approximator $\hat{Q}_{\omega_{k}}(\cdot)\leftarrow\mathcal{T}^*\hat{Q}_{\omega_{k-1}}(\cdot)$ for $k>2$ \citep{qlearning}. See \citet{Yang19} for a theoretical exposition of the convergence this $Q$-learning algorithm using function approximators. When partial EM steps are carried out, we can view this algorithm as a variational approximation to $Q$-learning. 
\section{Recovering MPO}
\label{sec:recovering_mpo}
We now derive the MPO objective from our framework. Under the probabilistic interpretation in \cref{sec:probablistic_interpretation}, the objective can be derived using the prior $p_\phi(h)=\mathcal{U}(s)\pip$ instead of the uniform distribution. Following the same analysis as in \cref{sec:probablistic_interpretation}, this yields an action-posterior:
\begin{align}
    p_{\omega,\phi}(a\vert s,\mathcal{O})&=\frac{\exp\left(\frac{\hQ}{\ew}\right)\pip}{\int\exp\left(\frac{\hQ}{\ew}\right)\pip da}.
\end{align}
Again, following the same analysis as in \cref{sec:probablistic_interpretation}, our ELBO objective is:
\begin{align}
    \mathcal{L}(\omega,\theta,\phi)&=\mathbb{E}_{d(s)}\left[\mathbb{E}_{\pit} \left[\frac{\hQ}{\ew}\right]-\kl{\pit}{\pip}\right]. \label{eq:mpo_objective}
\end{align}
Including a hyper-prior $p(\phi)$ over $\phi$ adds an additional term to $\mathcal{L}(\omega,\theta,\phi)$:
\begin{align}
    \mathcal{L}(\omega,\theta,\phi)&=\mathbb{E}_{d(s)}\left[\mathbb{E}_{\pit} \left[\frac{\hQ}{\ew}\right]-\kl{\pit}{\pip}\right]+\log p(\phi). 
\end{align}
which is exactly the MPO objective, with an adaptive scaling constant $\ew$ to balance the influence of $\kl{\pit}{\pip}$. Without loss of generality, we ignore the hyperprior and analyse \cref{eq:mpo_objective} instead. 

As discussed by \citet{Abdolmaleki18}, the MPO objective is similar to the PPO \citep{Schulman2017ppo} objective with the KL-direction reversed. In our E-step, we find a new variational distribution $\pi_{\theta_{k+1}}(a\vert s)$ that maximises the ELBO with $\omega_k$ fixed: Doing so yields an identical E-step to MPO. In parametric form, we can use gradient ascent and apply the same analysis as in \cref{proof:estep}, obtaining an update 
\paragraph{ E-Step (MPO):} $\theta_{i+1}\leftarrow\theta_{i} + \alpha_{\textrm{actor}}\left(\varepsilon_{\omega_k}\nabla_\theta \mathcal{L}(\omega_k,\phi_k,\theta)\vert_{\theta=\theta_i}\right)$,\\
\begin{align}
\varepsilon_{\omega_k}\nabla_\theta \mathcal{L}(\omega_k,\phi_k,\theta)=\mathbb{E}_{ d(s)}\left[\mathbb{E}_{\pit}\left[ \hat{Q}_{\omega_k}(h)\nabla_\theta\log\pit\right]-\varepsilon_{\omega_k}\nabla_\theta\kl{\pit}{\pi_{\phi_k}(a\vert s)}\right].\label{eq:e-mpo}
\end{align}
As a point of comparison, \citet{Abdolmaleki18} motivate the update in \cref{eq:e-mpo} by carrying out a partial E-step, maximising the ``one-step'' KL-regularised pseudo-likelihood objective. In our framework, maximising \cref{eq:e-mpo} constitutes a full E-step, without requiring approximation.

In our M-step, we maximise the LML using the posterior derived from the E-step, yielding the update:
\paragraph{ M-Step (MPO):} $\omega_{k+1},\phi_{k+1}\leftarrow\argmax_{\omega,\phi}\mathcal{L}(\omega,\phi,\theta_{k+1})$,\\
\begin{align}
\argmax_{\omega,\phi}\mathcal{L}(\omega,\phi,\theta_{k+1})=\argmax_{\omega,\phi}\left(\mathbb{E}_{ d(s)}\left[\mathbb{E}_{\pi_{\theta_{k+1}}}\left[\frac{\hQ}{\ew}\right]-\kl{\pi_{\theta_{k+1}}}{\pip}\right]\right).
\end{align}
Maximising for $\phi$ can be achieved exactly by setting $\pip=\pi_{\theta_{k+1}}(a\vert s)$, under which $\kl{\pi_{\theta_{k+1}}}{\pip}=0$. Maximising for $\omega$ is equivalent to finding $\argmax_\omega\mathbb{E}_{d(s)\pi_{\theta_{k+1}}}\left[\frac{\hQ}{\ew}\right]$, which accounts for the missing policy evaluation step, and can be implemented using the gradient ascent updates from \cref{eq:m_step}. Setting $\pip=\pi_{\theta_{k+1}}(a\vert s)$ is exactly the M-step update for MPO and, like in TRPO \citep{Schulman15b}, means that $\pip$ can be interpreted as the old policy, which is updated only after policy improvement. The objective in \cref{eq:mpo_objective} therefore prevents policy improvement from straying too far from the old policy, adding a penalisation term $\kl{\pit}{\pi_\textsc{old}(a\vert s)}$ to the classic RL objective.

\section{Variational Actor-Critic Algorithm Pseudocode}
\label{app:pseudocode}

\cref{alg:variatinal_actor_critic_virel,alg:variatinal_actor_critic_beta} show the pseudocode for the variational actor-critic algorithms $\textit{virel}$ and $\textit{beta}$  described in \cref{sec:experiments}. The respective objectives are:
\begin{align}
    J^V(\phi)=&\mathbb{E}_{s_t\sim\mathcal{D}}\left[\frac{1}{2}\left(V_{\phi}(s_t)-\mathbb{E}_{a_t\sim\pi_{\theta}}\left[Q_\omega(s_t,a_t)\right]\right)^2\right],\\
    J^Q(\omega)=&\mathbb{E}_{(h_t,r_t,s_{t+1})\sim\mathcal{D}}\bigg[\frac{1}{2}\left(r_t+\gamma V_{\bar{\phi}}(s_{t+1})-Q_\omega(h_t)\right)^2\bigg],\\
    J^{\pi^q}_{\textit{virel}}(\theta)=&\mathbb{E}_{h_t\sim\mathcal{D}}\bigg[\log\pi_\theta(a_t\vert s_t)(\alpha-(Q_\omega(h_t)-V_{\bar{\phi}}(s_{t})))\bigg],\\
     J^{\pi^q}_{\textit{beta}}(\theta)=&\mathbb{E}_{h_t\sim\mathcal{D}}\bigg[\log\pi_\theta(a_t\vert s_t)\bigg( \frac{1-\gamma}{r_{avg}}\ew-(Q_\omega(h_t)-V_{\bar{\phi}}(s_{t}))\bigg)\bigg].
\end{align}
Note that the derivative of the policy objectives can be found using the reparametrisation trick \citep{auto_bayes,svg}, which we use for our implementation.
\begin{multicols}{2}
\begin{minipage}{.5\textwidth}
  \begin{algorithm}[H]
\caption{Variational Actor-Critic: $\textit{virel}$}
	\label{alg:variatinal_actor_critic_virel}
	\begin{algorithmic}
		\STATE \mbox{Initialize parameter vectors $\phi$, $\bar{\phi}$, $\theta$, $\omega$,  $\mathcal{D} \leftarrow \left\{ \right\}$ }
		\FOR{each iteration}
		\FOR{each environment step}
		\STATE $a_t \sim {\pi^q}(a \vert s; \theta)$
		\STATE $s_{t+1} \sim p(s_{t+1} \vert s_t, a_t)$
		\STATE $\mathcal{D} \leftarrow \mathcal{D} \cup \left\{(s_t, a_t, r(s_t, a_t), s_{t+1} )\right\}$
		\ENDFOR
		\FOR{each gradient step}
		\item $\phi \leftarrow \phi - \lambda_V \hat \nabla_\phi J^V(\phi)$ (M-step)
		\STATE $\omega \leftarrow \omega - \lambda_Q \hat \nabla_{\omega} J^Q(\omega)$ (M-step)
		\STATE $\theta \leftarrow \theta - \lambda_{\pi^q} \hat \nabla_\theta J^{\pi^q}_{\textit{virel}}(\theta)$ (E-step)
		\STATE $\bar{\phi} \leftarrow \tau \phi+ (1-\tau)\bar{\phi}$
		\ENDFOR
		\ENDFOR
	\end{algorithmic}
\end{algorithm}
\end{minipage}
\vfill\null
\columnbreak
\begin{minipage}{.5\textwidth}
  \begin{algorithm}[H]
    \caption{Variational Actor-Critic: $\textit{beta}$}
	\label{alg:variatinal_actor_critic_beta}
	\begin{algorithmic}
		\STATE \mbox{Initialize parameter vectors $\phi$, $\bar{\phi}$, $\theta$, $\omega$,  $\mathcal{D} \leftarrow \left\{ \right\}$ }
		\FOR{each iteration}
		\FOR{each environment step}
		\STATE $a_t \sim {\pi^q}(a \vert s; \theta)$
		\STATE $s_{t+1} \sim p(s_{t+1} \vert s_t, a_t)$
		\STATE $\mathcal{D} \leftarrow \mathcal{D} \cup \left\{(s_t, a_t, r(s_t, a_t), s_{t+1} )\right\}$
		\ENDFOR
		\FOR{each gradient step}
		\item $\ew \leftarrow \mathbb{E}_{\mathcal{D}}\left[\left(r_t+\gamma V_{\bar{\phi}}(s_{t+1})-Q_{\omega}(h_t)\right)^2\right]$
		\item $\phi \leftarrow \phi - \lambda_V \hat \nabla_\phi J^V(\phi)$ (M-step)
		\STATE $\omega \leftarrow \omega - \lambda_Q \hat \nabla_{\omega} J^Q(\omega)$ (M-step)
		\STATE $\theta \leftarrow \theta - \lambda_{\pi^q} \hat \nabla_\theta J^{\pi^q}_{\textit{beta}}(\theta)$ (E-step)
		\STATE $\bar{\phi} \leftarrow \tau \phi + (1-\tau)\bar{\phi}$
		\ENDFOR
		\ENDFOR
	\end{algorithmic}
\end{algorithm}
\end{minipage}
\end{multicols}

\newpage
\section{Experimental details}

\subsection{Parameter Values}
Note that instead of specifying temperature $c$, we fix $c=1$ for all implementations and scale reward instead.
\label{paremeter_values}
\begin{table}[h!]
\caption{Summary of Experimental Parameter Values} \label{table:parameters}
\vspace{0.3cm}
	\begin{center}
		\begin{tabular}{lc}
			\textbf{PARAMETER }&\textbf{VALUE} \\
			\hline \\
			Steps per evaluation & 1000 \\
			Path Length & 999 \\
			Discount factor & 0.99\\
			\addlinespace[0.3cm]
			\hline \\
			\multicolumn{2}{l}{\textbf{Mujoco-v2 Experiments:}} \\
			\addlinespace[0.3cm]
			Batch size & 128 \\
			Net size & 300 \\
			\addlinespace[0.2cm]
			\multirow{2}{*}{$\displaystyle \lambda_{\beta}\approx  \frac{1-\gamma}{r_{avg}}$} & \makecell[c]{Humanoid \\ 4e-4} \\
			 &\makecell[c]{All other \\ 4e-3} \\
			\addlinespace[0.2cm]
			\multirow{3}{*}{Reward scale} & \makecell[c]{Hopper, Half-Cheetah \\ 5} \\
			& \makecell[c]{Walker \\ 3} \\
			&\makecell[c]{All other \\ 1} \\
			\addlinespace[0.2cm]
			\makecell[l]{Value function\\learning rate} & 3e-4 \\
			\addlinespace[0.2cm]
			\makecell[l]{Policy\\learning rate} & 3e-4 \\
			\addlinespace[0.2cm]
			\multicolumn{2}{l}{\makecell[l]{MLP layout as given in\\ \url{https://github.com/vitchyr/rlkit}}}\\
			\addlinespace[0.3cm]
			\hline \\
			\multicolumn{2}{l}{\textbf{Mujoco-v1 Experiments:}} \\
			\addlinespace[0.3cm]
			\multicolumn{2}{l}{\makecell[l]{Values as used by \citet{Haarnoja18} in\\ \url{https://github.com/haarnoja/sac}}}
		\end{tabular}
	\end{center}
\end{table}
\onecolumn

\subsection{Additional MuJoCo-v1 Experiments}
\label{sec:v1_experiments}

\begin{figure*}[!htb]
\begin{center}
\minipage{0.33\textwidth}
  \includegraphics[width=\linewidth]{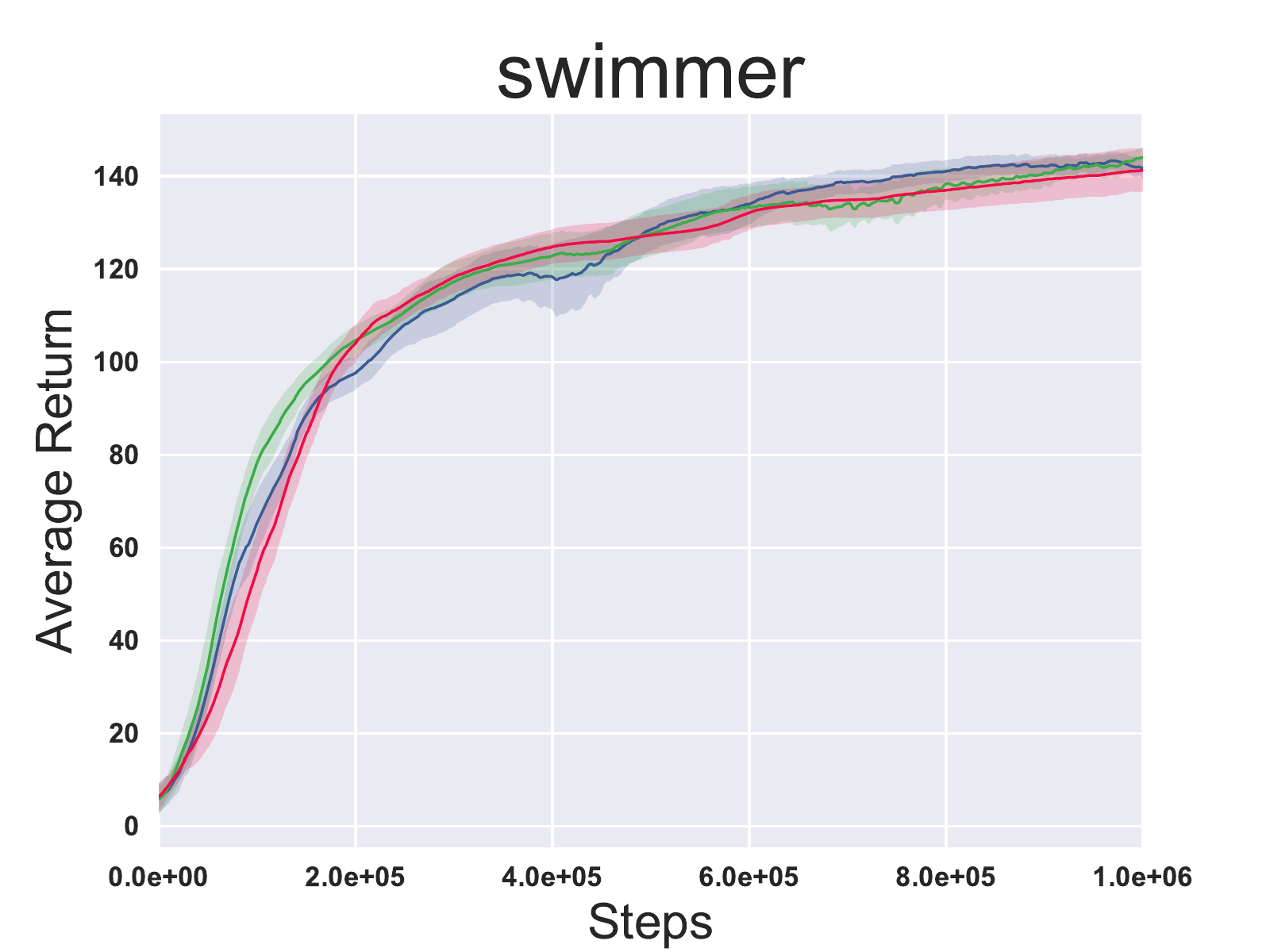}
  \label{fig:swimmer}
\endminipage
\minipage{0.33\textwidth}%
  \includegraphics[width=\linewidth]{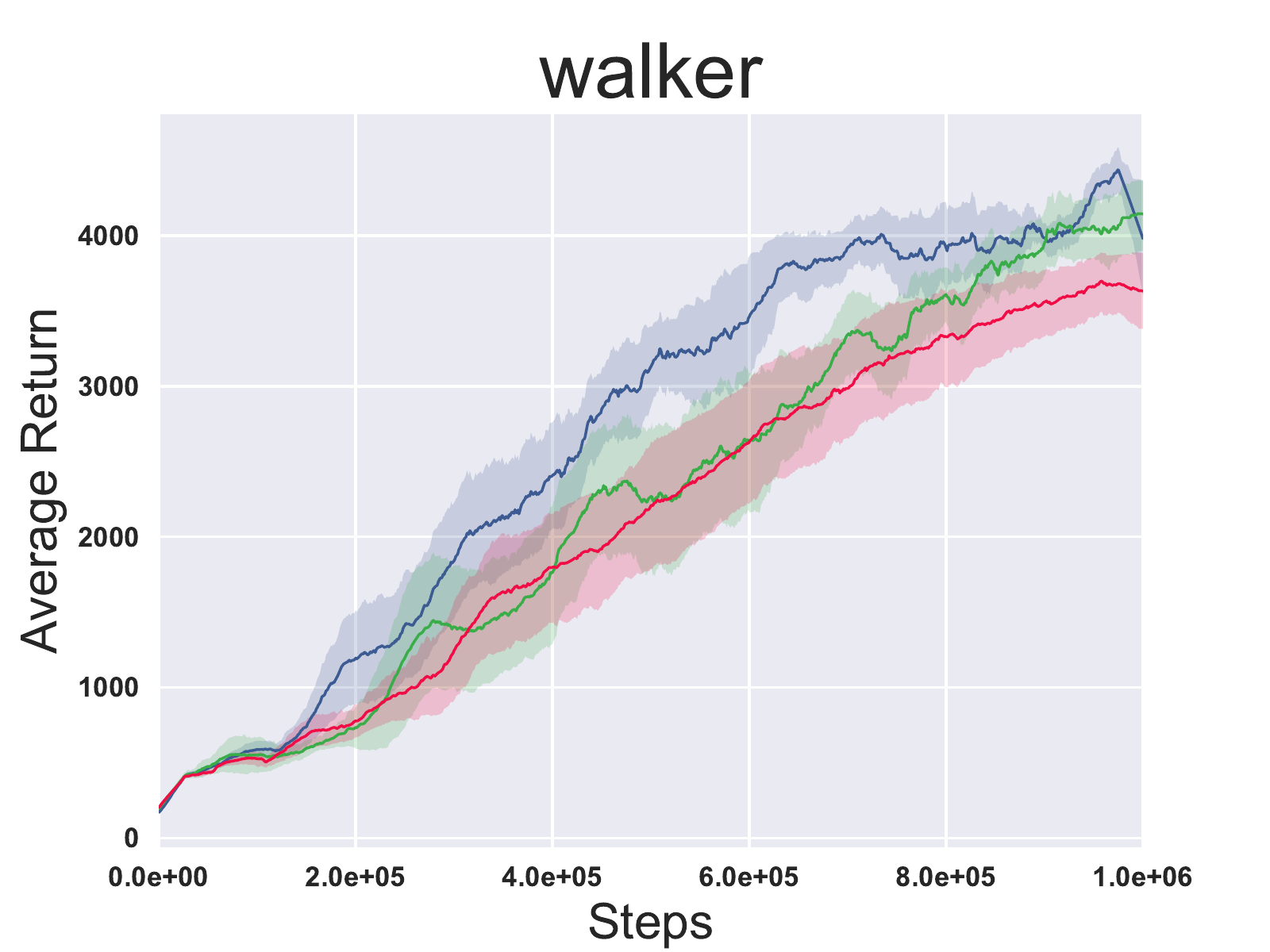}
  \label{fig:hopper}
\endminipage
\minipage{0.33\textwidth}
  \includegraphics[width=\linewidth]{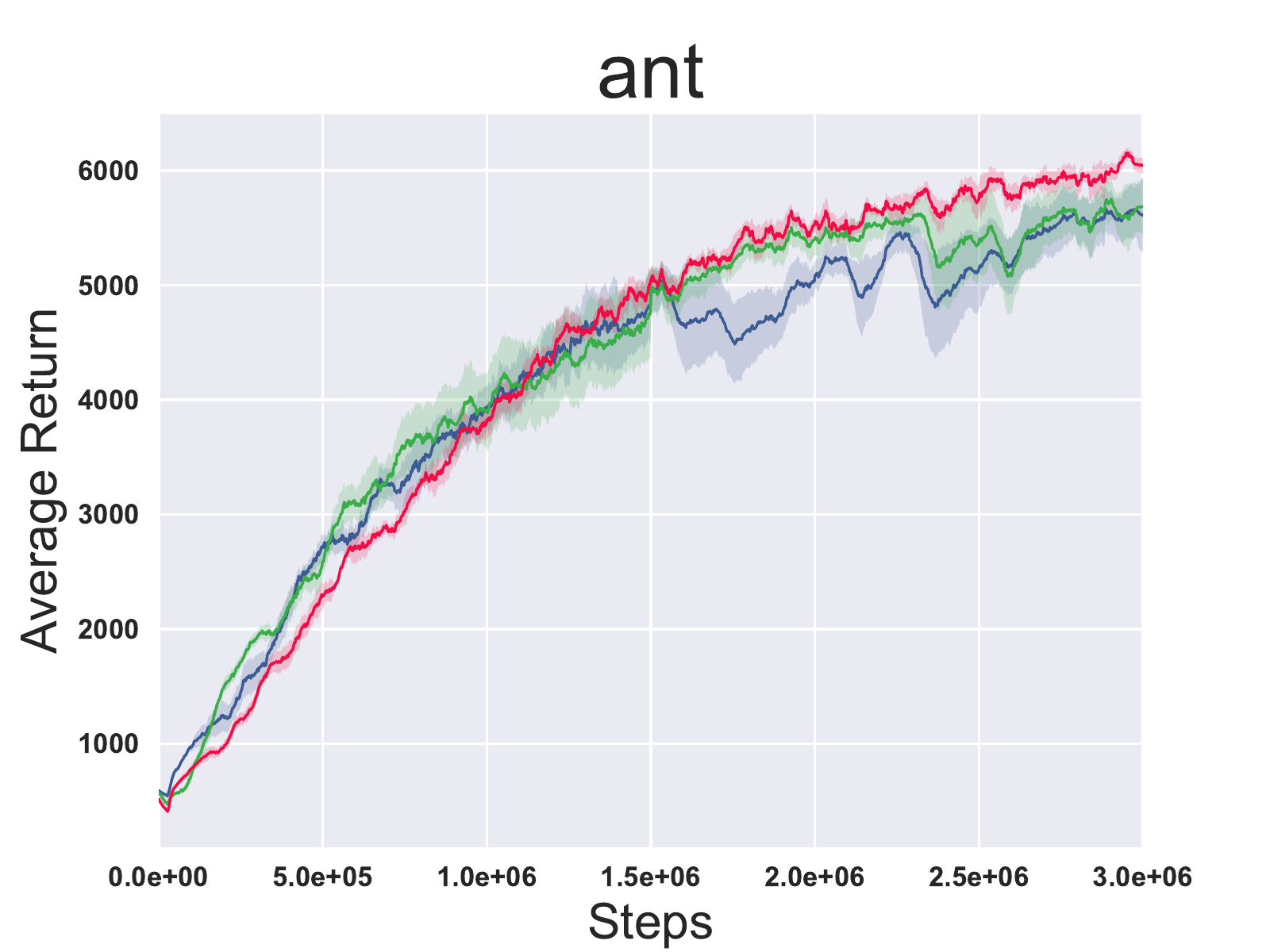}
  \label{fig:hopper}
\endminipage\\
\end{center}
\caption{Training curves on additional continuous control benchmarks Mujoco-v1.}
\label{experimentsv1_app}
\end{figure*}

\subsection{Additional MuJoCo-v2 Experiments}
\label{sec:v2_experiments}
\begin{figure*}[!htb]
	\begin{center}
		\minipage{0.33\textwidth}%
		\includegraphics[width=\linewidth]{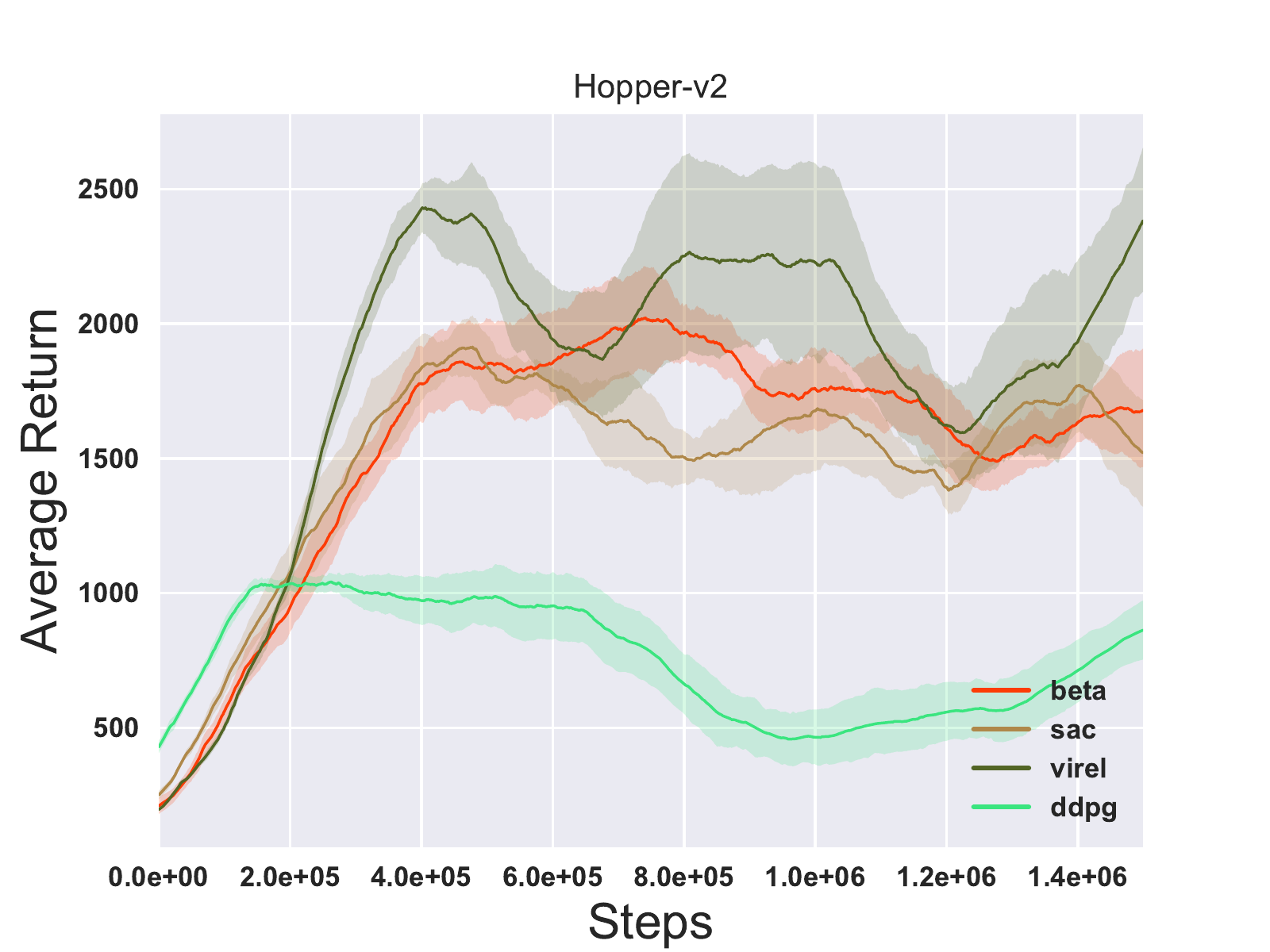}
		\label{fig:hopperv2}
		\endminipage
		\minipage{0.33\textwidth}
		\includegraphics[width=\linewidth]{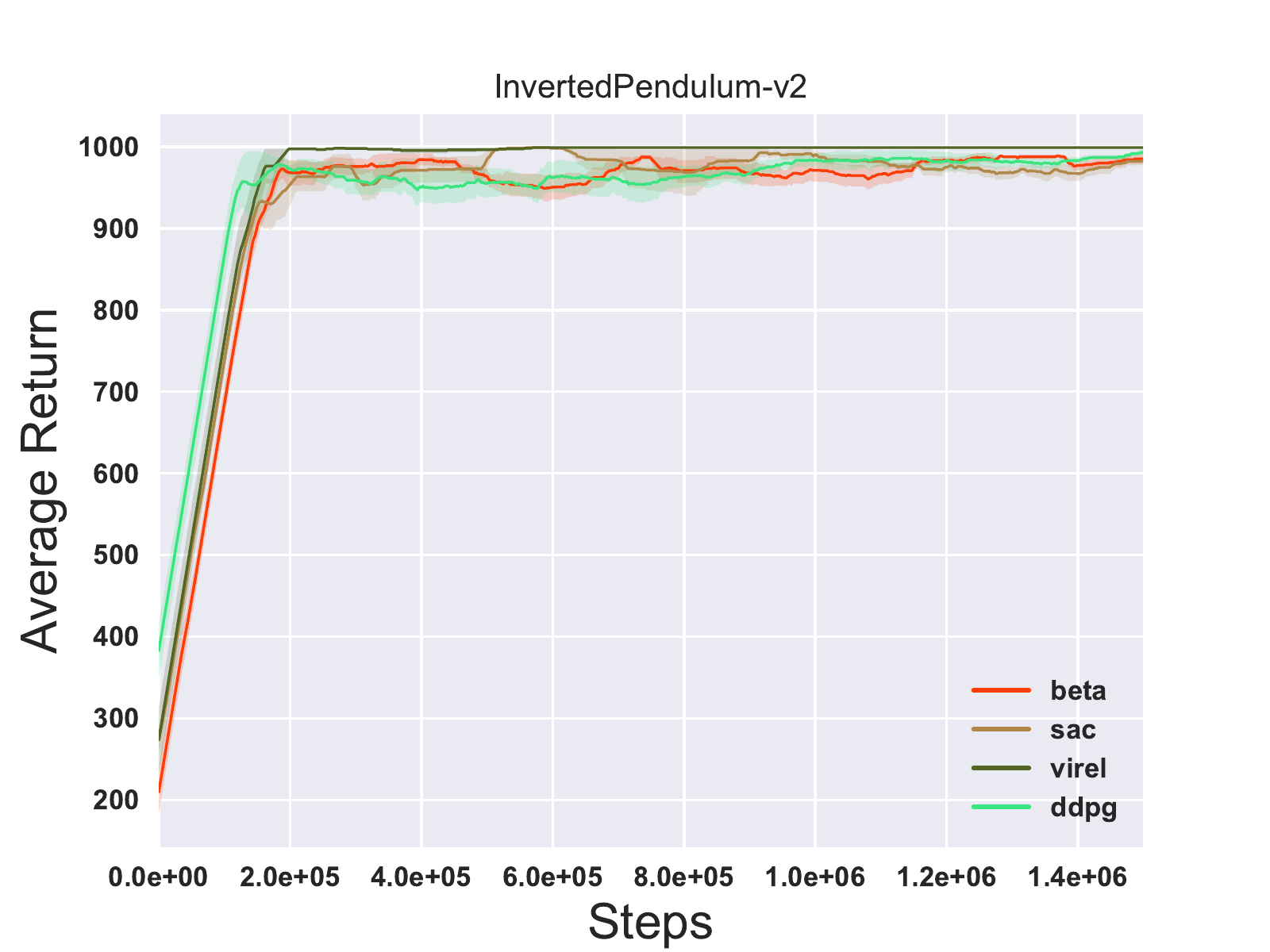}
		\label{fig:ipv2}
		\endminipage

		\minipage{0.33\textwidth}
		\includegraphics[width=\linewidth]{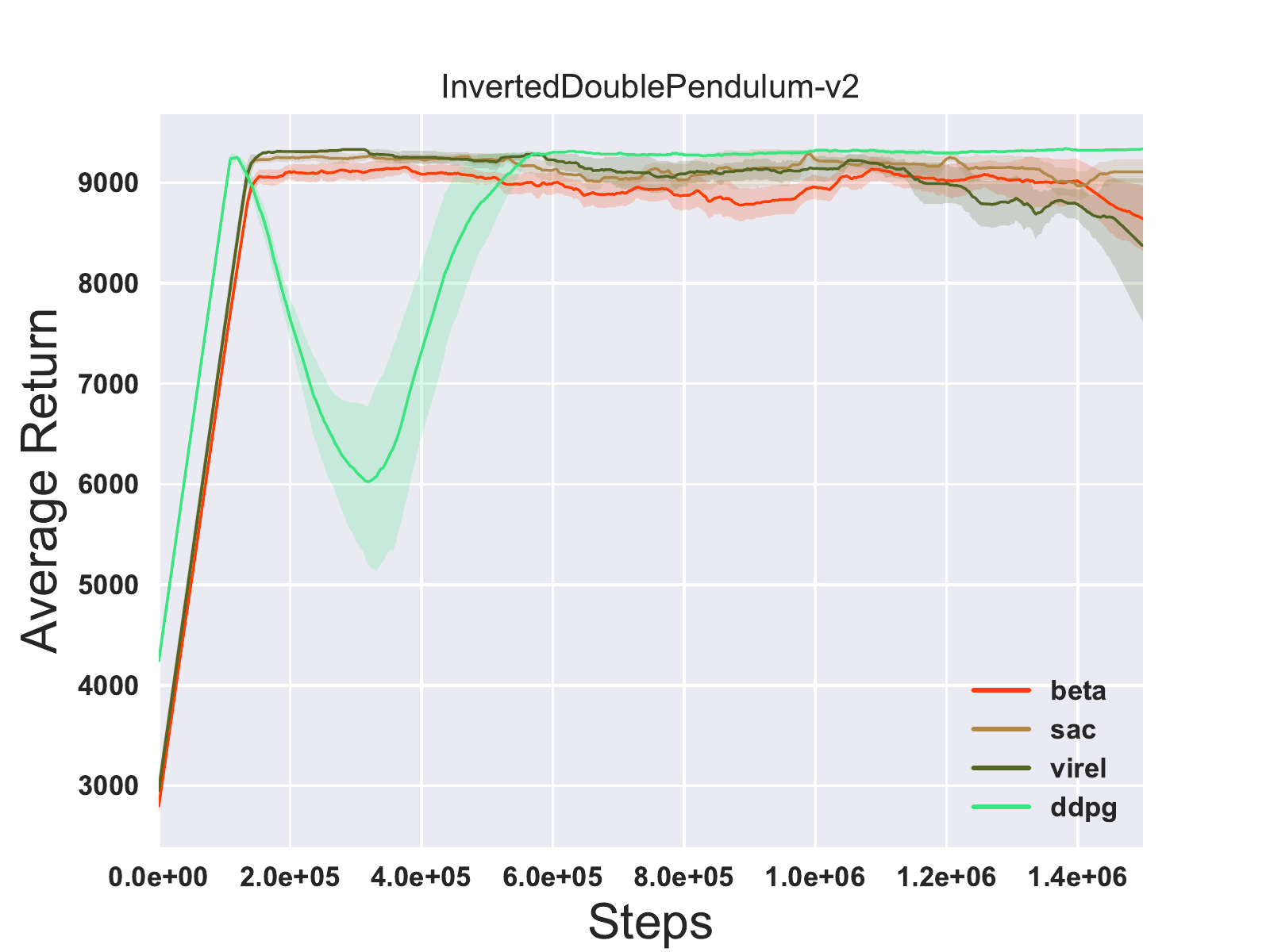}
		\label{fig:idpv2}
		\endminipage\\

		\minipage{0.33\textwidth}
		\includegraphics[width=\linewidth]{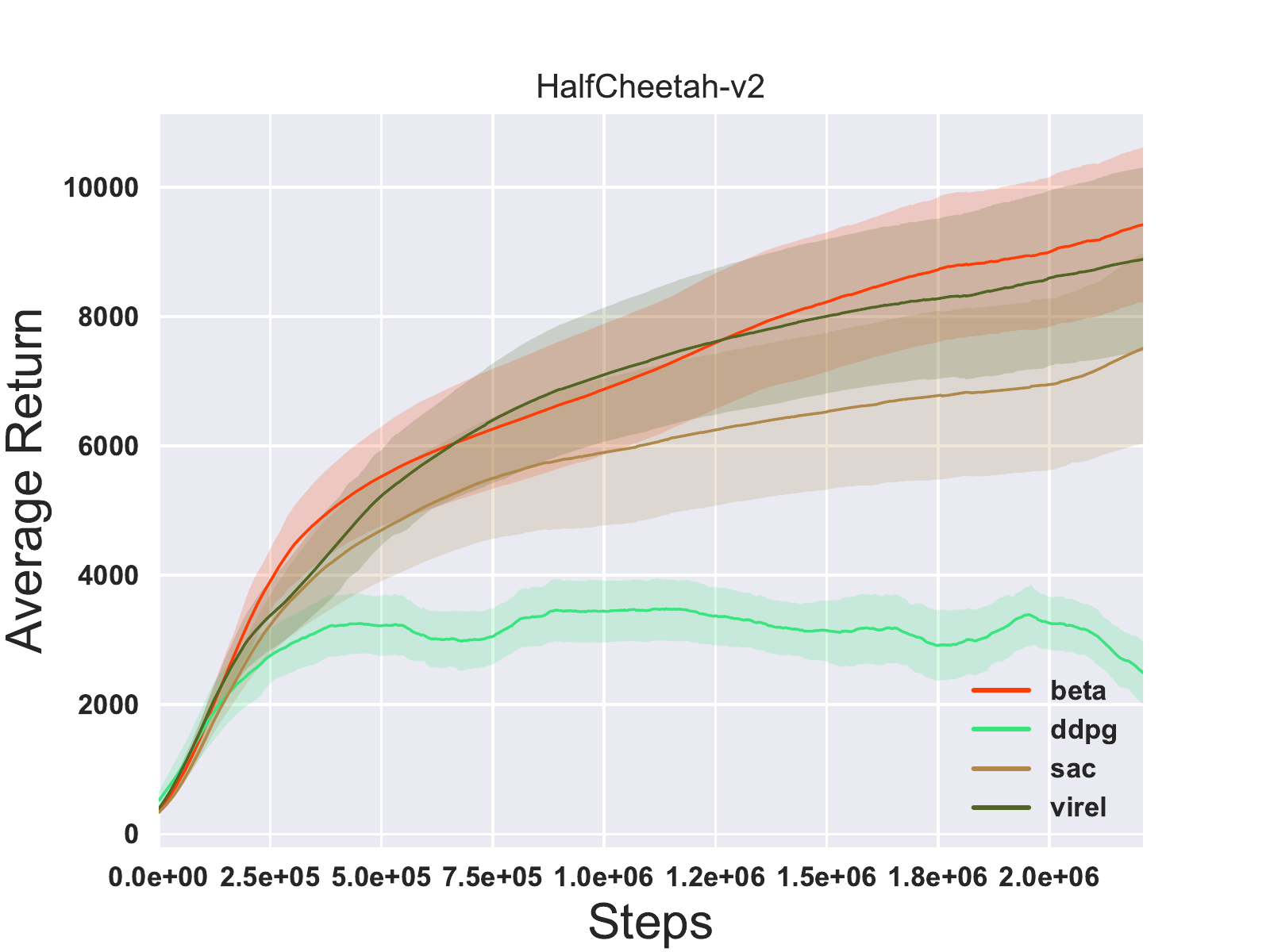}
		\label{fig:antv2}
		\endminipage
		\minipage{0.33\textwidth}%
		\includegraphics[width=\linewidth]{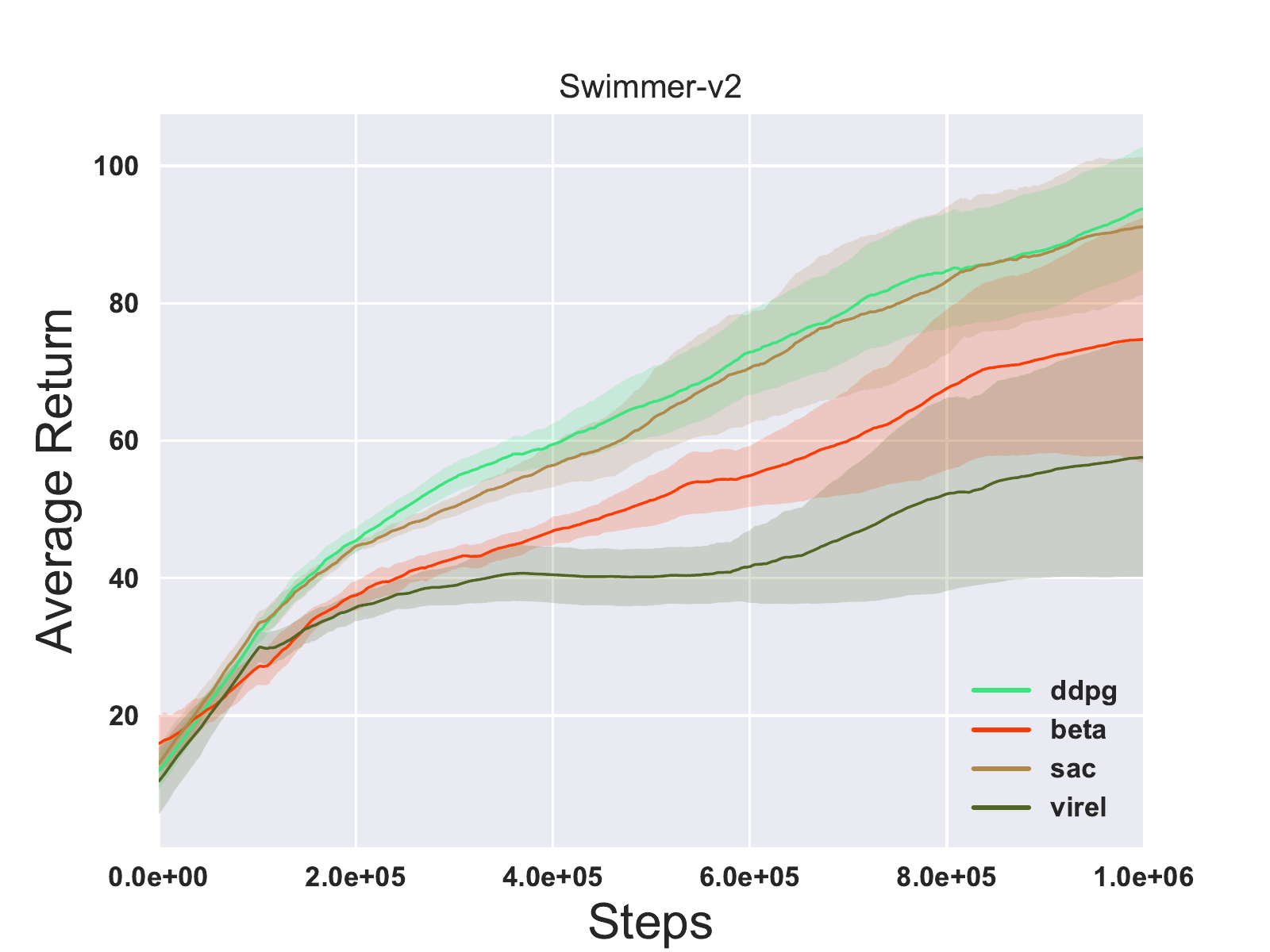}
		\label{fig:hopperv2}
		\endminipage
		\minipage{0.33\textwidth}%
		\includegraphics[width=\linewidth]{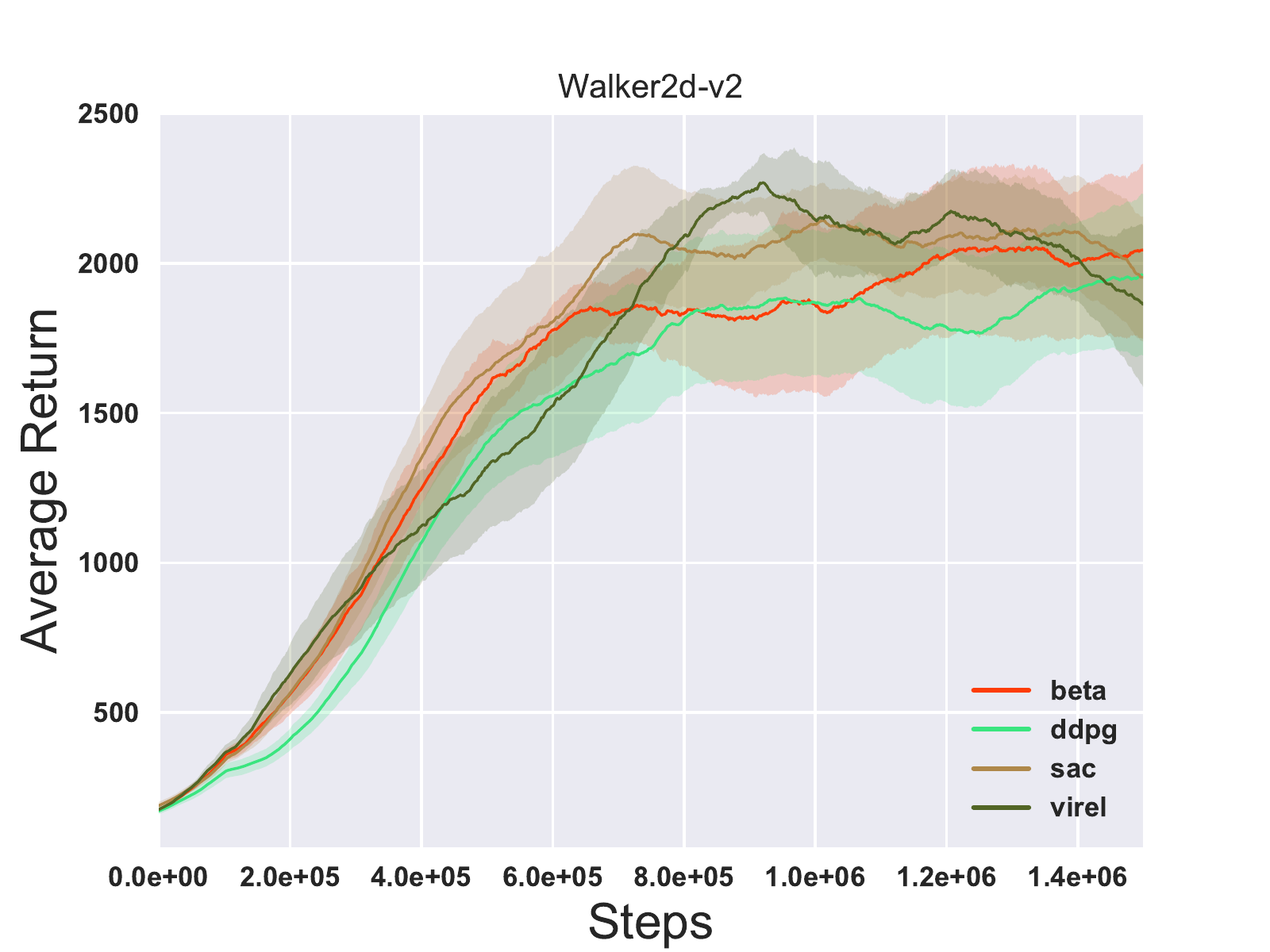}
		\label{fig:hopperv2}
		\endminipage\\
	\end{center}
	\caption{Training curves on additional continuous control benchmarks gym-Mujoco-v2.}
	\label{experimentsv2_app}
\end{figure*}